\documentclass[10pt,twocolumn,letterpaper]{article}

\usepackage[pagenumbers]{cvpr} %

\usepackage{animate}
\usepackage{makecell}
\usepackage{graphicx}
\usepackage{tabularx}
\usepackage{amsmath}
\usepackage{amssymb}
\usepackage{booktabs}
\usepackage{url}
\usepackage{epsfig}
\usepackage{lipsum}
\usepackage{amsthm}
\usepackage{comment}
\usepackage{multirow}
\usepackage{subcaption}
\usepackage{microtype}
\usepackage{xspace}
\usepackage[dvipsnames,table]{xcolor}
\usepackage{enumitem}
\usepackage{tabularx}
\usepackage{graphbox} %
\usepackage{xcolor}         %
\usepackage[accsupp]{axessibility}
\definecolor{myblue}{HTML}{5364cc}
\definecolor{myorange}{HTML}{ED7D31}

\usepackage[pagebackref,breaklinks,colorlinks]{hyperref}

\usepackage{nimbusmononarrow}

\usepackage[capitalize]{cleveref}
\crefname{section}{Sec.}{Secs.}
\Crefname{section}{Section}{Sections}
\Crefname{table}{Table}{Tables}
\crefname{table}{Tab.}{Tabs.}

\def\cvprPaperID{???} %
\def\confName{CVPR}
\def\confYear{2023}

\usepackage{amsmath,amsfonts,bm}

\def\eqref#1{equation~\ref{#1}}

\def\1{\bm{1}}

\def\vs{{\bm{s}}}

\DeclareMathAlphabet{\mathsfit}{\encodingdefault}{\sfdefault}{m}{sl}
\SetMathAlphabet{\mathsfit}{bold}{\encodingdefault}{\sfdefault}{bx}{n}

\definecolor{citecolor}{rgb}{0.4,0.7,0.25}
\hypersetup{
    colorlinks,
    citecolor=citecolor,
}

\newcommand{\name}{{Magic3D}\xspace}
\newcommand{\ediffi}{{eDiff-I}\xspace}
\newcommand{\V}{\texttt{[V]}\xspace}

\def\vs{\emph{vs}\onedot}

\begin{document}

\title{Magic3D: High-Resolution Text-to-3D Content Creation}

\author{Chen-Hsuan Lin$^*$ \quad Jun Gao$^*$ \quad Luming Tang$^*$ \quad Towaki Takikawa$^*$ \quad Xiaohui Zeng$^*$ \\
Xun Huang \quad Karsten Kreis \quad Sanja Fidler$^\dagger$ \quad Ming-Yu Liu$^\dagger$ \quad Tsung-Yi Lin
\vspace{0.2cm} \\
NVIDIA Corporation \vspace{2pt} \\
{\fontsize{10}{10}\selectfont \url{https://research.nvidia.com/labs/dir/magic3d}}
}

\maketitle
\def\thefootnote{*$\dagger$}\footnotetext{:~equal contribution.}\def\thefootnote{\arabic{footnote}}

\begin{abstract}

DreamFusion~\cite{poole2022dreamfusion} has recently demonstrated the utility of a pre-trained text-to-image diffusion model to optimize Neural Radiance Fields (NeRF)~\cite{mildenhall2020nerf},
achieving remarkable text-to-3D synthesis results.
However, the method has two %
inherent limitations: (a) extremely slow optimization of NeRF and
(b) low-resolution image space supervision on NeRF,
leading to low-quality 3D models with a long processing time.
In this paper, we address these limitations by utilizing a two-stage optimization framework.
First, we obtain a coarse model using a low-resolution diffusion prior and accelerate with a sparse 3D hash grid structure.
Using the coarse representation as the initialization, we further optimize a textured 3D mesh model with an efficient differentiable renderer interacting with a high-resolution latent diffusion model.
Our method, dubbed \name, can create %
high quality 3D mesh models in 40 minutes, which is $2\times$ faster than DreamFusion (reportedly taking 1.5 hours on average), while also achieving higher resolution.
User studies show 61.7\% raters to prefer our approach over DreamFusion.
Together with the image-conditioned generation capabilities,
we provide users with new ways to control 3D synthesis, opening up new avenues to various creative applications.

\end{abstract}

\section{Introduction}
3D digital content has been in high demand for a variety of applications, including gaming, entertainment, architecture, and robotics simulation.
It is slowly finding its way into virtually every possible domain: retail, online conferencing, virtual social presence, education, \etc.
However, creating professional 3D content is not for anyone --- it requires immense artistic and aesthetic training with 3D modeling expertise.
Developing these skill sets takes a significant amount of time and effort.
Augmenting 3D content creation with natural language could considerably help democratize 3D content creation for novices and turbocharge expert artists.

Image content creation from text prompts~\cite{nichol2021glide,saharia2022photorealistic,ramesh2022hierarchical,balaji2022ediffi} has seen significant progress with the advances of diffusion models~\cite{sohl2015deep,song2019generative,ho2020denoising} for generative modeling of images.
The key enablers are large-scale datasets comprising billions of samples (images with text) scrapped from the Internet and massive amounts of compute.
In contrast, 3D content generation has progressed at a much slower pace.
Existing 3D object generation models~\cite{chan2022efficient,gao2022get3d,zeng2022lion} are mostly categorical.
A trained model can only be used to synthesize objects for a single class, with early signs of scaling to multiple classes shown recently by Zeng~\etal~\cite{zeng2022lion}.
Therefore, what a user can do with these models is extremely limited and not yet ready for artistic creation.
This limitation is largely due to the lack of diverse large-scale 3D datasets --- compared to image and video content, 3D content is much less accessible on the Internet.
This naturally raises the question of whether 3D generation capability can be achieved by leveraging powerful text-to-image generative models.

Recently, DreamFusion~\cite{poole2022dreamfusion} demonstrated its remarkable ability for text-conditioned 3D content generation by utilizing a pre-trained text-to-image diffusion model~\cite{saharia2022photorealistic} that generates images as a strong image prior.
The diffusion model acts as a critic to optimize the underlying 3D representation.
The optimization process ensures that rendered images from a 3D model, represented by Neural Radiance Fields (NeRF)~\cite{mildenhall2020nerf}, match the distribution of photorealistic images across different viewpoints, given the input text prompt.
Since the supervision signal in DreamFusion operates on very low-resolution images ($64\times64$), DreamFusion cannot synthesize high-frequency 3D geometric and texture details.
Due to the use of inefficient MLP architectures for the NeRF representation, practical high-resolution synthesis may not even be possible as the required memory footprint and the computation budget grows quickly with the resolution.
Even at a resolution of $64\times64$, optimization times are in hours (1.5 hours per prompt on average using TPUv4). 
\begin{figure*}[!]
    \centering
    \includegraphics[width=\linewidth,trim={0cm 1.9cm 0.0cm 0.0cm}, clip]{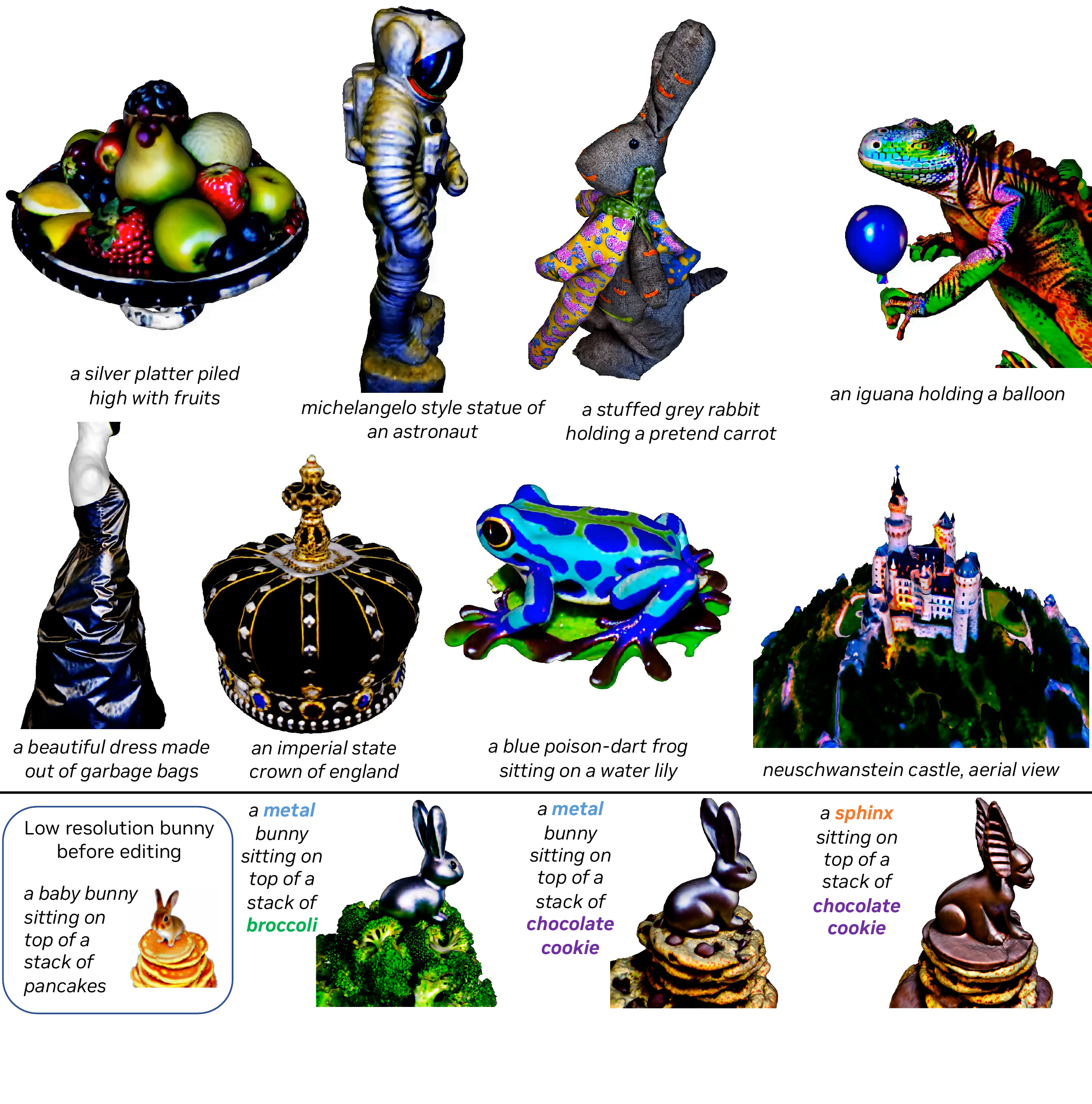}
    \caption{
    \textbf{Results and applications of \name.}
    \textbf{Top: high-resolution text-to-3D generation}.
    \name can generate high-quality and high-resolution 3D models from text prompts.
    \textbf{Bottom: high-resolution prompt-based editing}.
    \name can edit 3D models by fine-tuning with the diffusion prior using a different prompt.
    Taking the low-resolution 3D model as the input (left), \name can modify different parts of the 3D model corresponding to different input text prompts.
    Together with various creative controls on the generated 3D models, \name is a convenient tool for augmenting 3D content creation.
    }
    \label{fig:teaser_2x4}
\end{figure*}

In this paper, we present a method that can synthesize highly detailed 3D models from text prompts within a reduced computation time.
Specifically, we propose a coarse-to-fine optimization approach that uses multiple diffusion priors at different resolutions to optimize the 3D representation, 
enabling the generation of both view-consistent geometry as well as high-resolution details.
In the first stage, we optimize a coarse neural field representation akin to DreamFusion, but with
a memory- and compute-efficient scene representation based on a hash grid~\cite{muller2022instant}.
In the second stage, 
we switch to optimizing mesh representations, a critical step that allows us to utilize diffusion priors at resolutions as high as $512\times 512$.
As 3D meshes are amenable to fast graphics renderers that can render high-resolution images in real-time, we leverage an efficient differentiable rasterizer~\cite{nvdiffrec,gao2022get3d} and make use of camera close-ups to recover high-frequency details in geometry and texture.
As a result, our approach produces high-fidelity 3D content (see Fig.~\ref{fig:teaser_2x4}) that can conveniently be imported and visualized in standard graphics software and does so at 2$\times$ the speed of DreamFusion.
Furthermore, we showcase various creative controls over the 3D synthesis process by leveraging the advancements developed for text-to-image editing applications~\cite{balaji2022ediffi, ruiz2022dreambooth}.
Our approach, dubbed {\name}, endows users with unprecedented control in crafting their desired 3D objects with text prompts and reference images, bringing this technology one step closer to democratizing 3D content creation.

In summary, our work makes the following contributions:
\begin{itemize}[leftmargin=*]
    \setlength\itemsep{0pt}
    \item We propose \name, a framework for high-quality 3D content synthesis using text prompts by improving several major design choices made in DreamFusion.
    It consists of a coarse-to-fine strategy that leverages both low- and high-resolution diffusion priors for learning the 3D representation of the target content.
    \name, which synthesizes 3D content with an 8$\times$ higher resolution supervision, is also 2$\times$ faster than DreamFusion. 3D content synthesized by our approach is significantly preferable by users (61.7\%).
    \item We extend various image editing techniques developed for text-to-image models to 3D object editing and show their applications in the proposed framework.
\end{itemize}

\section{Related Work}

\vspace{4pt}
\noindent\textbf{Text-to-image generation.}
We have witnessed significant progress in text-to-image generation with diffusion models in recent years.
With improvements in modeling and data curation, diffusion models can compose complex semantic concepts from text descriptions (nouns, adjectives, artistic styles, \etc) to generate high-quality images of objects and scenes~\cite{ramesh2022hierarchical,saharia2022photorealistic,Rombach_2022_CVPR, balaji2022ediffi}.
Sampling images from diffusion models is time consuming.
To generate high-resolution images, these models either utilize a cascade of super-resolution models~\cite{saharia2022photorealistic,balaji2022ediffi} or sample from a lower-resolution latent space and decode latents into high-resolution images~\cite{Rombach_2022_CVPR}.
Despite the advances in high-resolution image generation, using language to describe and control 3D properties (\eg camera viewpoints) while maintaining coherency in 3D remains an open, challenging problem.

\vspace{4pt}
\noindent\textbf{3D generative models.} %
There is a large body of work on 3D generative modeling,  exploring different types of 3D representations such as 3D voxel grids~\cite{wu2016learning,gadelha20173d,henzler2019platonicgan,lunz2020inverse,smith2017improved}, point-clouds~\cite{achlioptas2018learning, pointflow,mo2019structurenet,zhou2021pvd,luo2021diffusion,zeng2022lion}, meshes~\cite{zhang2020image,gao2022get3d}, implicit~\cite{occnet,chen2019learning}, or octree~\cite{ibing2021octree} representations.
Most of these approaches rely on training data in the form of 3D assets, which are hard to acquire at scale.
Inspired by the success of neural volume rendering~\cite{mildenhall2020nerf}, recent works started investing in 3D-aware image synthesis~\cite{nguyen2019hologan,chan2021pi,niemeyer2021giraffe,hao2021GANcraft,gu2021stylenerf,chan2022efficient,orel2021styleSDF,schwarz2022voxgraf}, which has the advantage of learning 3D generative models directly from images --- a more widely accessible resource.
However, volume rendering networks are typically slow to query, leading to a trade-off between long training time~\cite{chan2021pi,niemeyer2021giraffe} and lack of multi-view consistency~\cite{gu2021stylenerf}.
EG3D~\cite{chan2022efficient} partially mitigates this problem by utilizing a dual discriminator.
While obtaining promising results, these works remain limited to modeling objects within a single object category, such as cars, chairs, or human faces, thus lacking scalability and the creative control desired for 3D content creation.
In our paper, we focus on text-to-3D synthesis, aiming to generate a 3D renderable representation of a scene based on a text prompt.

\vspace{4pt}
\noindent\textbf{Text-to-3D generation.}
With the recent success in text-to-image generative modeling in recent years, text-to-3D generation has also gained a surge of interest from the learning community.
Earlier works such as CLIP-forge~\cite{sanghi2022clip} synthesizes objects by learning a normalizing flow model to sample shape embeddings from textual input.
However, it requires 3D assets in voxel representations during training, making it challenging to scale with data.
DreamField~\cite{jain2021dreamfields} and CLIP-mesh~\cite{khalid2022clip} mitigate the training data issue by relying on a pre-trained image-text model~\cite{radford2021learning} to optimize the underlying 3D representations (NeRFs and meshes), such that all 2D renderings reach high text-image alignment scores.
While these approaches avoid the requirement of expensive 3D training data and mostly rely on pre-trained large-scale image-text models, they tend to produce less realistic 2D renderings. 

Recently, DreamFusion~\cite{poole2022dreamfusion} %
showcased impressive capability in text-to-3D synthesis by utilizing a powerful pre-trained text-to-image diffusion model~\cite{saharia2022photorealistic} as a strong image prior.
We build upon this work and improve over several design choices to bring significantly higher-fidelity 3D models into hands of users with a much reduced generation time.

\begin{figure*}[t!]
    \centering
    \includegraphics[width=\linewidth,page=1]{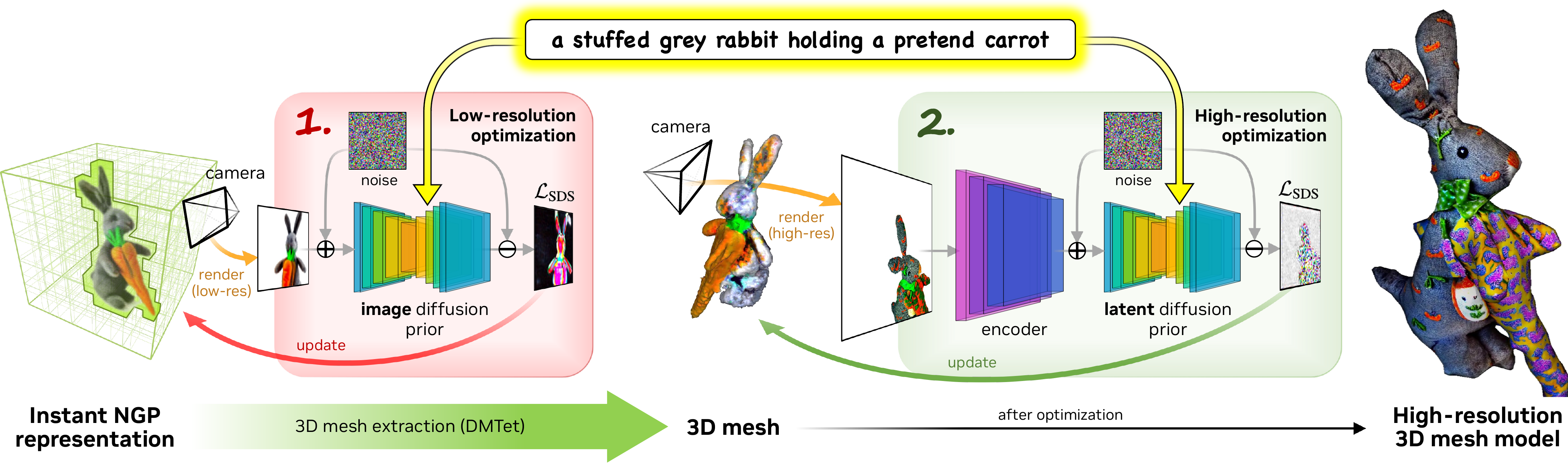}
    \vspace{-3mm}
    \caption{
        \textbf{Overview of \name.}
        We generate high-resolution 3D content from an input text prompt in a coarse-to-fine manner.
        In the first stage, we utilize a low-resolution diffusion prior and optimize neural field representations (color, density, and normal fields) to obtain the coarse model.
        We further differentiably extract textured 3D mesh from the density and color fields of the coarse model. Then we fine-tune it using a high-resolution latent diffusion model.
        After optimization, our model generates high-quality 3D meshes with detailed textures. 
    }
    \label{fig:diagram}
\end{figure*}

\section{Background: DreamFusion}
DreamFusion~\cite{poole2022dreamfusion} achieves text-to-3D generation with two key components: a neural scene representation which we refer to as the scene model, and a pre-trained text-to-image diffusion-based generative model.
The scene model is a parametric function $x = g(\theta)$, which can produce an image $x$ at the desired camera pose.
Here, $g$ is a volumetric renderer of choice, and $\theta$ is a coordinate-based MLP representing a 3D volume.
The diffusion model $\phi$ comes with a learned denoising function $\epsilon_\phi(x_t;y,t)$ that predicts the sampled noise $\epsilon$ given the noisy image $x_t$, noise level $t$, and text embedding $y$.
It provides the gradient direction to update $\theta$ such that all rendered images are pushed to the high probability density regions conditioned on the text embedding under the diffusion prior.
Specifically, DreamFusion introduces Score Distillation Sampling (SDS), which computes the gradient: 
\begin{equation}
    \nabla_\theta \mathcal{L}_\text{SDS}(\phi, g(\theta)) = 
    \mathbb{E}_{t, \epsilon} \! \! \left[ w(t)(\epsilon_\phi(x_t;y,t) - \epsilon)\frac{\partial x}{\partial \theta} \right]. 
\end{equation}
Here, $w(t)$ is a weighting function. We view the scene model $g$ and the diffusion model $\phi$ as modular components of the framework, amenable to choice.
In practice, the denoising function $\epsilon_\phi$ is often replaced with another function $\tilde{\epsilon}_\phi$ that uses classifier-free guidance~\cite{ho2021classifierfree}, which allows one to carefully weigh the strength of the text conditioning (see Sec.~\ref{sec:controllable}).
DreamFusion relies on large classifier-free guidance weights to obtain results with better quality.

DreamFusion adopts a variant of Mip-NeRF 360~\cite{barron2022mip} with an explicit shading model for the scene model and Imagen~\cite{saharia2022photorealistic} as the diffusion model.
These choices result in two key limitations.
First, high-resolution geometry or textures cannot be obtained since the diffusion model only operates on $64 \times 64$ images.
Second, the utility of a large global MLP for volume rendering is both computationally expensive as well as memory intensive, making this approach scale poorly with the increasing resolution of images.

\section{High-Resolution 3D Generation}

\name is a two-stage coarse-to-fine 
framework that uses efficient scene models that enable high-resolution text-to-3D synthesis (Fig.~\ref{fig:diagram}).
We describe our method and key differences from DreamFusion~\cite{poole2022dreamfusion} in this section.

\subsection{Coarse-to-fine Diffusion Priors}
\label{diffusion_prior}
\name uses two different diffusion priors in a coarse-to-fine fashion to generate high-resolution geometry and textures.
In the first stage, we use the base diffusion model described in \ediffi~\cite{balaji2022ediffi}, which is similar to the base diffusion model of Imagen~\cite{saharia2022photorealistic} used in DreamFusion.
This diffusion prior is used to compute gradients of the scene model via a loss defined on rendered images at a low resolution $64\times 64$.
In the second stage, we use the latent diffusion model (LDM)~\cite{Rombach_2022_CVPR} that allows backpropagating gradients into rendered images at a high resolution $512\times 512$; in practice, we choose to use the publicly available \textit{Stable Diffusion} model~\cite{Rombach_2022_CVPR}.
Despite generating high-resolution images, the computation of LDM is manageable because the diffusion prior acts on the latent $z_t$ with resolution $64\times 64$: 
\begin{equation}
\label{eq:sds_high}
    \hspace*{-8pt} \nabla_\theta \mathcal{L}_\text{SDS}(\phi, g(\theta)) = 
    \mathbb{E}_{t, \epsilon} \!\! \left[ w(t)(\epsilon_\phi(z_t;y,t) \!-\! \epsilon)\frac{\partial z}{\partial x}\frac{\partial x}{\partial \theta} \! \right].
\end{equation} 
The increase in computation time mainly comes from computing ${\partial x} /\ { \partial \theta}$ (the gradient of the high-resolution rendered image) and ${\partial z} /\ {\partial x}$ (the gradient of the encoder in LDM).

\subsection{Scene Models}
\label{sec:scene_model}

We cater two different 3D scene representations to the two different diffusion priors at coarse and fine resolutions to accommodate the increased resolution of rendered images for the input of high-resolution priors, discussed as follows.

\vspace{4pt}
\noindent\textbf{Neural fields as coarse scene models.}
The initial coarse stage of the optimization requires finding the geometry and textures from scratch.
This can be challenging as we need to accommodate complex topological changes in the 3D geometry and depth ambiguities from the 2D supervision signals.
In DreamFusion~\cite{poole2022dreamfusion}, the scene model is a neural field (a coordinate-based MLP) based on Mip-NeRF 360~\cite{barron2022mip} that predicts albedo and density.
This is a suitable choice as neural fields can handle topological changes in a smooth, continuous fashion.
However, Mip-NeRF 360~\cite{barron2022mip} is computationally expensive as it is based on a large global coordinate-based MLP.
As volume rendering requires dense samples along a ray to accurately render high-frequency geometry and shading, the cost of having to evaluate a large neural network at every sample point quickly stacks up.

For this reason, we opt to use the hash grid encoding from Instant NGP~\cite{muller2022instant}, which allows us to represent high-frequency details at a much lower computational cost.
We use the hash grid with two single-layer neural networks, one predicting albedo and density and the other predicting normals.
We additionally maintain a spatial data structure that encodes scene occupancy and utilizes empty space skipping~\cite{liu2020neural,takikawa2021neural}.
Specifically, we use the density-based voxel pruning approach from Instant NGP~\cite{muller2022instant} with an octree-based ray sampling and rendering algorithm~\cite{KaolinWispLibrary}.
With these design choices, we drastically accelerate the optimization of coarse scene models while maintaining quality.

\begin{figure}[t!]
\centering 
    \includegraphics[width=\linewidth]{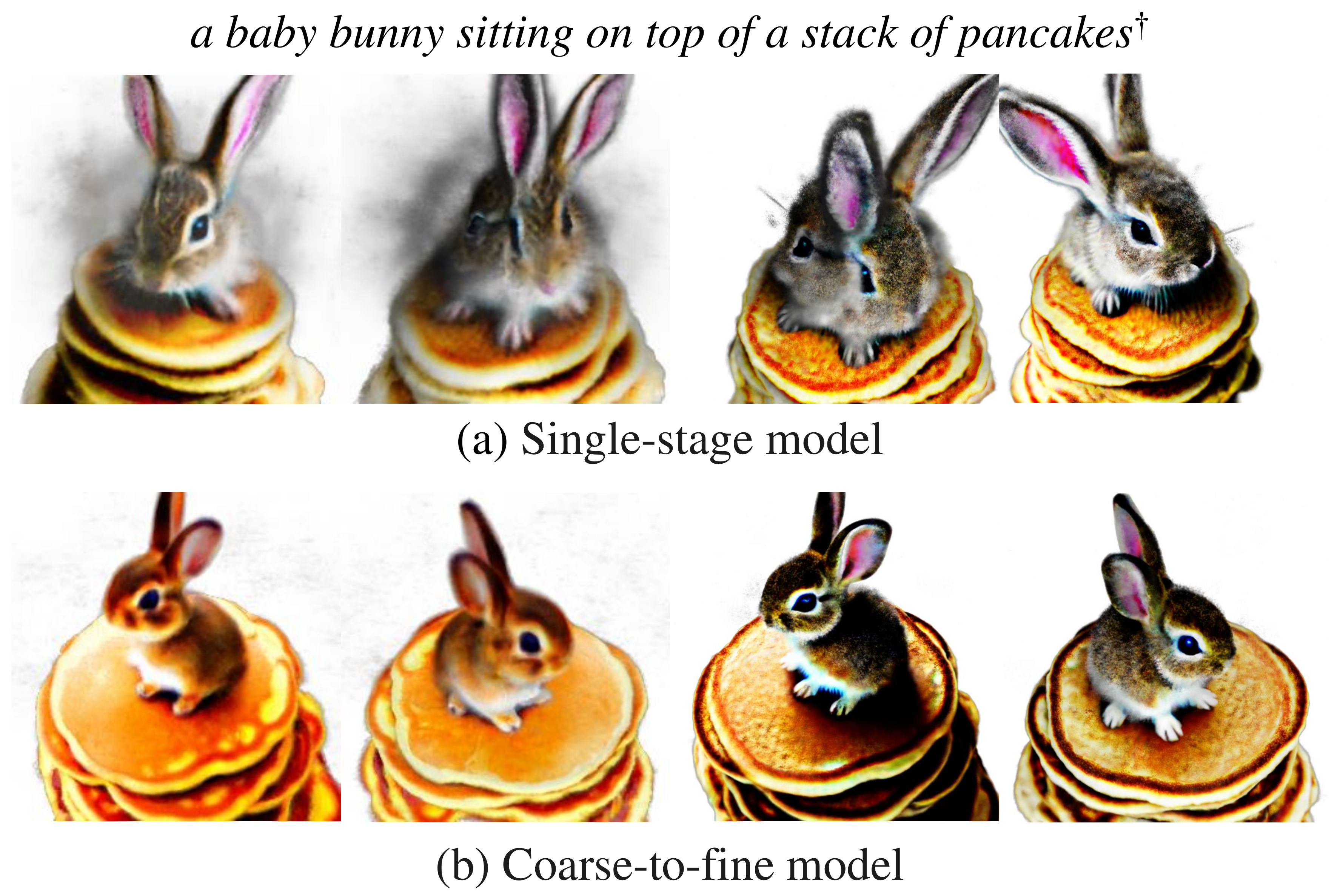}
    \caption{\textbf{Single-stage (top) vs. coarse-to-fine models (bottom).} Both use NeRF for the scene model. The left two columns use 64$\times$64 rendering resolution during optimization while the right two use 256$\times$256. Compared to our coarse-to-fine approach, the single-stage method can generate details but with worse shapes.
    }
    \label{fig:scratch_vs_2stage}
\end{figure}

\begin{figure*}
    \centering 
    \setlength{\tabcolsep}{0.2pt}
    \begin{tabular}{ccccc} 
        \makecell{\scriptsize Coarse\\ \scriptsize NeRF}& \includegraphics[align=c, width=0.23\linewidth,trim=0 0 0 100,clip]{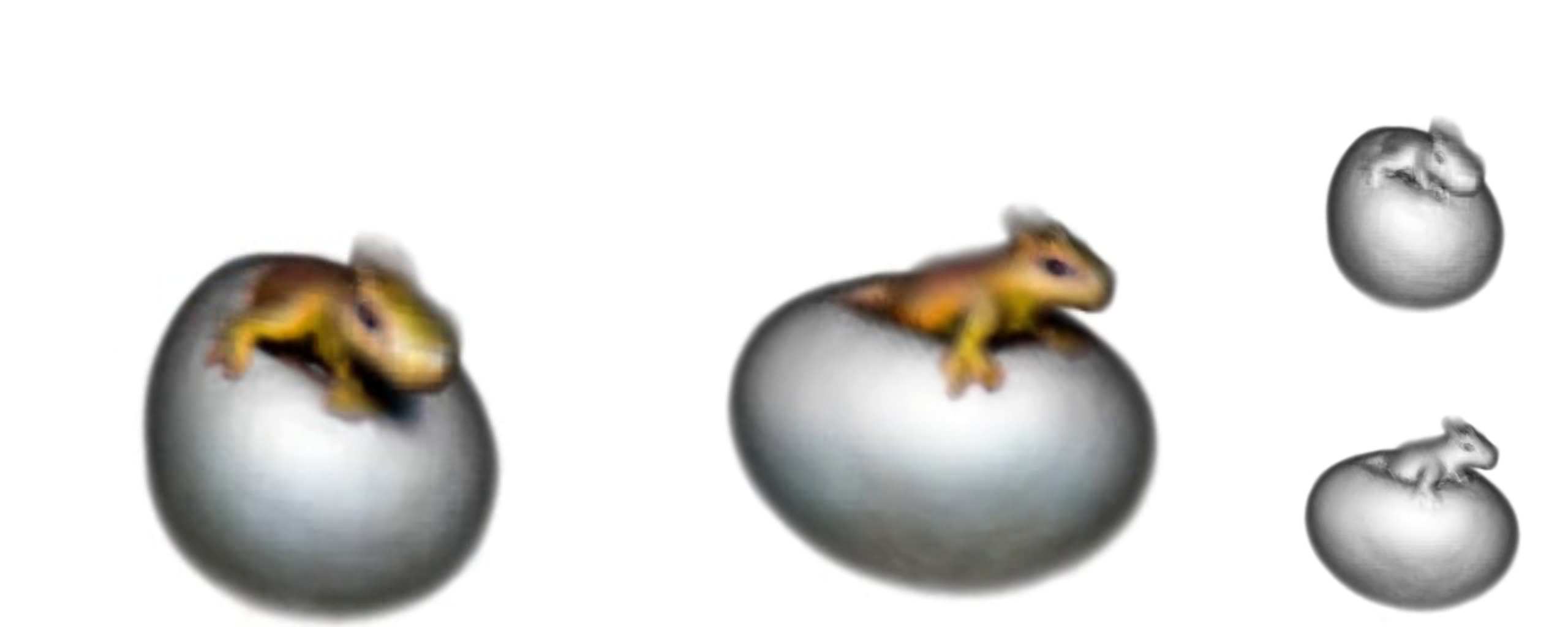}&
        \includegraphics[align=c, width=0.23\linewidth,trim=0 50 0 50,clip]{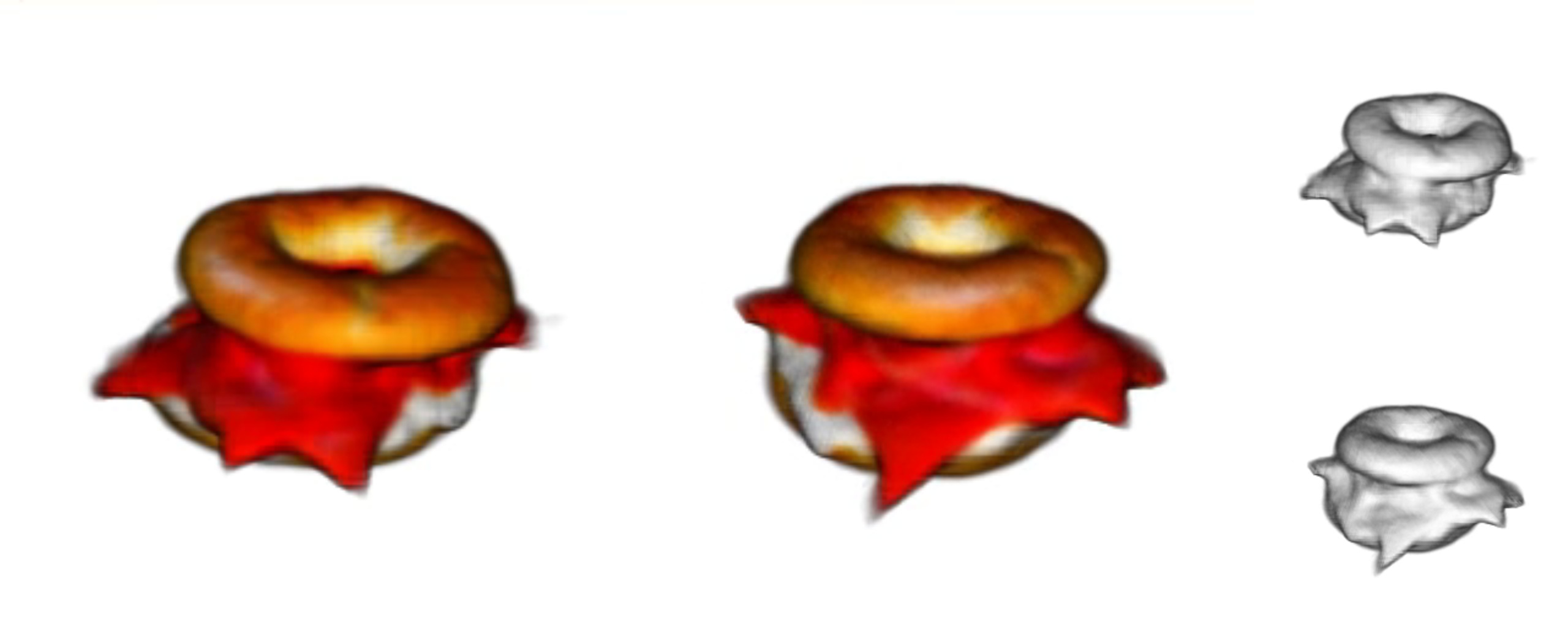} &
        \includegraphics[align=c, width=0.23\linewidth,trim=0 0 0 0,clip]{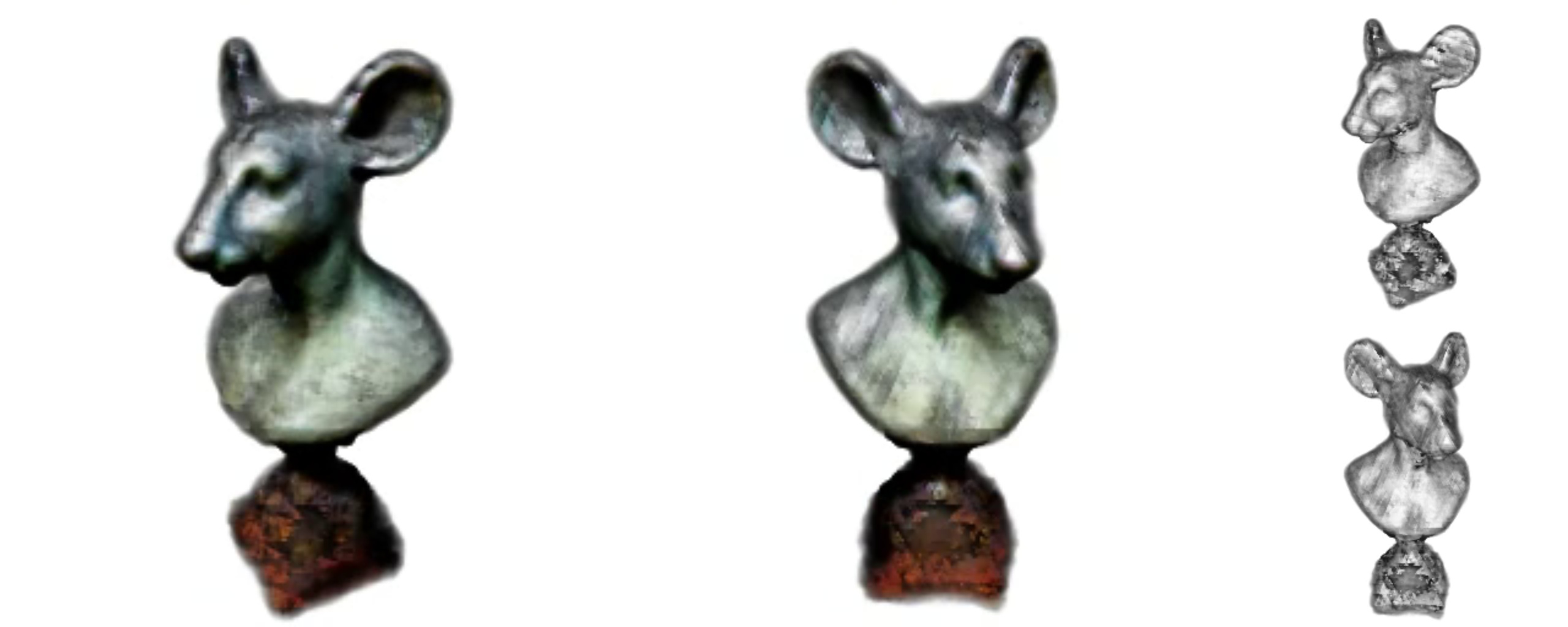}&
        \includegraphics[align=c, width=0.23\linewidth,trim=0 0 0 100,clip]{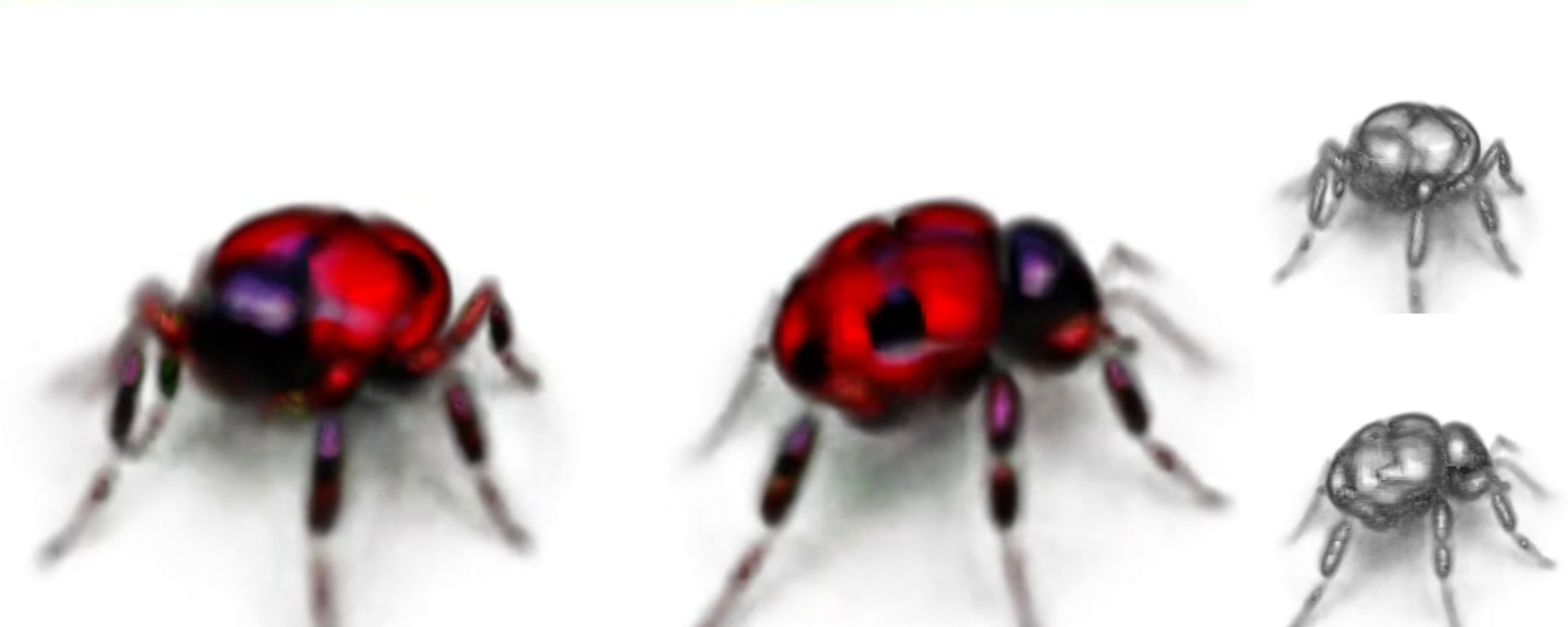} \\
         \makecell{\scriptsize Fine-tuned \\ \scriptsize  NeRF}& \includegraphics[align=c, width=0.23\linewidth,trim=0 0 0 100,clip]{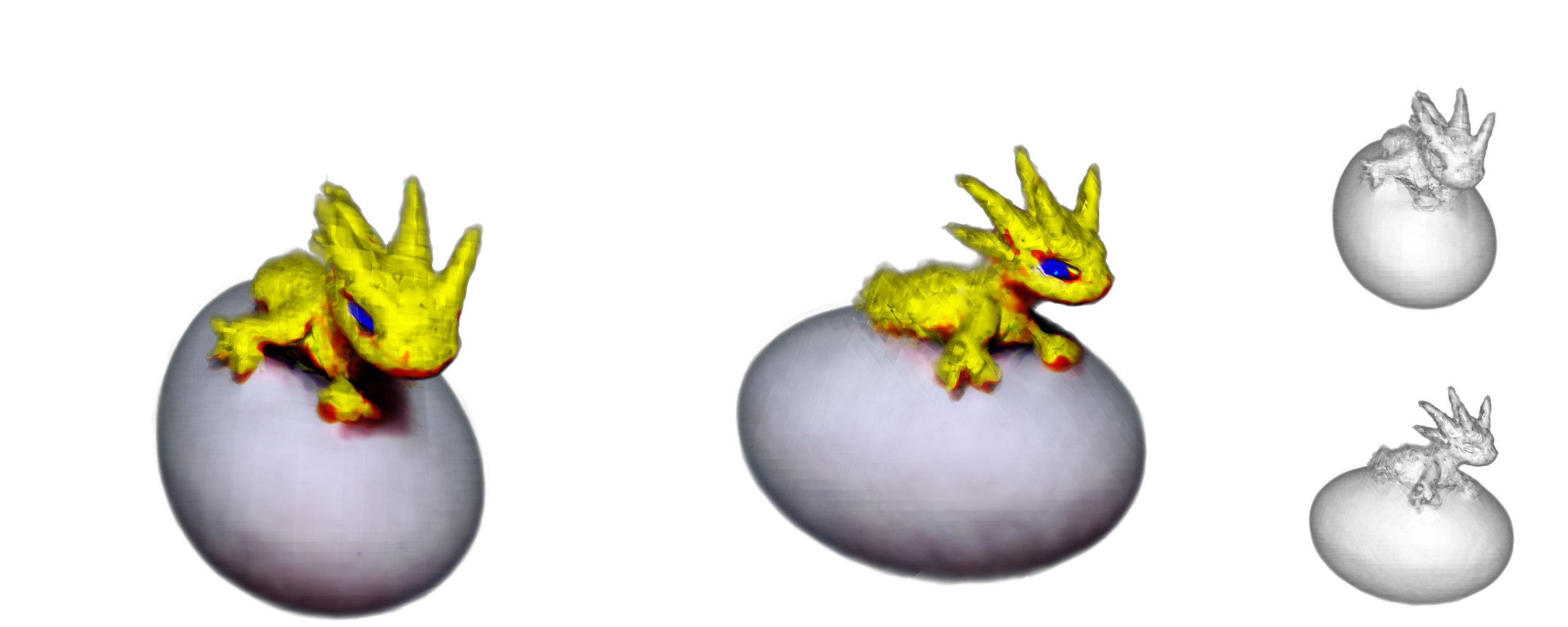}&
        \includegraphics[align=c, width=0.23\linewidth,trim=0 50 0 50,clip]{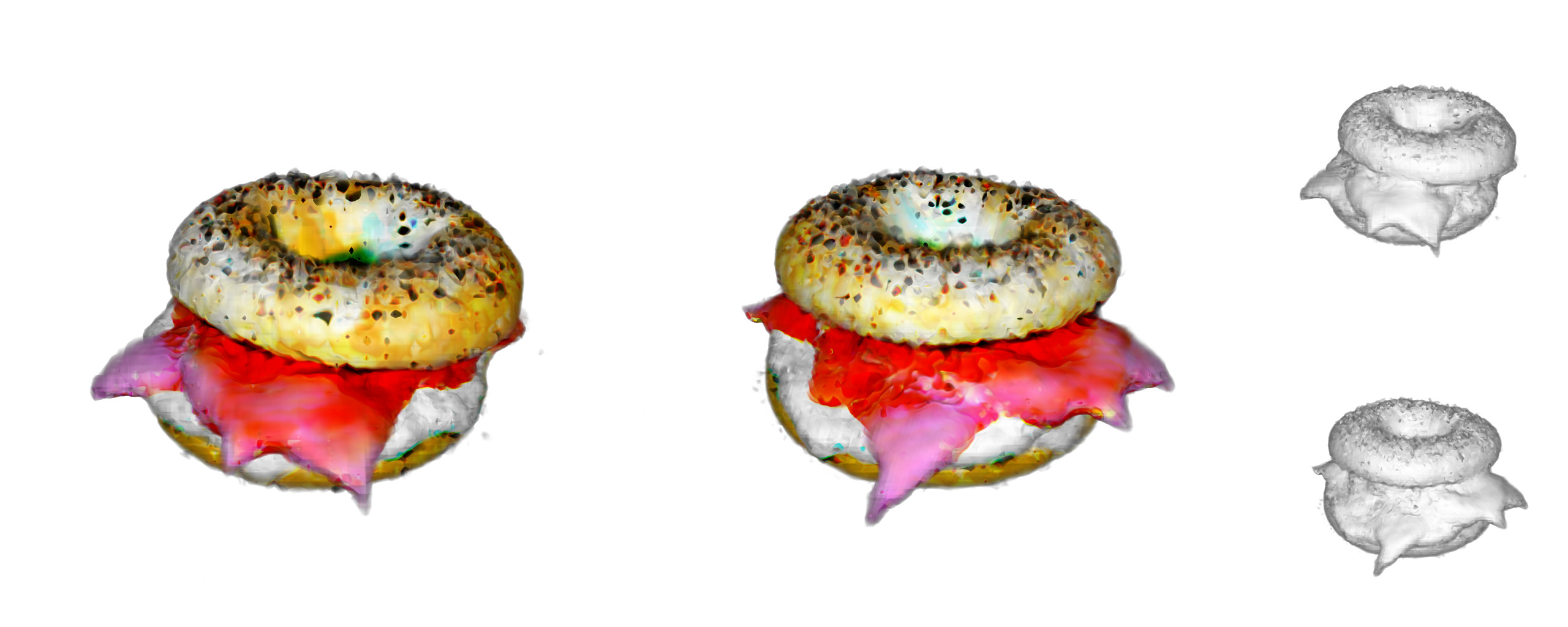} &
        \includegraphics[align=c, width=0.23\linewidth,trim=0 0 0 0,clip]{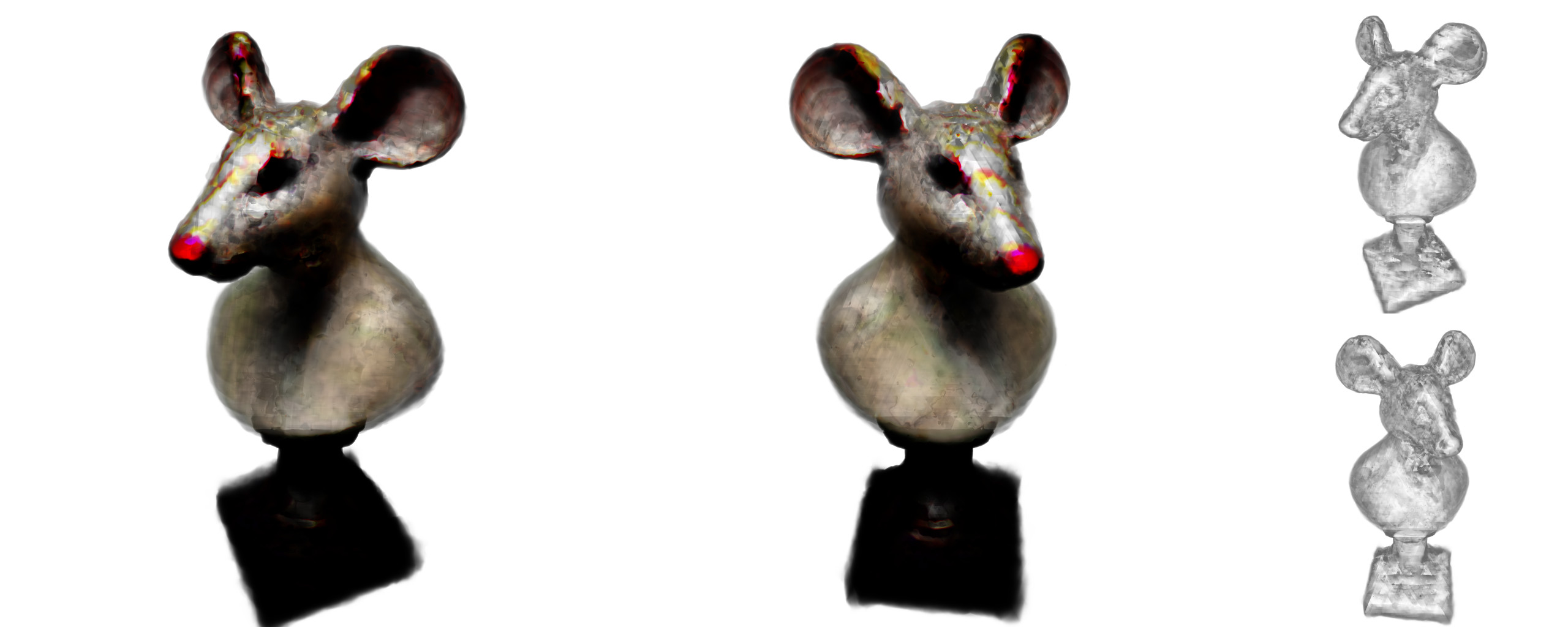}&
        \includegraphics[align=c, width=0.23\linewidth,trim=0 0 0 100,clip]{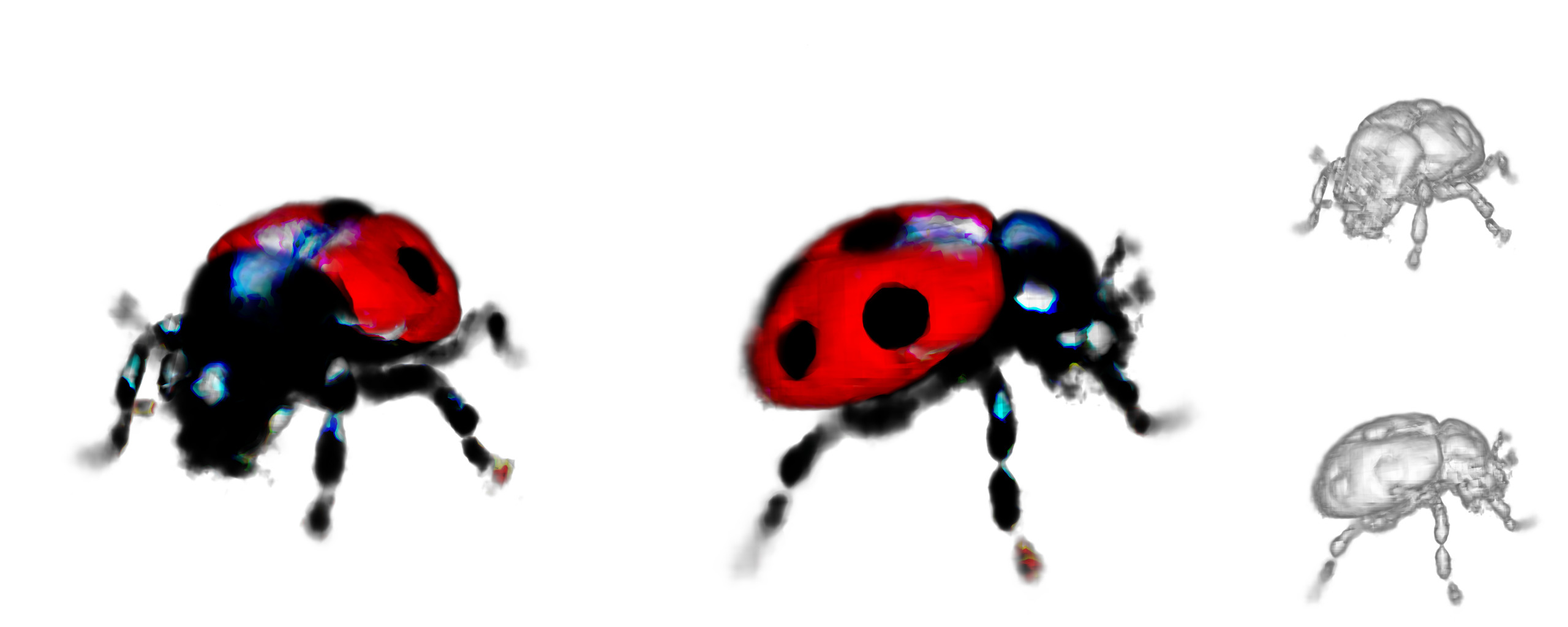} \\
        \makecell{\scriptsize Fine-tuned \\ \scriptsize Mesh}& \includegraphics[align=c, width=0.23\linewidth,trim=0 0 0 100,clip]{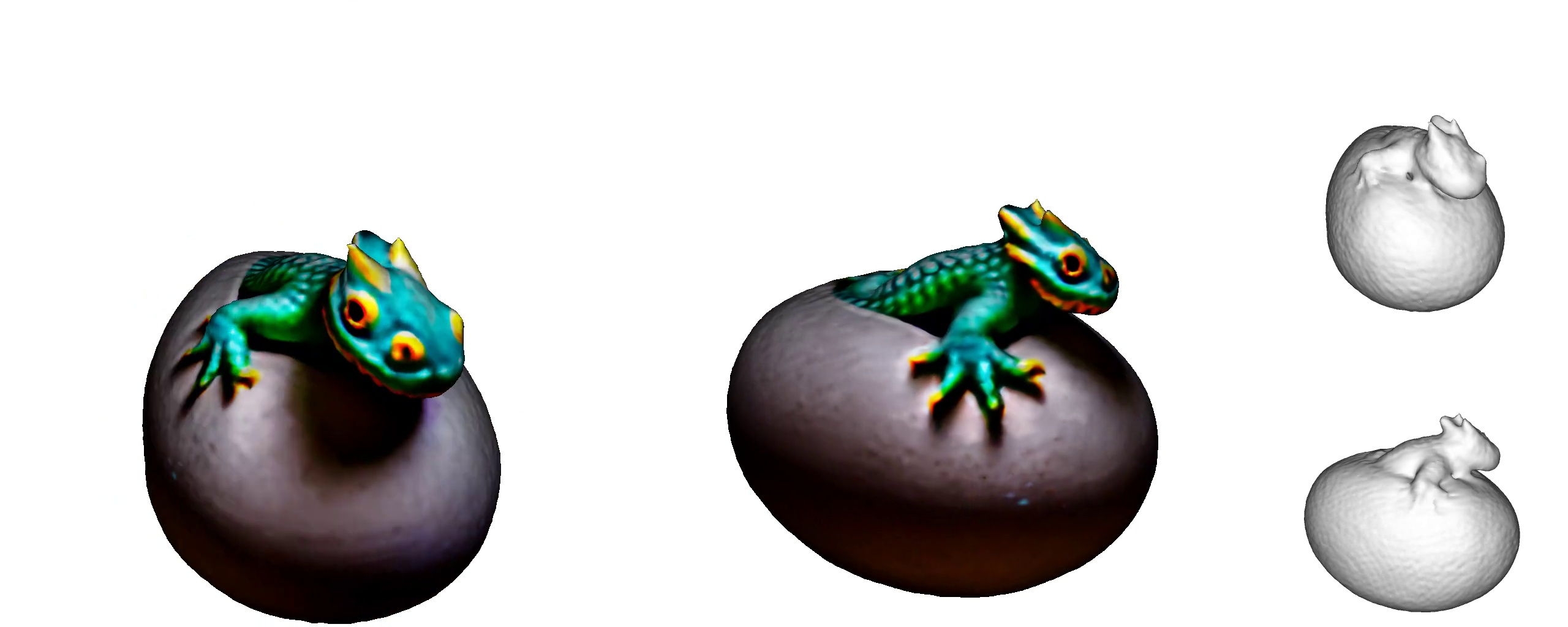}&
        \includegraphics[align=c, width=0.23\linewidth,trim=0 50 0 50,clip]{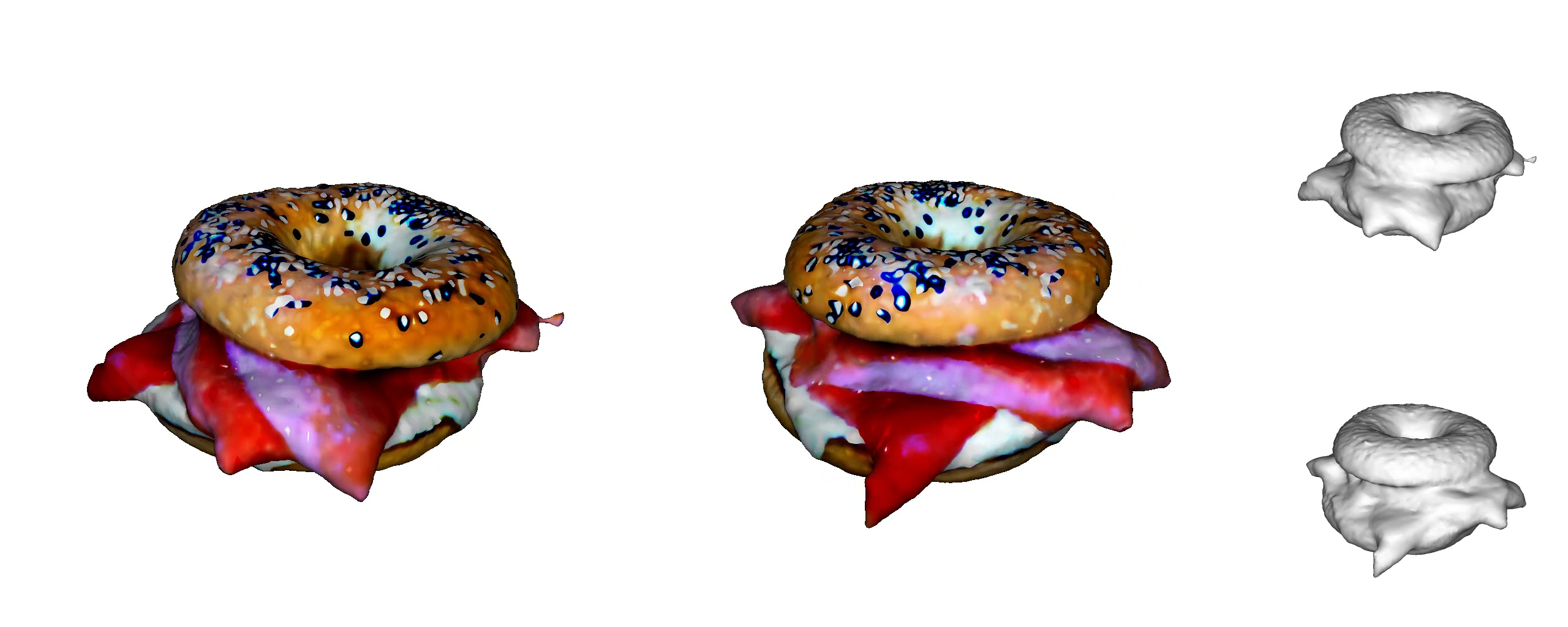} &
        \includegraphics[align=c, width=0.23\linewidth,trim=0 0 0 0,clip]{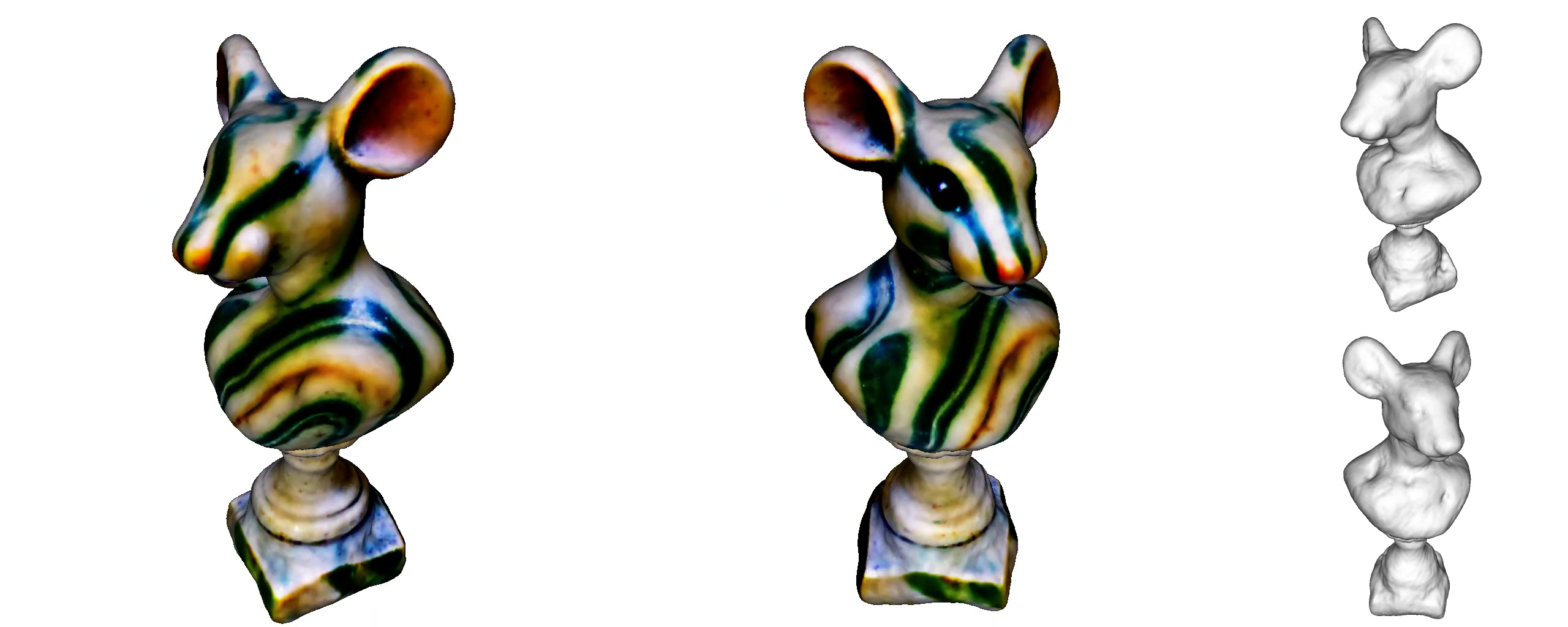}&
        \includegraphics[align=c, width=0.23\linewidth,trim=0 0 0 100,clip]{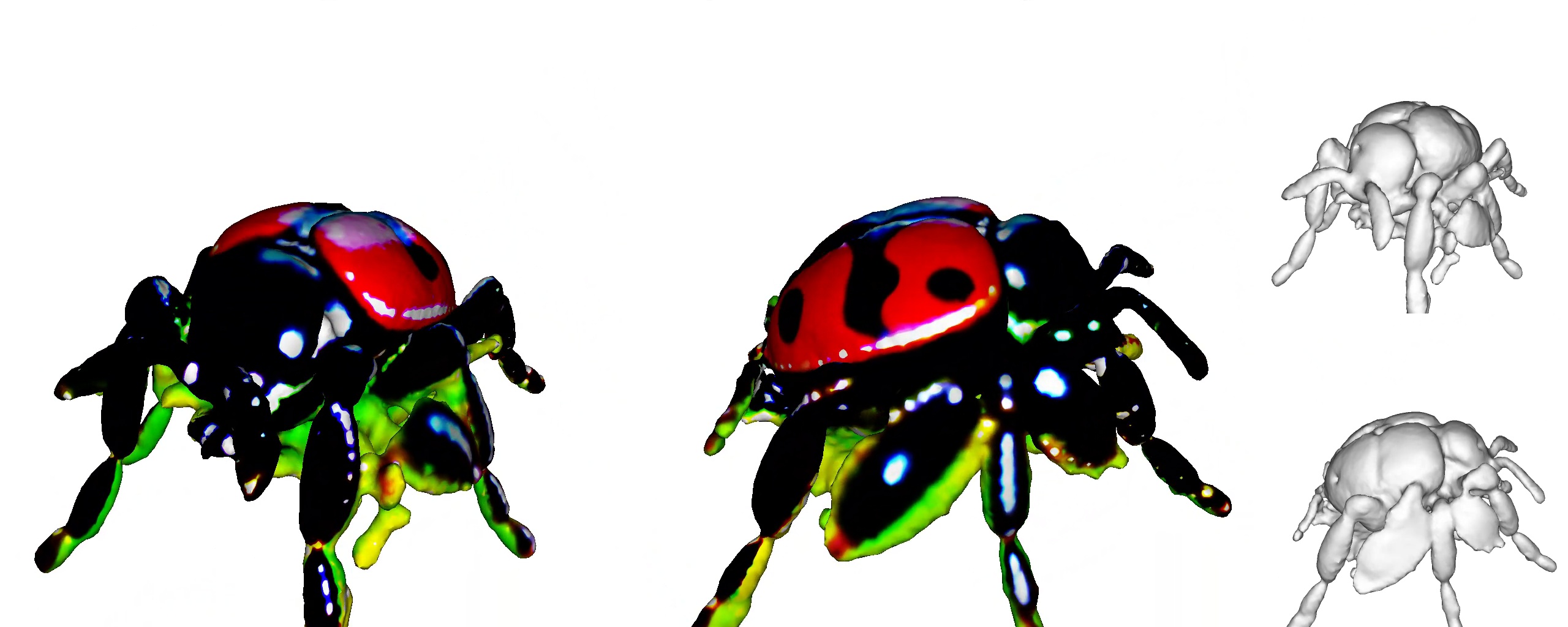} \\
         &{\scriptsize \textit{baby dragon hatching out of a stone egg$^{\ast}$}} & 
        {\scriptsize \textit{ bagel filled with cream cheese and lox$^{\ast}$}}& {\scriptsize \textit{a marble bust of a mouse}} & {\scriptsize \textit{a ladybug$^{\dagger}$}}  
        \end{tabular}
    \caption{\textbf{Ablation on the fine-tuning stage.}
    For each text prompt, we compare coarse and fine models with mesh and NeRF representations.
    Mesh fine-tuning significantly improve the visual quality of generated 3D assets, providing more photo-realistic details on the 3D shapes.}
    \label{fig:ablate_mesh_two_stage_nerf}
\end{figure*}

\vspace{4pt}
\noindent\textbf{Textured meshes as fine scene models.}
In our fine stage of optimization, we need to be able to accommodate very high-resolution rendered images to fine-tune our scene model with high-resolution diffusion priors.
Using the same scene representation (the neural field) from the initial coarse stage of optimization could be a natural choice since the weights of the model can directly carry over.
Although this strategy can work to some extent (Figs.~\ref{fig:scratch_vs_2stage} and \ref{fig:ablate_mesh_two_stage_nerf}), it struggles to render very high-resolution (\eg, $512 \times 512$) images within reasonable memory constraints and computation budgets. 

To resolve this issue, we use textured 3D meshes as the scene representation for the fine stage of optimization.
In contrast to volume rendering for neural fields, rendering textured meshes with differentiable rasterization can be performed efficiently at very high resolutions, making meshes a suitable choice for our high-resolution optimization stage.
Using the neural field from the coarse stage as the initialization for the mesh geometry, we can also sidestep the difficulty of learning large topological changes in meshes.

Formally, we represent the 3D shape using a deformable tetrahedral grid ($V_T, T$), where $V_T$ is the vertices in the grid $T$~\cite{gao2020deftet,dmtet}.
Each vertex $\mathbf{v}_{i} \in V_T \subset \mathbb{R}^3$ contains a signed distance field (SDF) value $s_i \in \mathbb{R}$ and a deformation $\Delta \mathbf{v}_{i} \in \mathbb{R}^3$ of the vertex from its initial canonical coordinate.
Then, we extract a surface mesh from the SDF using a differentiable marching tetrahedra algorithm~\cite{dmtet}.
For textures, we use the neural color field as a volumetric texture representation.

\subsection{Coarse-to-fine Optimization}
\label{sec:c2fgen}

We describe our coarse-to-fine optimization procedure, which first operates on a coarse neural field representation and subsequently a high-resolution textured mesh. 

\vspace{4pt}
\noindent\textbf{Neural field optimization.} 
Similarly to Instant NGP~\cite{muller2022instant}, we initialize an occupancy grid of resolution $256^3$ with values to 20 to encourage shapes to grow in the early stages of optimization.
We update the grid every 10 iterations and generate an octree for empty space skipping.
We decay the occupancy grid by 0.6 in every update and follow Instant NGP with the same update and thresholding parameters.

Instead of estimating normals from density differences, we use an MLP to predict the normals.
Note that this does not violate geometric properties since volume rendering is used instead of surface rendering; as such, the orientation of particles at continuous positions need not be oriented to the level set surface.
This helps us significantly reduce the computational cost of optimizing the coarse model by avoiding the use of finite differencing.
Accurate normals can be obtained in the fine stage of optimization when we use a true surface rendering model.

Similar to DreamFusion, we also model the background using an environment map MLP, which predicts RGB colors as a function of ray directions.
Since our sparse representation model does not support scene reparametrization as in Mip-NeRF 360~\cite{barron2022mip},
the optimization has a tendency to ``cheat'' by learning the essence of the object using the background environment map.
As such, we use a tiny MLP for the environment map (hidden dimension size of 16) and weigh down the learning rate by $10\times$ to allow the model to focus more on the neural field geometry.

\vspace{4pt}
\noindent\textbf{Mesh optimization.}
To optimize a mesh from the neural field initialization, we convert the (coarse) density field to an SDF by subtracting it with a non-zero constant, yielding the initial $s_i$.
We additionally initialize the volume texture field directly with the color field optimized from the coarse stage.

During optimization, we render the extracted surface mesh into high-resolution images using a differentiable rasterizer~\cite{nvdiffrast,nvdiffrec}.
We optimize both $s_i$ and $\Delta \mathbf{v}_{i}$ for each vertex $\mathbf{v}_{i}$ via backpropagation using the high-resolution SDS gradient (Eq.~\ref{eq:sds_high}).
When rendering the mesh to an image, we also track the 3D coordinates of each corresponding pixel projection, which would be used to query colors in the corresponding texture field for joint optimization.

When rendering the mesh, we increase the focal length to zoom in on object details, which is a critical step towards recovering high-frequency details. We keep the same pre-trained environment map from the coarse stage of optimization and composite the rendered background with the rendered foreground object using differentiable antialiasing~\cite{nvdiffrast}.
To encourage the smoothness of the surface, we further regularize the angular differences between adjacent faces on the mesh.
This allows us to obtain well-behaved geometry even under supervision signals with high variance, such as the SDS gradient $\nabla_\theta \mathcal{L}_\text{SDS}$.

\begin{figure*}
    \centering 
    \setlength{\tabcolsep}{0.2pt}
    \begin{tabular}{cc} 
      \begin{tabularx}{0.5\textwidth}{ >{\centering\arraybackslash}X >{\centering\arraybackslash}X}
        Ours & DreamFusion~\cite{poole2022dreamfusion}
    \end{tabularx}& 
    \begin{tabularx}{0.5\textwidth}{ >{\centering\arraybackslash}X >{\centering\arraybackslash}X}
        Ours & DreamFusion~\cite{poole2022dreamfusion}
    \end{tabularx}\\
         \includegraphics[align=c, width=0.49\linewidth]{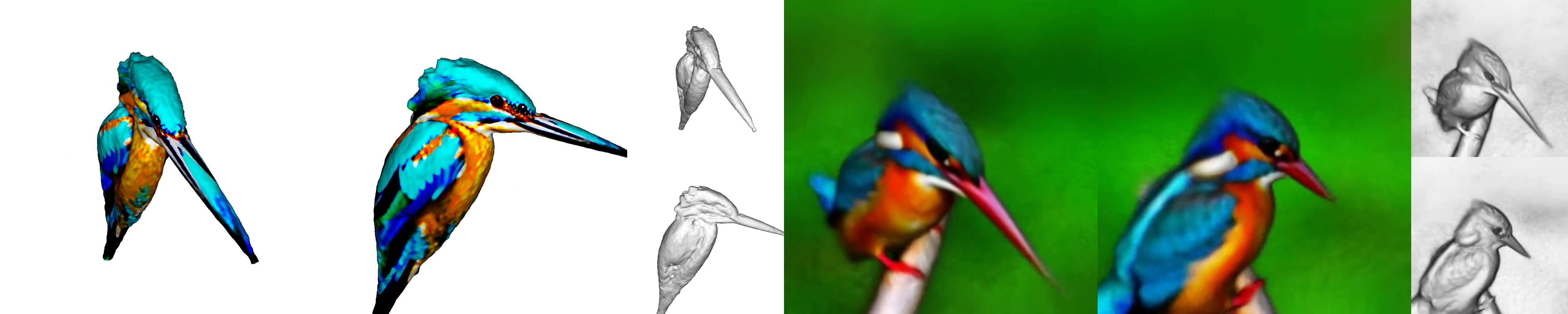}&
        \includegraphics[align=c, width=0.49\linewidth]{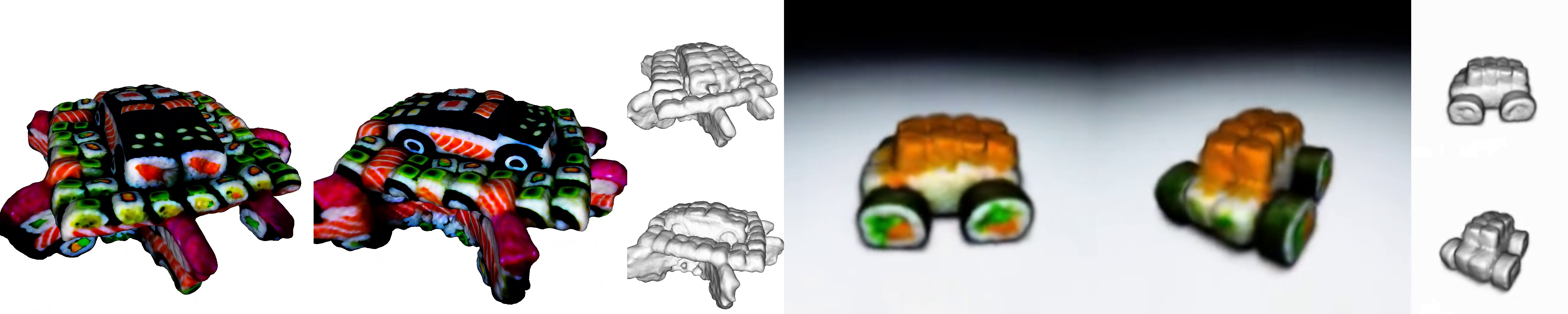} \\
        {\footnotesize \textit{a kingfisher bird$^{\dagger}$}} & 
        {\footnotesize \textit{car made out of sushi$^{\ast}$}}\\
        \includegraphics[align=c, width=0.49\linewidth]{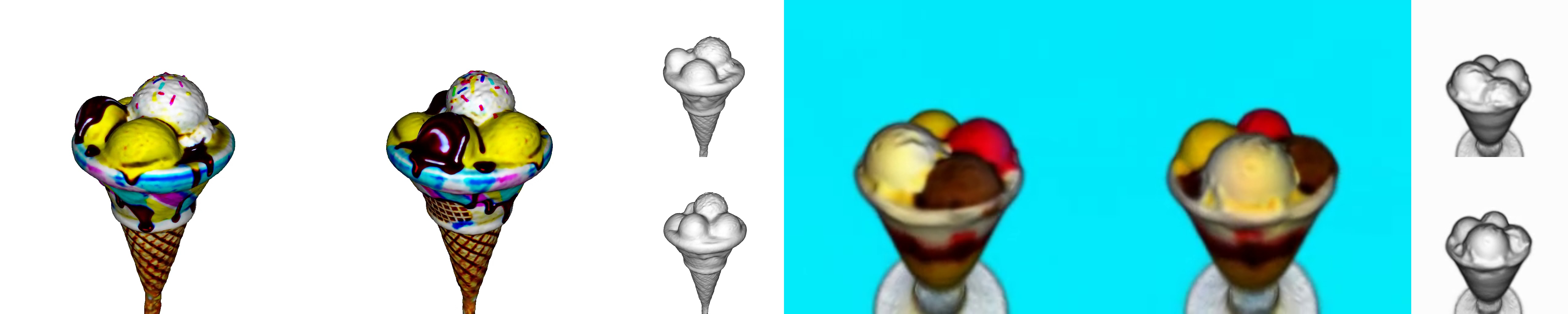} &\includegraphics[align=c, width=0.49\linewidth]{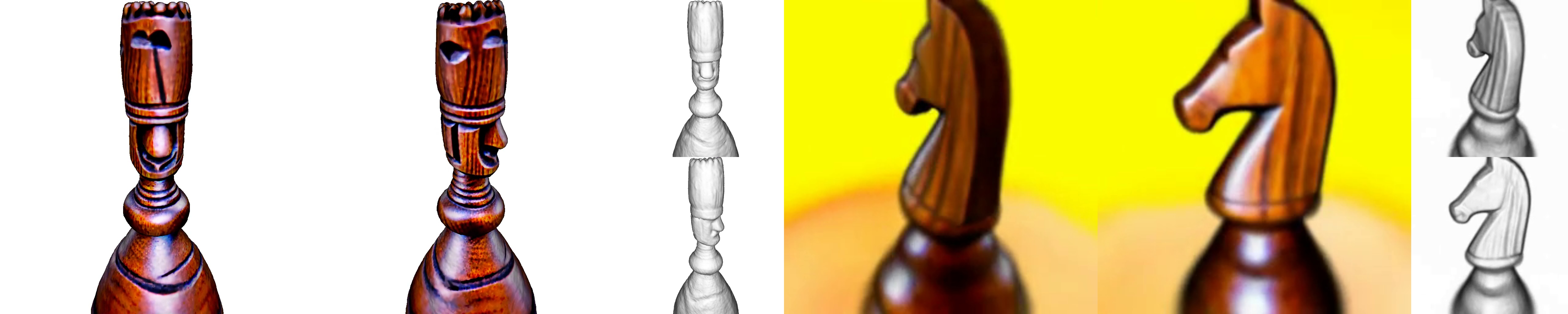}
       \\
        {\footnotesize \textit{ an ice cream sundae$^{\ast}$}} & {\footnotesize \textit{a beautifully carved wooden knight chess piece$^{\dagger}$}}  \\
        \includegraphics[align=c, width=0.49\linewidth]{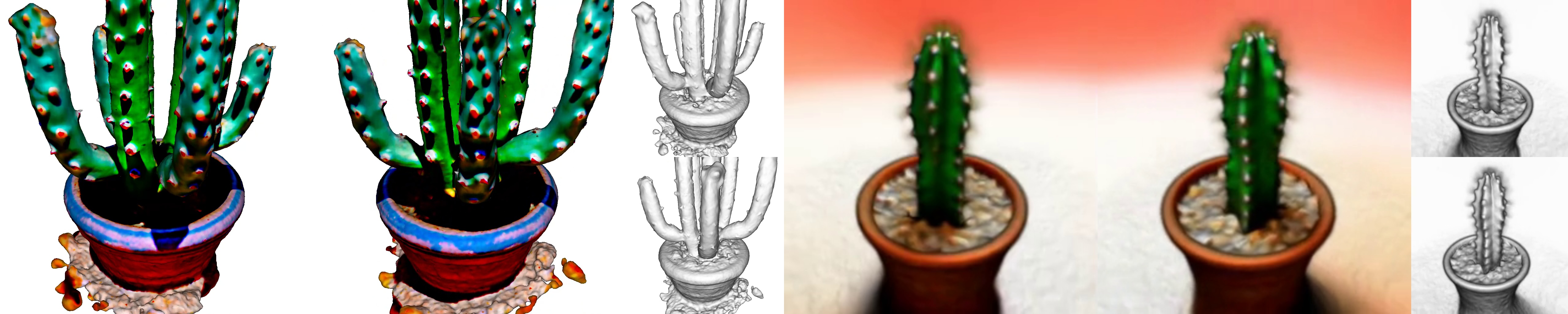} &
        \includegraphics[align=c, width=0.49\linewidth]{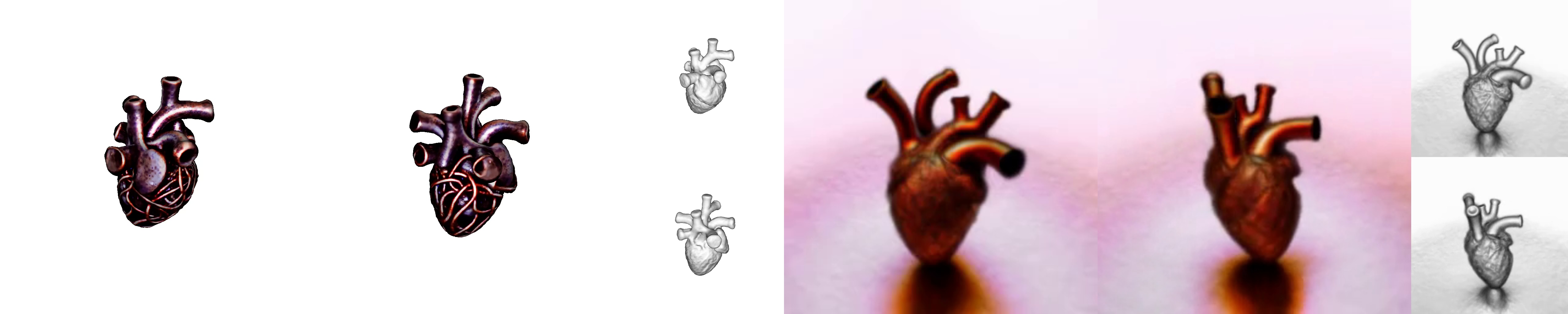} \\
         {\footnotesize \textit{a small saguaro cactus planted in a clay pot$^{\ast}$}} & {\begin{tabular}[c]{@{}c@{}}\footnotesize \textit{A very beautiful tiny human heart organic sculpture made of copper wire }
         \vspace{-1mm}
         \\ \footnotesize \textit{and threaded pipes, very intricate, curved, Studio lighting, high resolution$^{\ast}$}\end{tabular}} \\
       
       \includegraphics[align=c, width=0.49\linewidth]{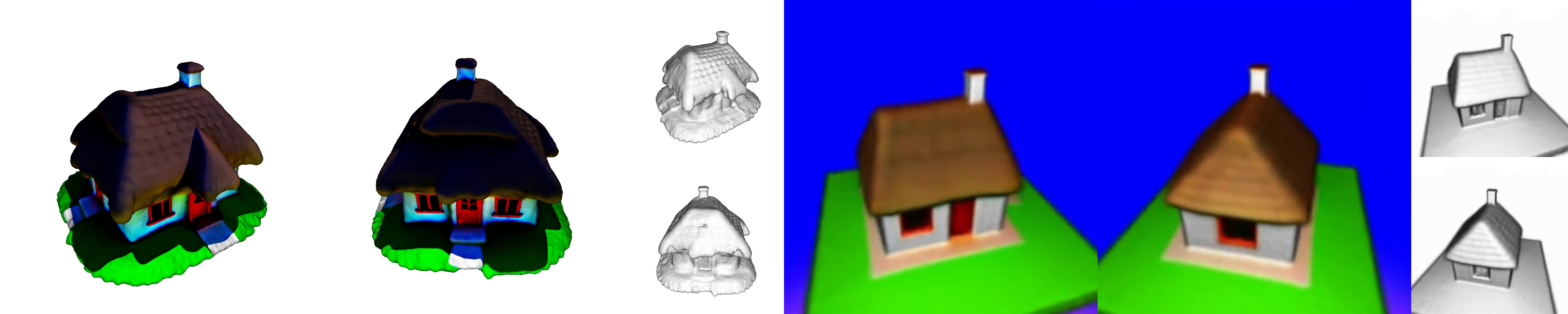}&
        \includegraphics[align=c, width=0.49\linewidth]{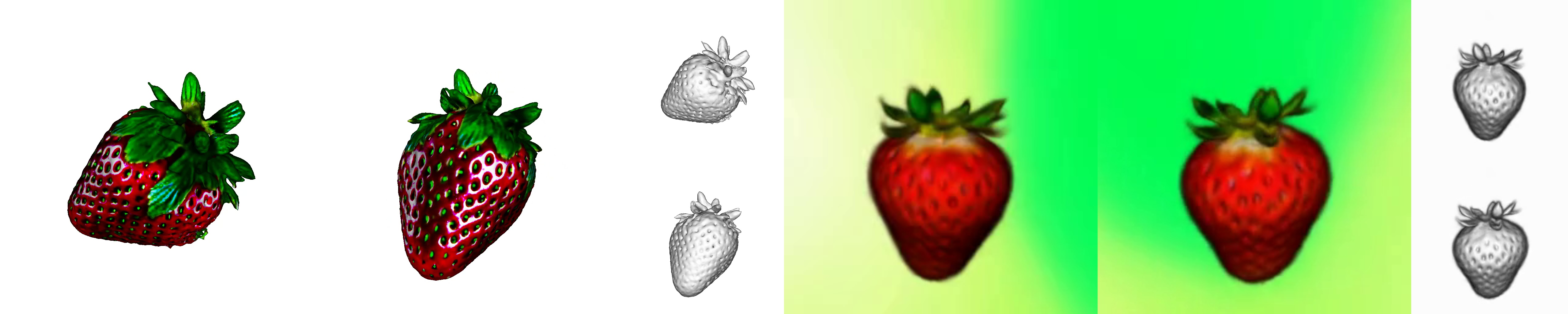} \\
        {\footnotesize \textit{a 3D model of an adorable cottage with a thatched roof$^{\dagger}$}}& {\footnotesize \textit{a ripe strawberry}}
        \end{tabular}
    \caption{\textbf{Qualitative comparison with DreamFusion~\cite{poole2022dreamfusion}.} We use the same text prompt as in DreamFusion. For each 3D model, we render it from two views with a textureless rendering for each view and remove the background to focus on the actual 3D shape. For the DreamFusion results, we take frames from the videos published on the official webpage. Our \name generates much higher quality 3D shapes on both geometry and texture compared with DreamFusion. 
\textit{$\ast$ a DSLR photo of... $\dagger$ a zoomed out DSLR photo of...} }
    \label{fig:compare_dreamfusion}
\end{figure*}
\begin{table}
    \renewcommand{\arraystretch}{1.1}
    \center
    \caption{\textbf{User preference studies.} We conducted user studies to measure preference for 3D models generated using 397 prompts released by DreamFusion.
    Overlal, more raters (61.7\%) prefer 3D models generated by \name over DreamFusion.
    The majority of raters (87.7\%) prefer fine models over coarse models in \name, showing the effectiveness of our coarse-to-fine approach.}
    \rowcolors{2}{gray!10}{white}
    \begin{tabular}{lc}
    \hline
    \multicolumn{1}{c}{\textbf{Comparison}} & \textbf{Preference} \\
    \hline
    \name \vs DreamFusion~\cite{poole2022dreamfusion}  & \\
    $\;\;\;\bullet\;$ More realistic & 58.3\% \\
    $\;\;\;\bullet\;$ More detailed & 66.0\% \\
    $\;\;\;\bullet\;$ More realistic \& detailed  & 61.7\% \\
    \name \vs~\name (coarse only) & 87.7\% \\
    \hline
    \end{tabular}
    \label{tbl:user_study}
\end{table}

\section{Experiments}
\label{sec:experiments4}

We focus on comparing our method with DreamFusion~\cite{poole2022dreamfusion} on 397 text prompts taken from the website of DreamFusion\footnote{\url{https://dreamfusion3d.github.io/gallery.html}}.
We train \name on all of the text prompts and compare them with the results provided on the website.

\vspace{4pt}
\noindent\textbf{Speed evaluation.}
Unless otherwise noted, the coarse stage is trained for 5000 iterations with 1024 samples along the ray (subsequently filtered by the sparse octree) with a batch size of 32, with a total runtime of around $15$ minutes (upwards of $8$ iterations / second, variable due to differences in sparsity).
The fine stage is trained for 3000 iterations with a batch size of 32 with a total runtime of $25$ minutes ($2$ iterations / second).
Both runtimes combined are $40$ minutes.
All runtimes were measured on 8 NVIDIA A100 GPUs.

\vspace{4pt}
\noindent\textbf{Qualitative comparisons.}
We provide qualitative examples in Fig.~\ref{fig:compare_dreamfusion}. Qualitatively, our models achieve much higher 3D quality in terms of both geometry and texture. Notice that our model can generate candies on ice cream cones, highly detailed sushi-like cars, vivid strawberries, and birds. We also note that our resulting 3D models can be directly imported and visualized in standard graphics software. 

\vspace{4pt}
\noindent\textbf{User studies.}
We conduct user studies to evaluate different methods based on user preferences on Amazon MTurk.
We show users two videos side by side rendered from a canonical view by two different algorithms using the same text prompt.
We ask the users to select the one that is more realistic and detailed.
Each prompt is evaluated by $3$ different users, resulting in $1191$ pairwise comparisons.
As shown in Table~\ref{tbl:user_study}, users favor 3D models generated by \name, with 61.7\% of the users considering our results with higher quality.

\vspace{4pt}
\noindent\textbf{Can single-stage optimization work with LDM prior?}
We ablate scene models optimized with high-resolution LDM prior in a single-stage optimization setup.
We find that 3D meshes as the scene model fail to generate high-quality results if optimized from scratch.
This leaves our our memory-efficient sparse 3D representation as the ideal candidate for the scene model.
However, rendering $512\times512$ images is still too memory intensive to fit into modern GPUs.
Therefore, we render lower-resolution images from the scene model and upsample them to $512\times 512$ as input to the LDM.  
We find it generates objects with worse shapes.
Fig.~\ref{fig:scratch_vs_2stage} shows two examples with scene rendering resolution $64 \times 64$ and $256 \times 256$ respectively (top row).
While it generates furry details, the shape is worse than the coarse model.

\vspace{4pt}
\noindent\textbf{Can we use NeRF for the fine model?}
Yes. While optimizing a NeRF from scratch does not work well, we can follow the coarse-to-fine framework but replace the second-stage scene model with a NeRF. In the bottom right of Fig.~\ref{fig:scratch_vs_2stage}, we show the result of a fine NeRF model initialized with the coarse model on its left and fine-tuned with $256 \times 256$ rendered images. The two-stage approach retains good geometry in the initial model and adds more details, showing superior quality to its one-stage counterpart.

\vspace{4pt}
\noindent\textbf{Coarse models vs. fine models.}
Fig.~\ref{fig:ablate_mesh_two_stage_nerf} provides more visual results contrasting coarse and fine models. We try both NeRF and mesh for scene models and fine-tune them from the same coarse model above. We see significant quality improvements on both NeRF and mesh models, suggesting our coarse-to-fine approach works for general scene models.

\begin{figure}[t!] \setlength{\tabcolsep}{0.1pt}
    \renewcommand{\arraystretch}{0.7} 

        \begin{tabular}{ccc} 
            \includegraphics[width=0.33\linewidth, height=0.33\linewidth]{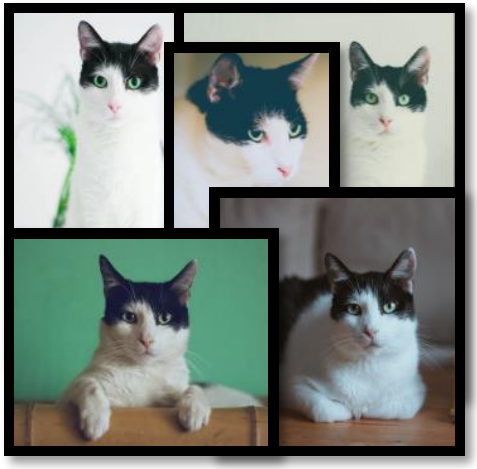} & 
            \includegraphics[width=.33\linewidth, height=.33\linewidth, trim={3cm 2cm 4cm 5cm}, clip]{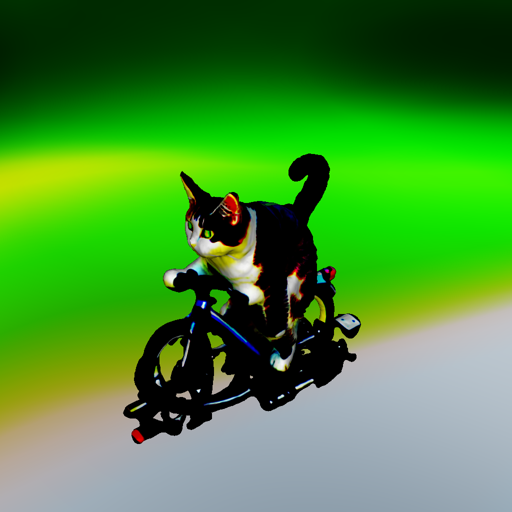} &
            \includegraphics[width=.33\linewidth, height=.33\linewidth, trim={3cm 2cm 4cm 5cm}, clip]{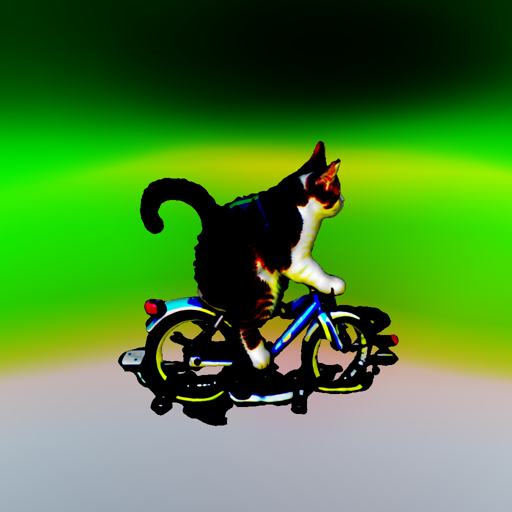} \\
            \footnotesize{Input images} & \multicolumn{2}{c}{\footnotesize \textit{a [V] cat riding a bike*}} \\ 
            \includegraphics[width=0.33\linewidth, height=0.33\linewidth]{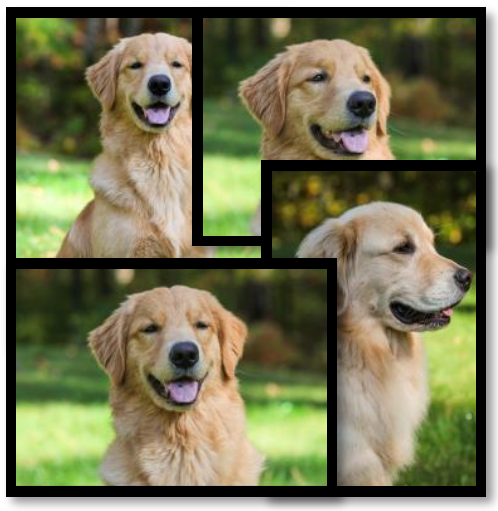} & 
            \includegraphics[width=.33\linewidth, height=.33\linewidth, trim={7cm 7cm 7cm 7cm}, clip]{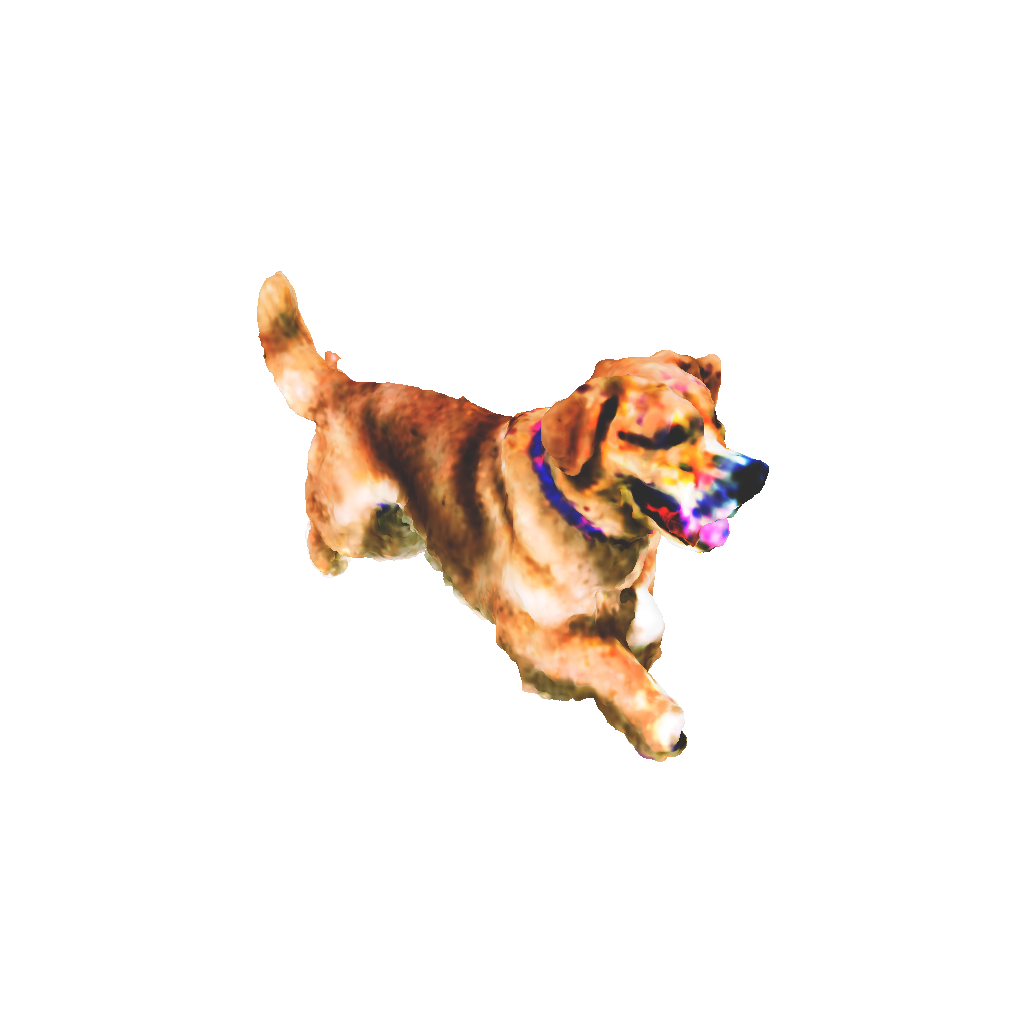} &
            \includegraphics[width=.33\linewidth, height=.33\linewidth, trim={8cm 8cm 8cm 8cm}, clip]{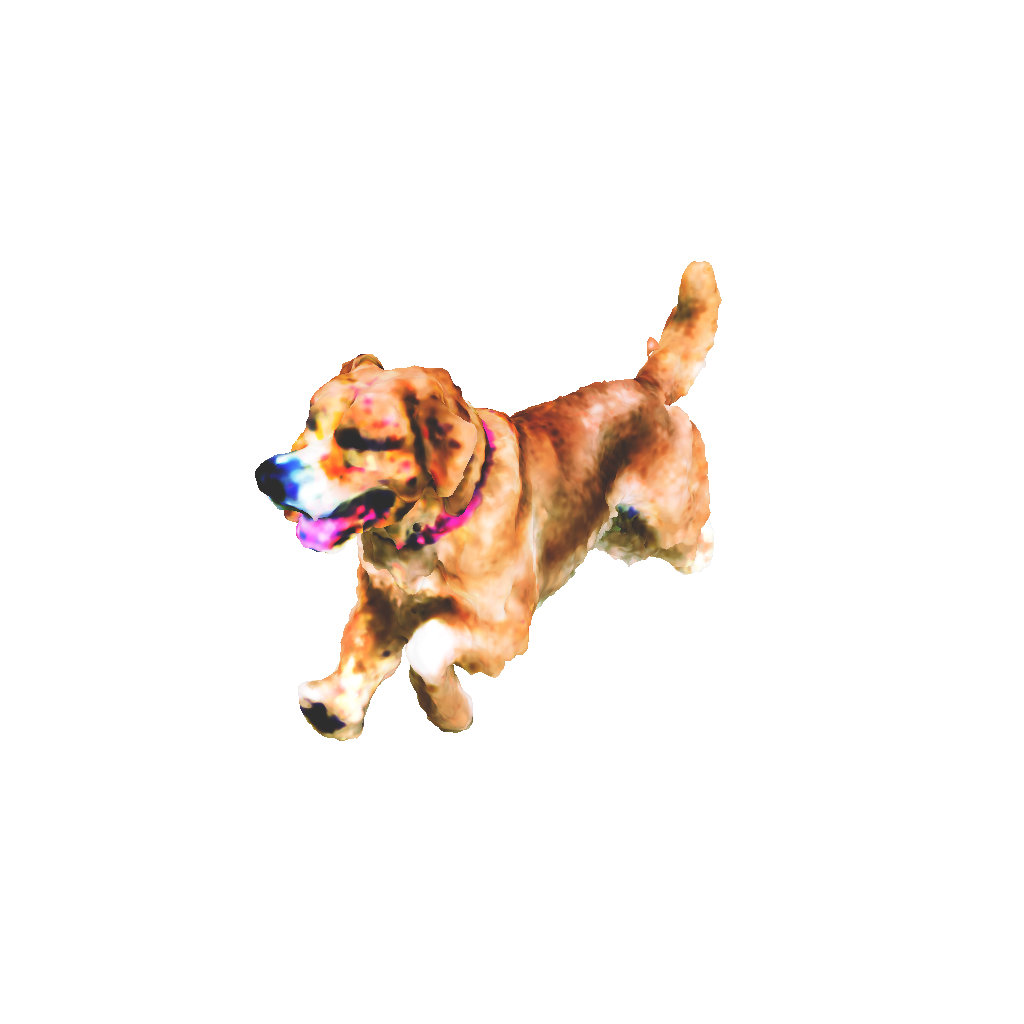} \\
            \footnotesize{Input images} & \multicolumn{2}{c}{\footnotesize \textit{a [V] dog running down the track*}} 
        \end{tabular}

    \caption{ \textbf{\name with DreamBooth-based personalization.}
         Given an input image of a particular instance, we fine-tune the diffusion models with DreamBooth and optimize the 3D models with the given prompts.
         The identity is well preserved in the generated 3D models.
         \textit{Image source (input images): Unsplash}. 
    }
    \label{fig:dreambooth}
\end{figure}

\begin{figure*}[t!] 
    \renewcommand{\arraystretch}{0.6}
    \centering 
    \setlength{\tabcolsep}{0.pt}
    \begin{tabular}{cccc ccc} 
    \includegraphics[align=c, width=0.1\linewidth, trim={0cm 0cm 0.3cm 0cm}, clip]{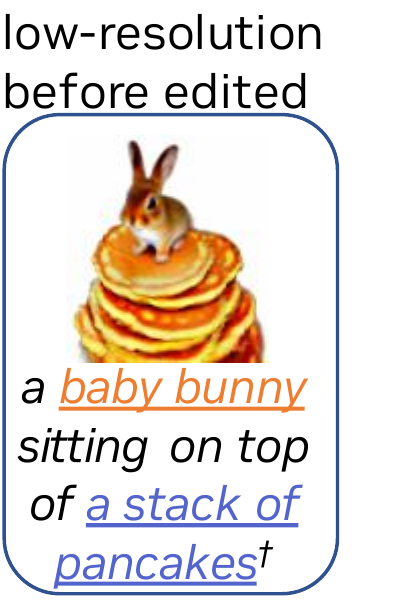} & 
        \includegraphics[align=c, width=0.15\linewidth]{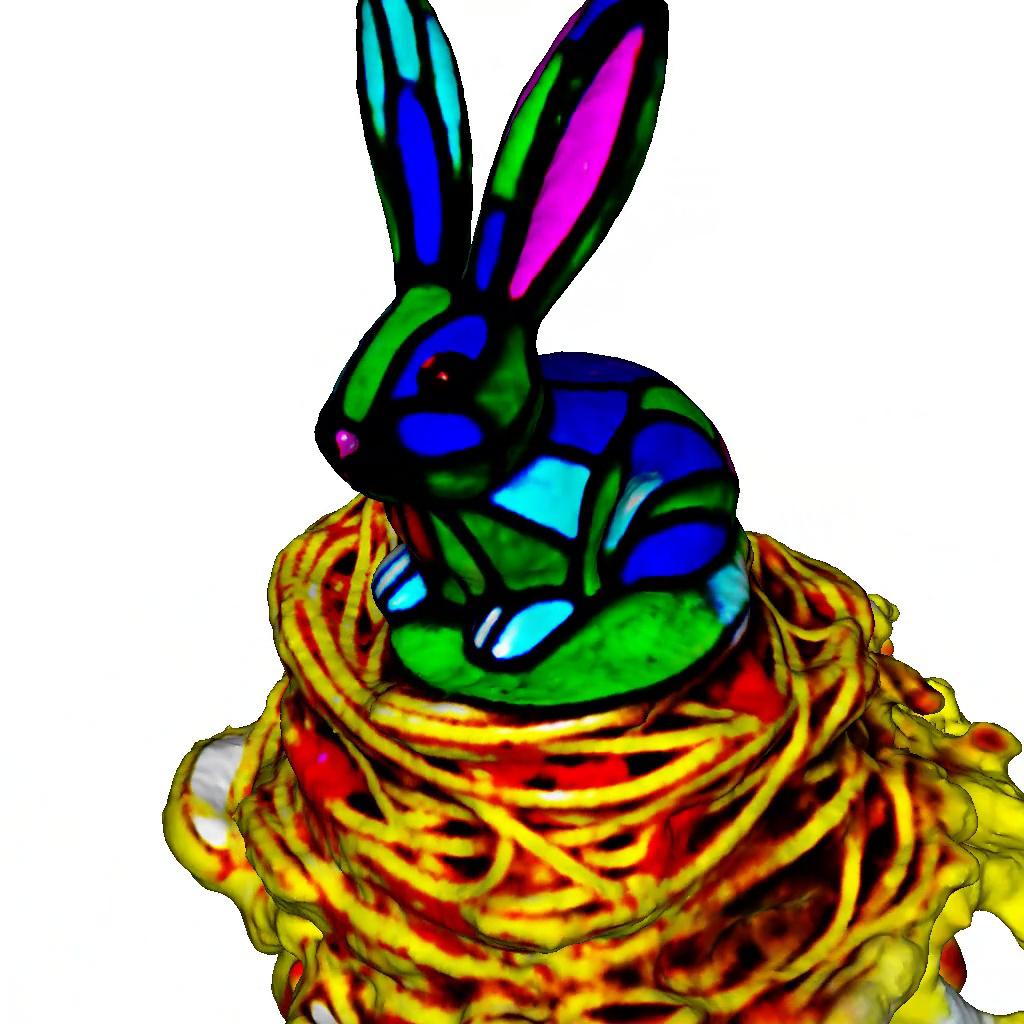} & 
        \includegraphics[align=c, width=0.15\linewidth]{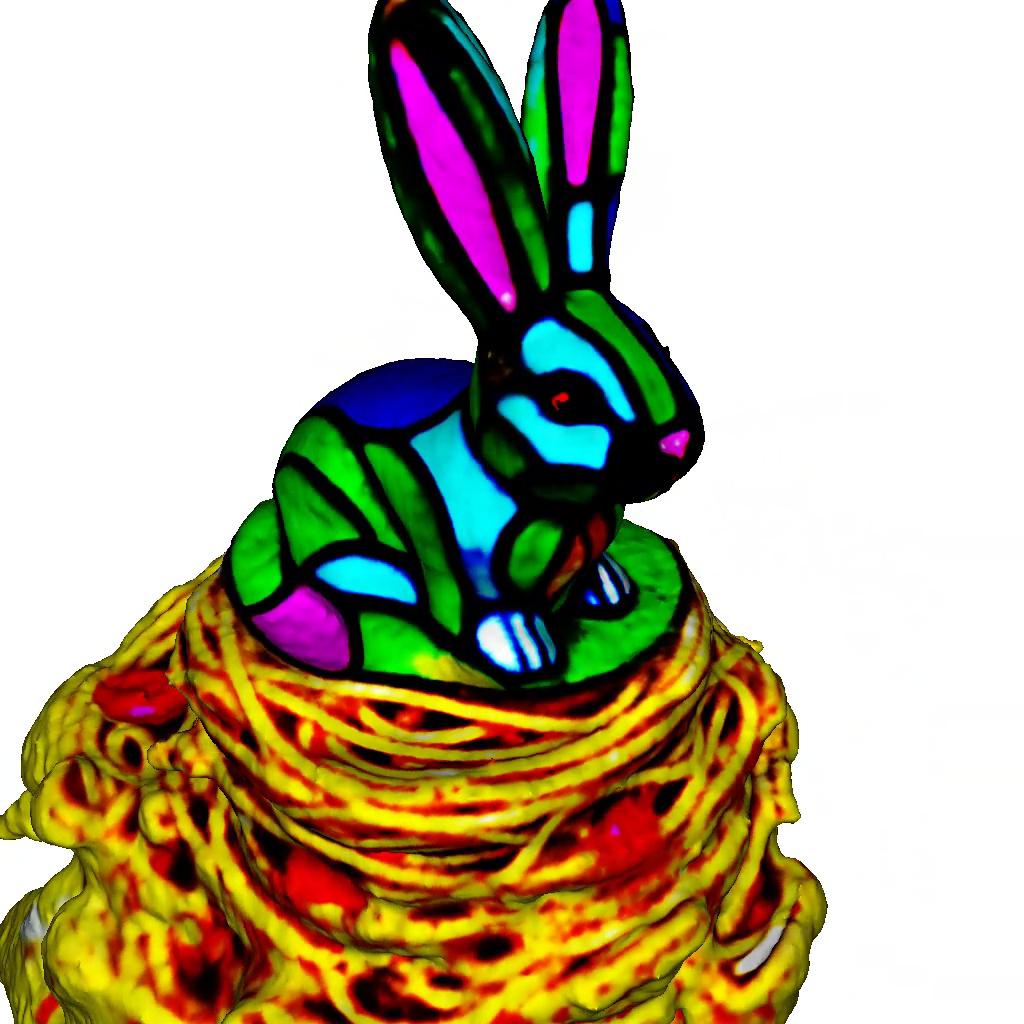} & 
        \includegraphics[align=c, width=0.15\linewidth]{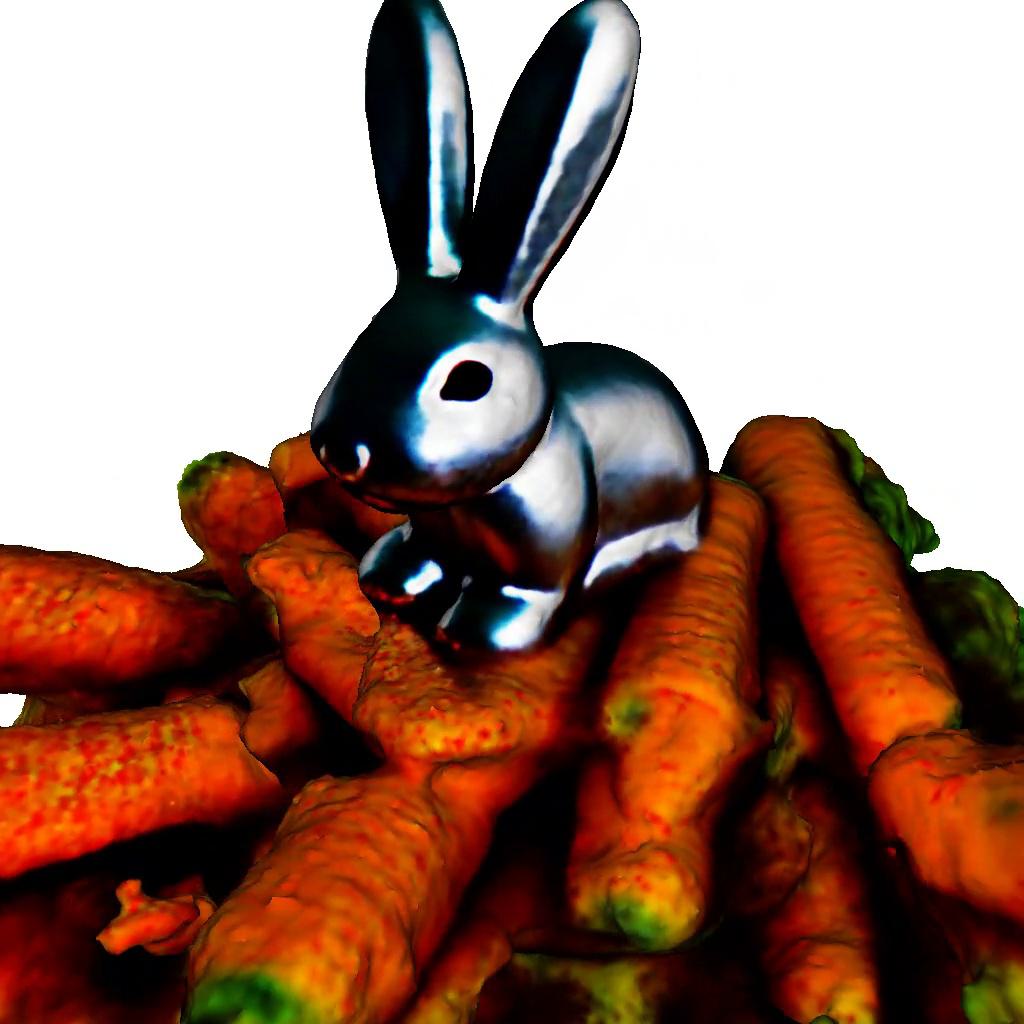} & 
        \includegraphics[align=c, width=0.15\linewidth]{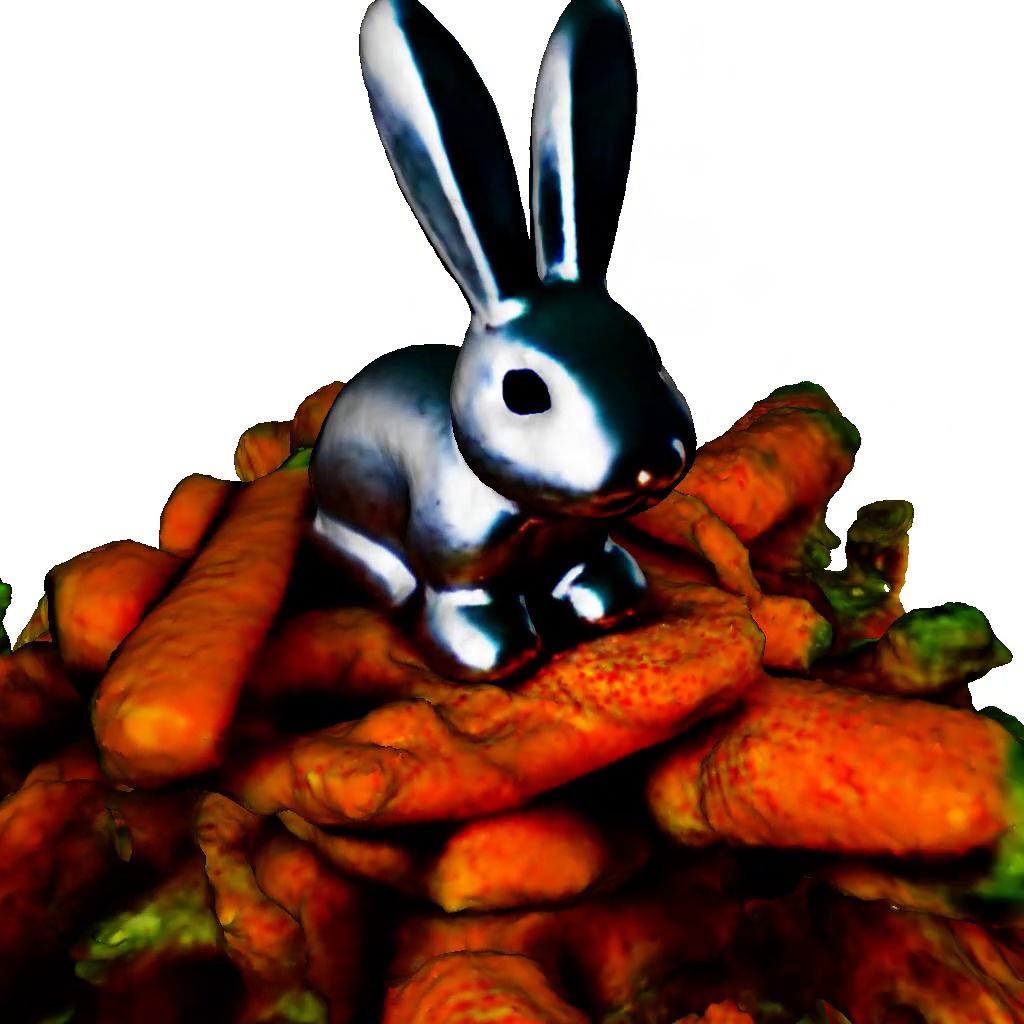} &
        \includegraphics[align=c, width=0.15\linewidth]{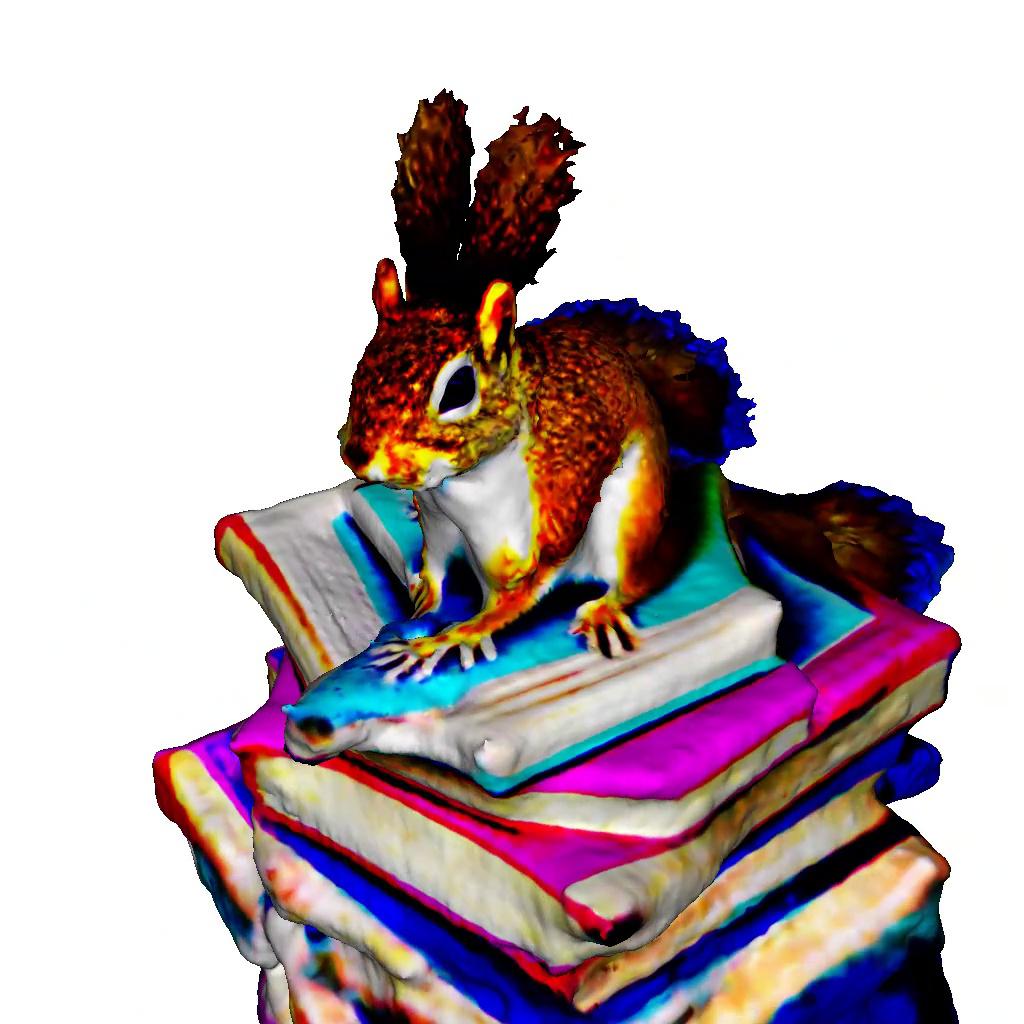} & 
        \includegraphics[align=c, width=0.15\linewidth]{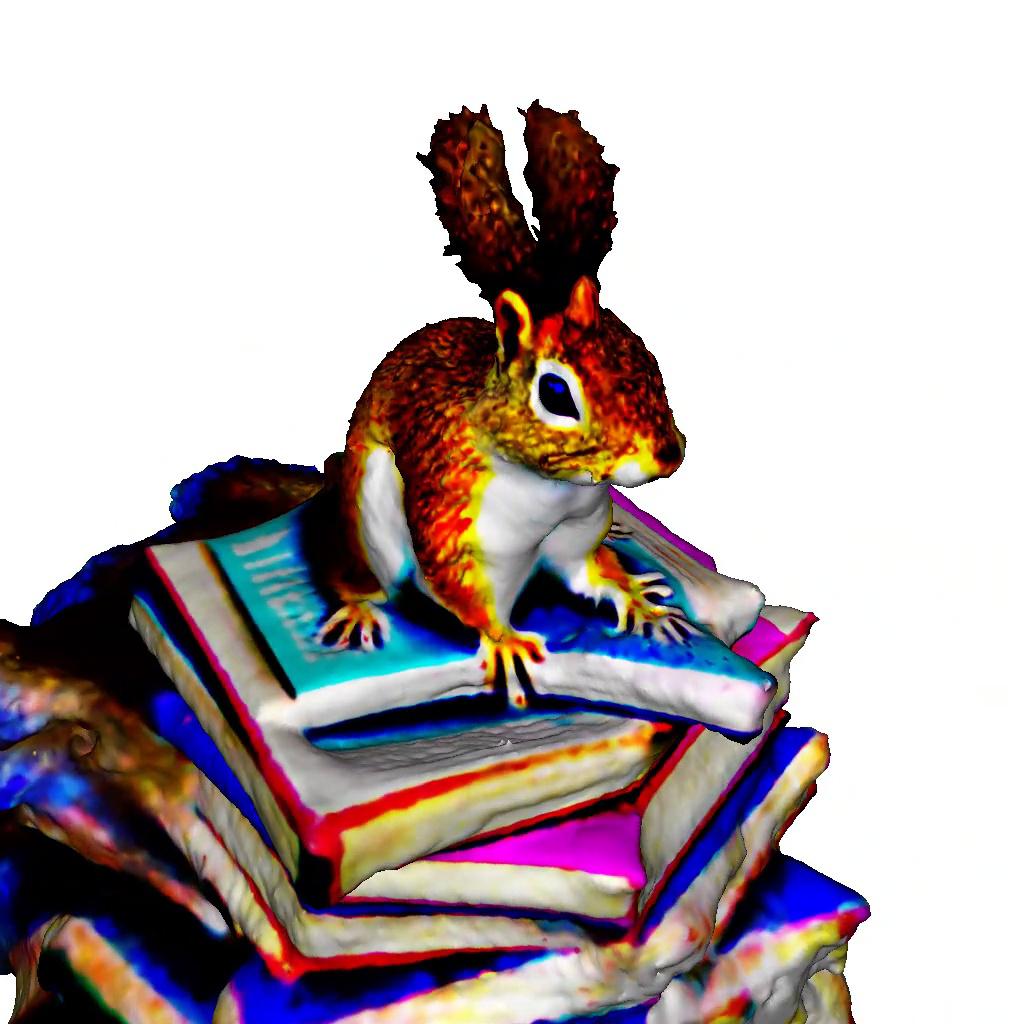} \\        
        & \multicolumn{2}{c}{\footnotesize \textit{\textcolor{myorange}{stained glass bunny}, \textcolor{myblue}{a plate of spaghetti}}} &
        \multicolumn{2}{c}{\footnotesize \textit{\textcolor{myorange}{metal bunny}, \textcolor{myblue}{a stack of carrots}}} &
        \multicolumn{2}{c}{\footnotesize \textit{\textcolor{myorange}{squirrel}, \textcolor{myblue}{a stack of books}}} \\
    \includegraphics[align=c, width=0.1\linewidth, trim={0cm 0cm 0.3cm 0cm}, clip]{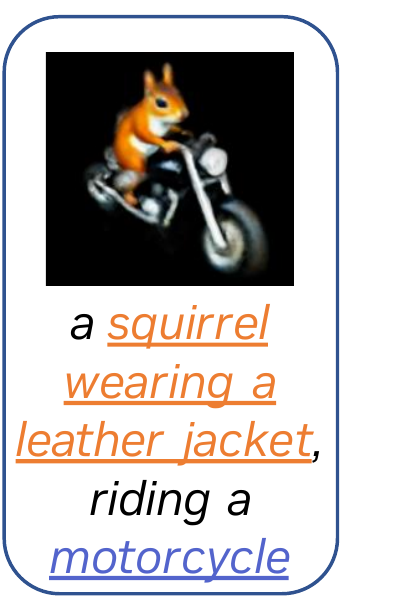}
       & \includegraphics[align=c, width=0.15\linewidth, trim={1.5cm 4cm 6cm 3.5cm}, clip]{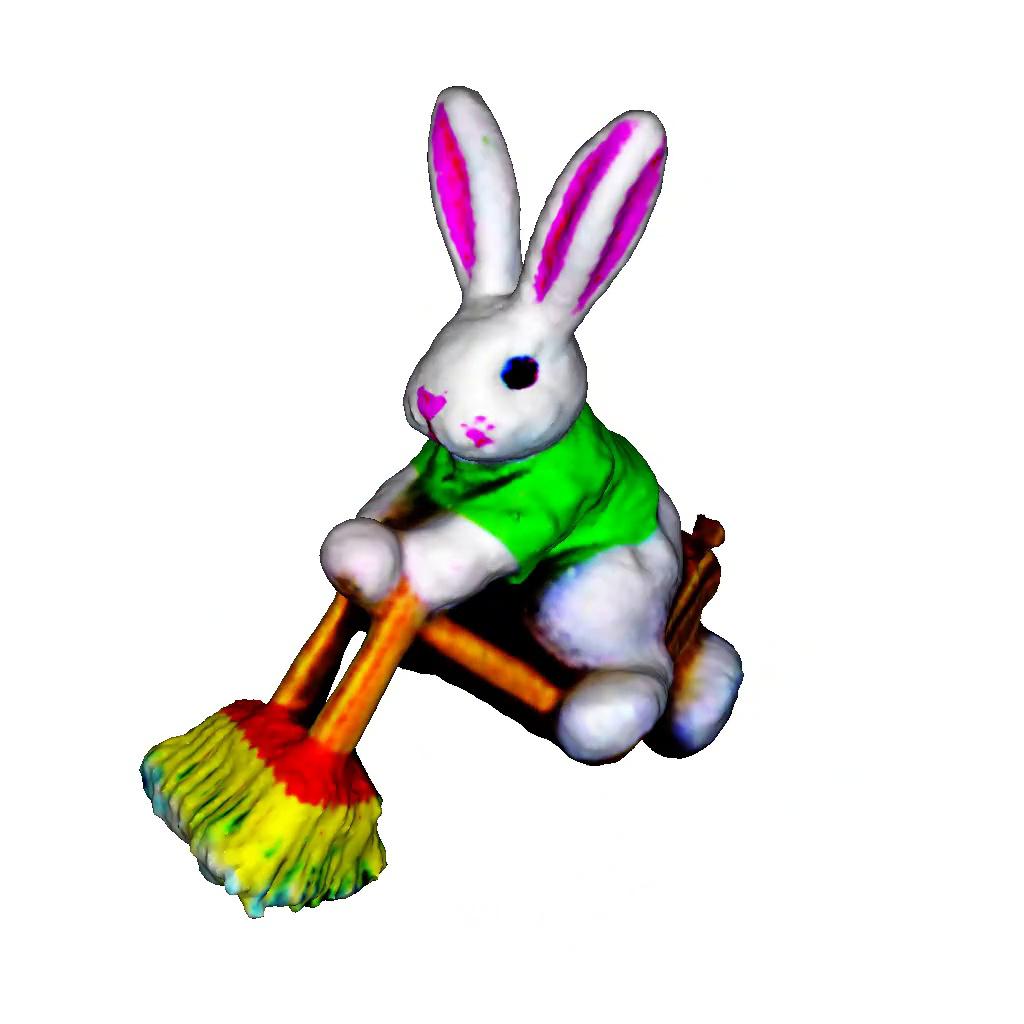} & 
        \includegraphics[align=c, width=0.15\linewidth, trim={3cm 1cm 2cm 4cm}, clip]{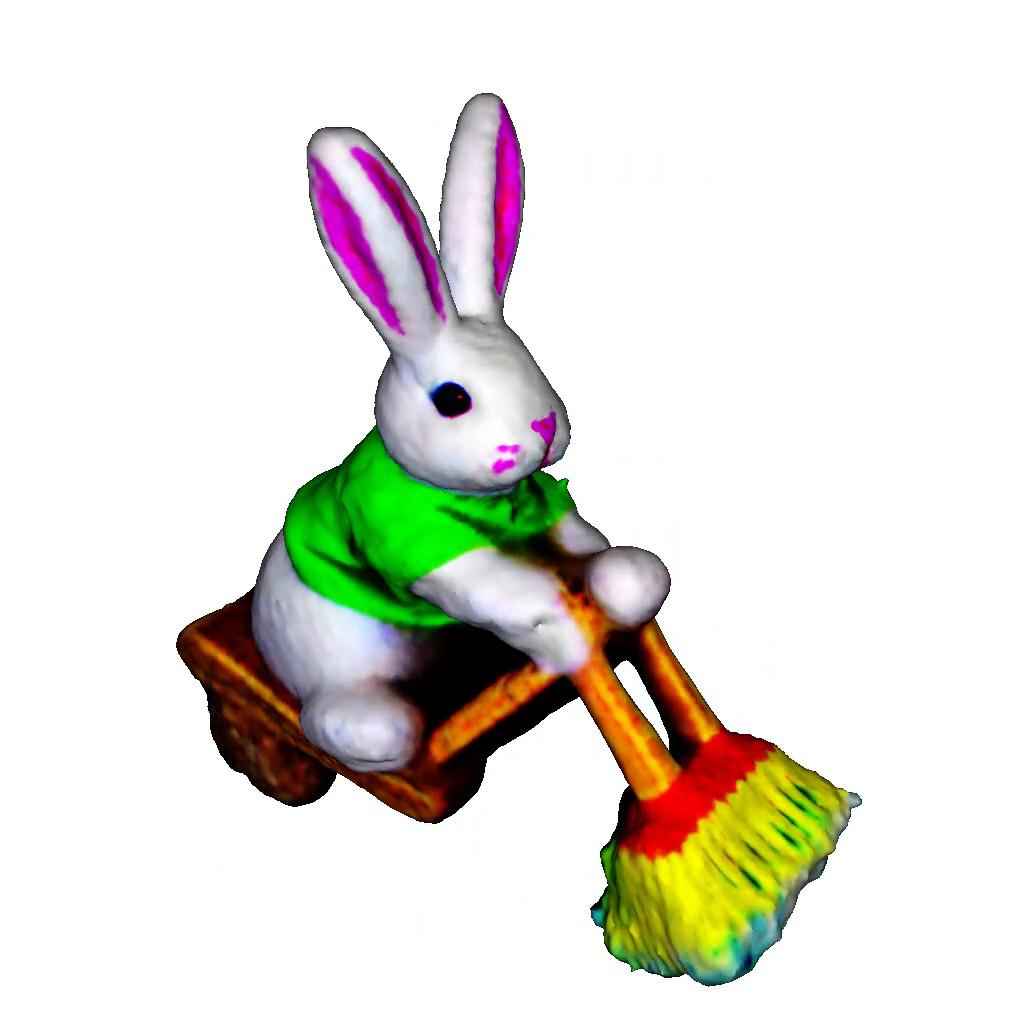} &
        \includegraphics[align=c, width=0.15\linewidth, trim={5cm 3cm 6cm 8cm}, clip]{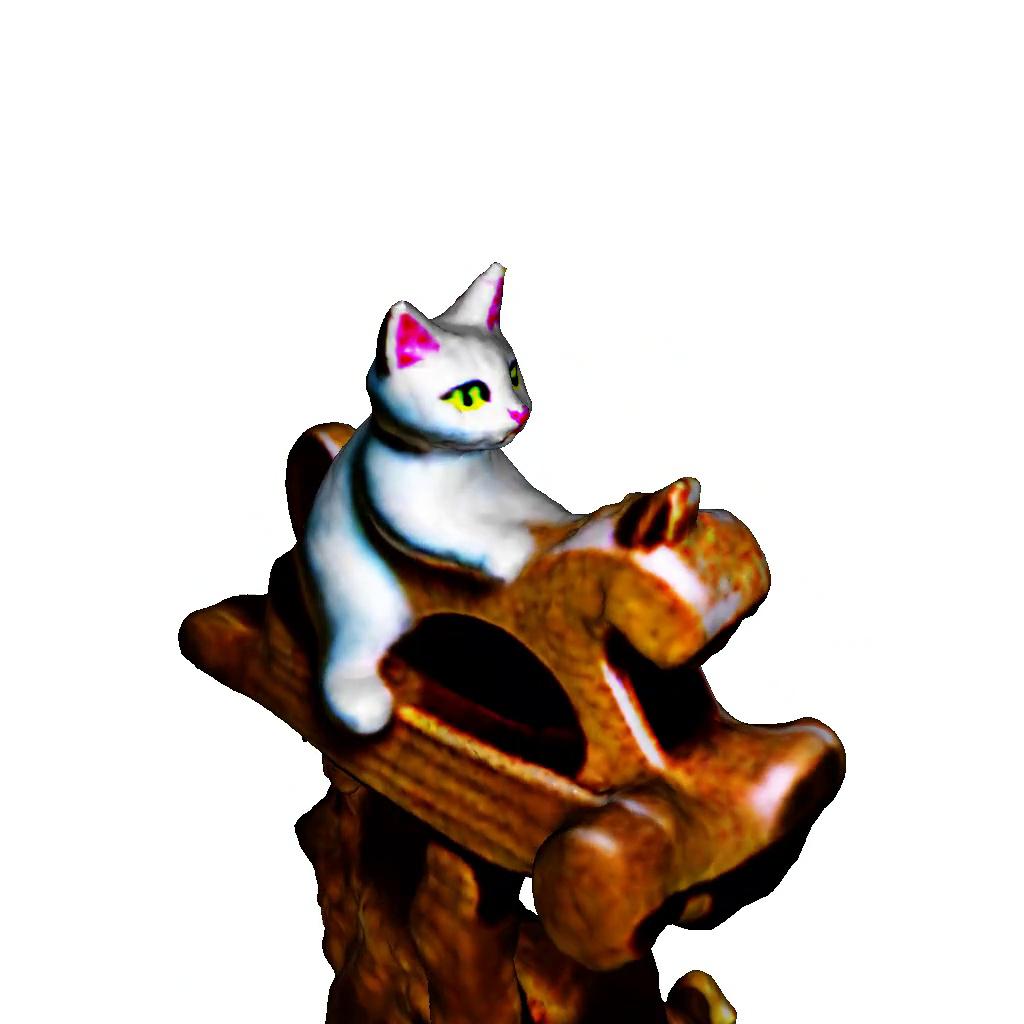} &  
        \includegraphics[align=c, width=0.15\linewidth, trim={5cm 3cm 6cm 8cm}, clip]{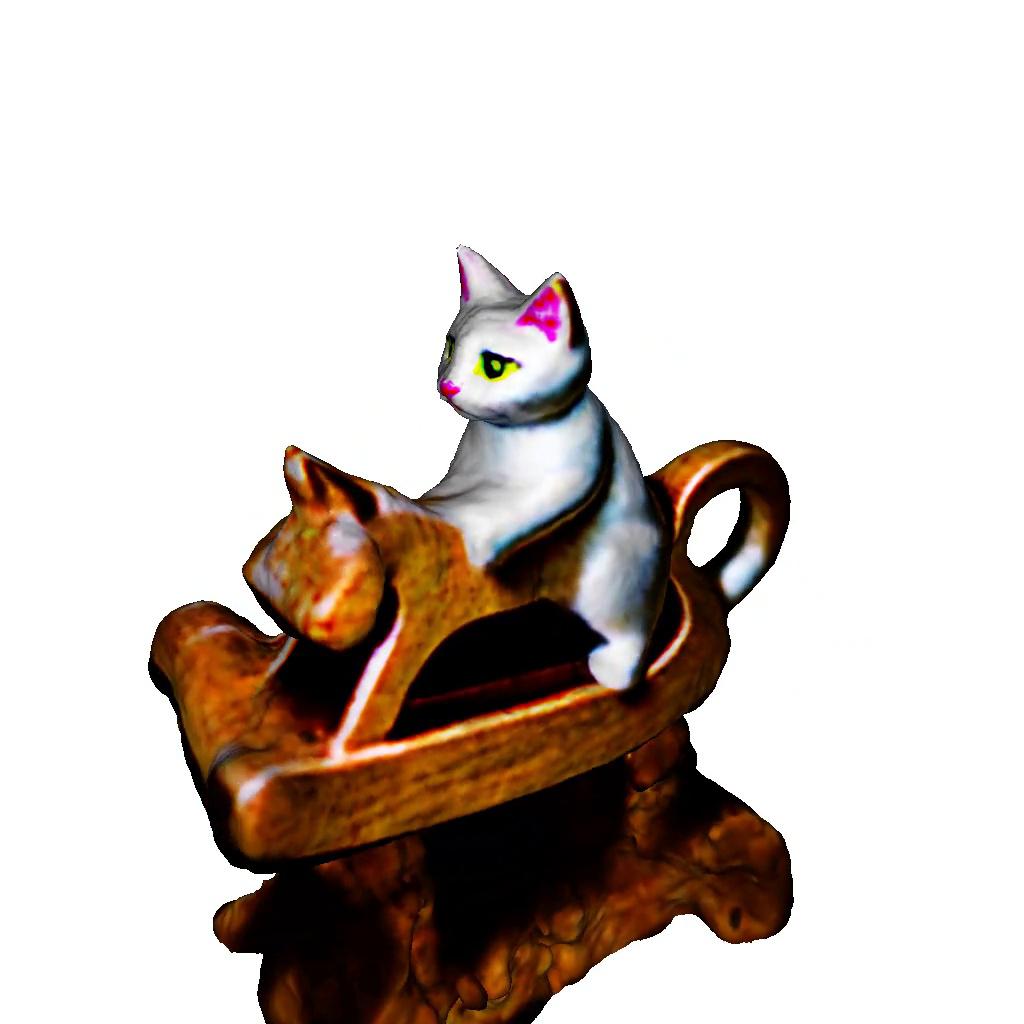} &
        \includegraphics[align=c, width=0.15\linewidth,trim={3cm 2cm 4cm 5cm}, clip]{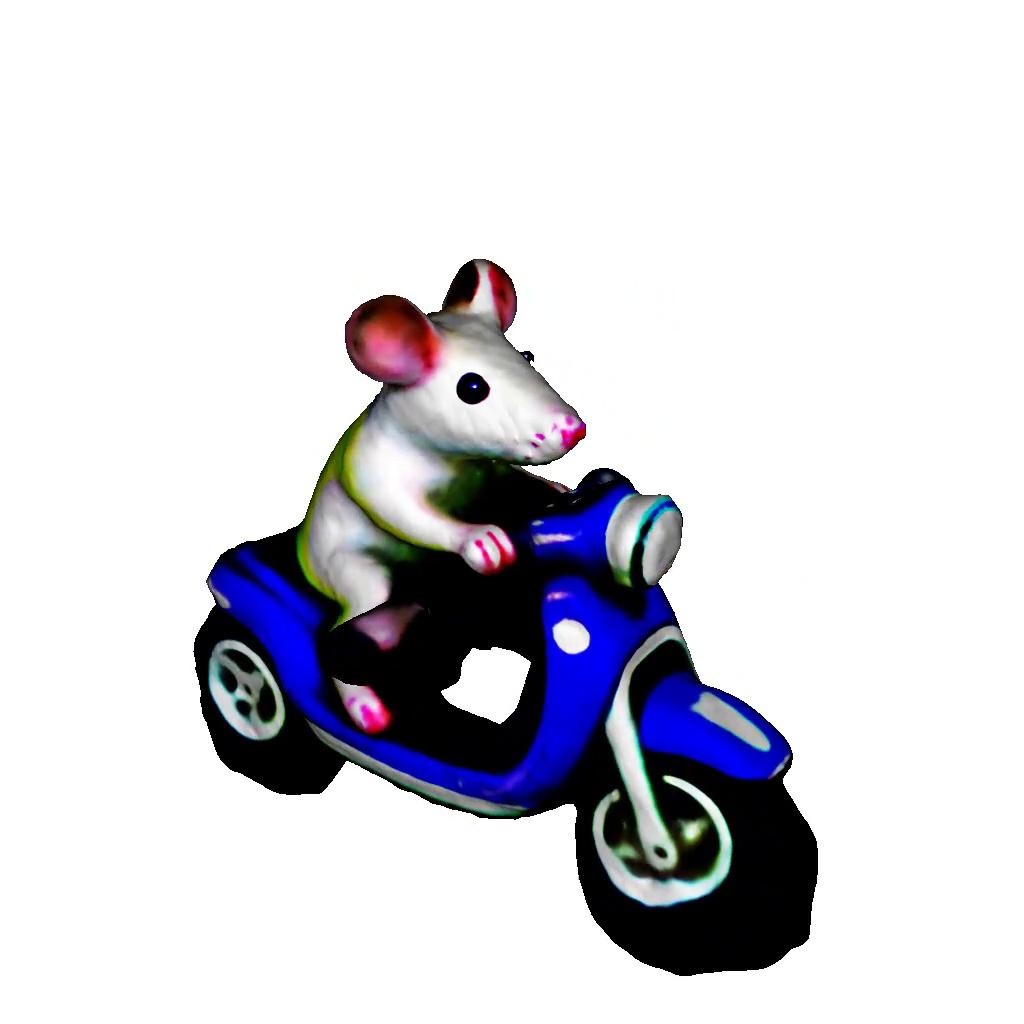} & 
        \includegraphics[align=c, width=0.15\linewidth,trim={3cm 2cm 4cm 5cm}, clip]{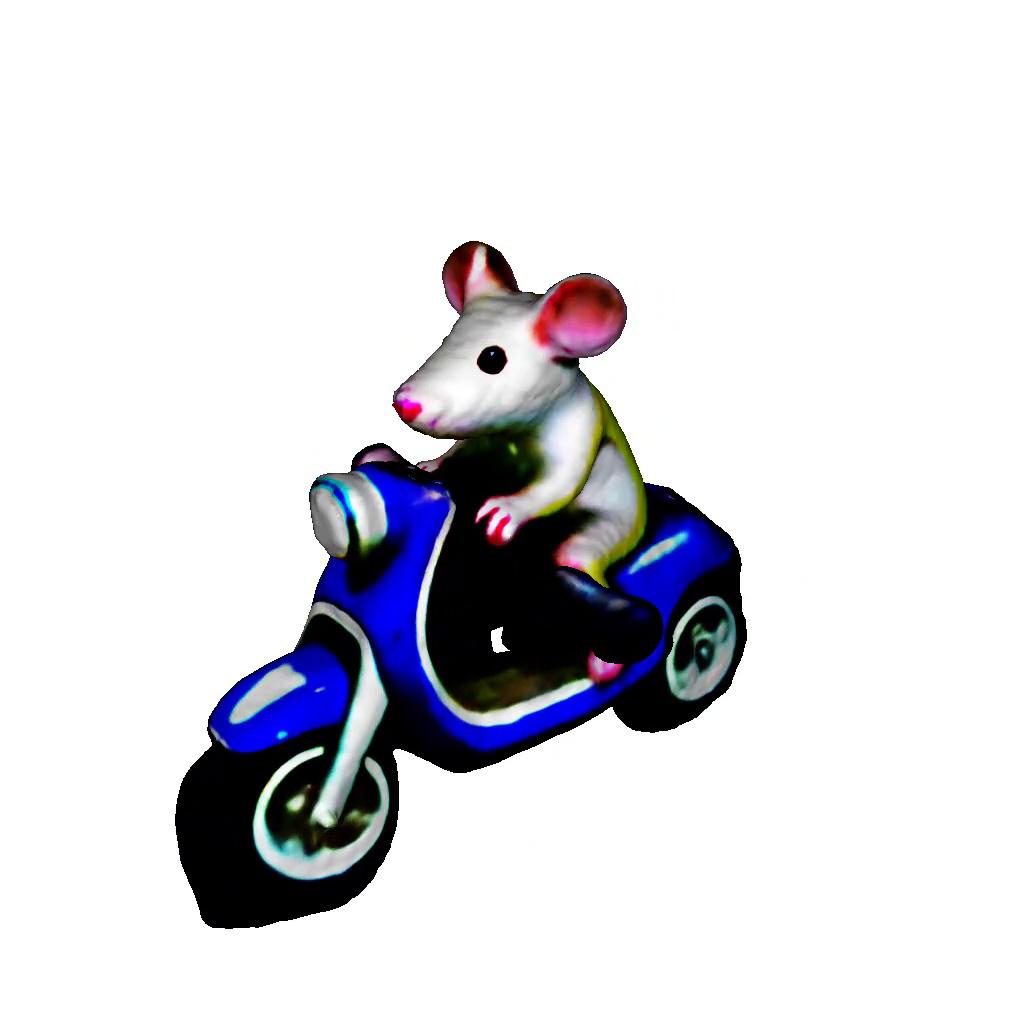} \\ 
        & \multicolumn{2}{c}{\footnotesize \textit{\textcolor{myorange}{bunny}, \textcolor{myblue}{broomstick}}} & 
        \multicolumn{2}{c}{\footnotesize \textit{\textcolor{myorange}{cat}, \textcolor{myblue}{rocking horse}}} & 
        \multicolumn{2}{c}{\footnotesize \textit{\textcolor{myorange}{rat}, \textcolor{myblue}{scooter}}} 
    \end{tabular}
   \vspace{-2mm}
   \caption{\textbf{\name with prompt-based editing}.
   Given a coarse model (first column) generated with a base prompt, we replace the underscored text with new text and fine-tune the NeRF to get a high-resolution NeRF model with LDM.
   We further fine-tune the high-resolution mesh with the NeRF model.
   Such a prompt-based editing method gives artists better control over the 3D generation output.}
   \label{fig:fine-tuneSD-prompt}
\end{figure*}

\section{Controllable 3D Generation}\label{sec:controllable}

As certain styles and concepts are difficult to express in words but easy with images, it is desirable to have a mechanism to influence the text-to-3D model generation with images.
We explore different image conditioning techniques as well as a prompt-based editing approach to provide users more control over the 3D generation outputs. 

\vspace{4pt}
\noindent\textbf{Personalized text-to-3D.}
DreamBooth~\cite{ruiz2022dreambooth} described a method to personalize text-to-image diffusion models by fine-tuning a pre-trained model on several images of a subject.
The fine-tuned model can learn to tie the subject to a unique identifier string, denoted as \V, and generate images of the subject when \V is included in the text prompt.
In the context of text-to-3D generation, we would like to generate a 3D model of a subject.
This can be achieved by first fine-tuning our diffusion prior models with the DreamBooth approach, and then using the fine-tuned diffusion priors with the \V identifier as part of the conditioning text prompt to provide the learning signal when optimizing the 3D model. 

To demonstrate the applicability of DreamBooth in our framework, we collect 11 images of one cat and 4 images of one dog.
We fine-tune \ediffi~\cite{balaji2022ediffi} and LDM~\cite{Rombach_2022_CVPR}, binding the text identifier \V to the given subject.
Then, we optimize the 3D model with \V in the text prompts.
We use a batch size of 1 for all fine-tuning.
For \ediffi, we use the Adam optimizer with learning rate $1\times 10^{-5}$ for 1,500 iterations; for LDM, we fine-tune with learning rate $1\times 10^{-6}$ for 800 iterations.
Fig.~\ref{fig:dreambooth} shows our personalized text-to-3D results: we are able to successfully modify the 3D models preserving the subjects in the given input images.

\vspace{4pt}
\noindent\textbf{Prompt-based editing through fine-tuning.}
Another way to control the generated 3D content is by fine-tuning a learned coarse model with a new prompt. Our prompt-based editing includes three stages.
(a) We train a coarse model with a base prompt.
(b) We modify the base prompt and fine-tune the coarse model with the LDM. 
This stage provides a well initialized NeRF model for the next step. Directly applying mesh optimization on a new prompt would generate highly detailed textures but could deform geometry only slightly.
(c) We optimize the mesh with the modified text prompt.
Our prompt-based editing can modify the texture of the shape or transform the geometry and texture according to the text.
The resulting scene models preserve the layer-out and overall structure. Such an editing capability makes the 3D content creation with \name more controllable. 
In Fig.~\ref{fig:fine-tuneSD-prompt}, we show two coarse NeRF models trained with the base prompt for the ``bunny'' and ``squirrel''.
We modify the base prompt, fine-tune the NeRF model in high resolution and optimize the mesh.
Results show that we can tune the scene model according to the prompt, \eg changing the ``baby bunny'' to ``stained glass bunny'' or ``metal bunny'' results in similar geometry but with a different texture.

\section{Conclusion}
We propose \name, a fast and high-quality text-to-3D generation framework. We benefit from both efficient scene models and high-resolution diffusion priors in a coarse-to-fine approach.
In particular, the 3D mesh models scale nicely with image resolution and enjoy the benefits of higher resolution supervision brought by the latent diffusion model without sacrificing its speed.
It takes 40 minutes from a text prompt to a high-quality 3D mesh model ready to be used in graphic engines.
With extensive user studies and qualitative comparisons, we show that \name is more preferable (61.7\%) by the raters compared to DreamFusion, while enjoying a $2\times$ speed-up.
Lastly, we propose a set of tools for better controlling style and content in 3D generation.
We hope with \name, we can democratize 3D synthesis and open up everyone's creativity in 3D content creation.

\vspace{8pt}
\noindent\textbf{Acknowledgements.}
We would like to thank Frank Shen, Yogesh Balaji, Seungjun Nah, James Lucas, David Luebke, Clement Fuji-Tsang, Charles Loop, Qinsheng Zhang, Zan Gojcic, and Jonathan Tremblay for helpful discussions and paper proofreading.
We would also like to thank Ben Poole, Ajay Jain, Jonathan T. Barron, and Ben Mildenhall for providing additional implementation details in DreamFusion.

\section*{\Large Appendix}
\appendix
\vspace{8pt}

\section{Author Contributions}
\textbf{All authors} have significant contributions on ideas, explorations, and paper writing.
Specifically, \textbf{CHL} and \textbf{TYL} led the research, developed fundamental code for experiments and organized team efforts.
\textbf{JG} led the experiments on generating high-resolution mesh models.
\textbf{LT} led the experiments on using high-resolution diffusion prior.
\textbf{TT} led the experiments on sparse scene representations.
\textbf{XZ} and \textbf{KK} led the experiments in controllable generation.
\textbf{XH} conducted the user study.
\textbf{SF} and \textbf{MYL} advised the research direction and designed the scope of the project.

\begin{figure*}
    \centering \setlength{\tabcolsep}{0.3pt} \renewcommand{\arraystretch}{0.5} 
     \begin{tabular}{cccc cccc}
        \multicolumn{8}{l}{\footnotesize{\textit{A corgi racing down the track*}}} \\
        \includegraphics[width=0.123\linewidth, height=0.123\linewidth]{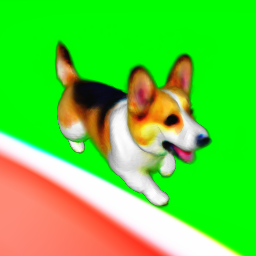}  
        &\includegraphics[width=0.123\linewidth]{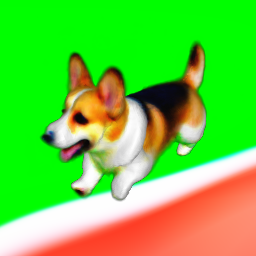} \hspace{1pt}
        &\includegraphics[width=0.123\linewidth]{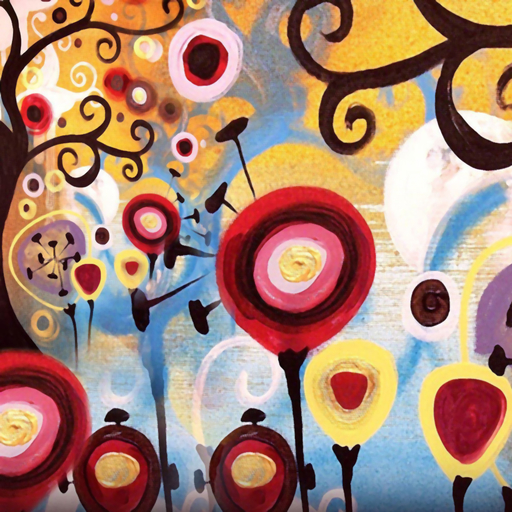} 
        &\includegraphics[width=0.123\linewidth]{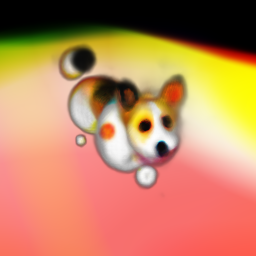}
        &\includegraphics[width=0.123\linewidth]{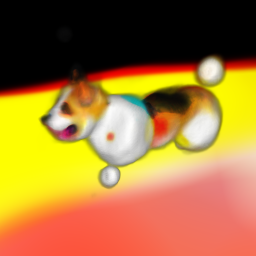} \hspace{1pt}
        &\includegraphics[width=0.123\linewidth, height=0.123\linewidth]{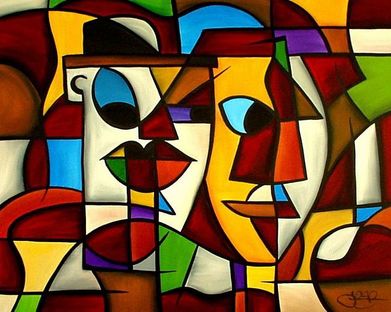}
        &\includegraphics[width=0.123\linewidth]{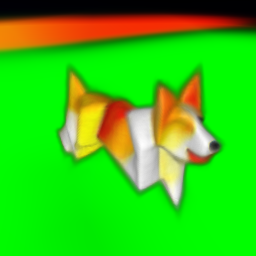}
        &\includegraphics[width=0.123\linewidth]{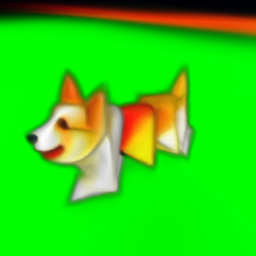}  \\
        \multicolumn{8}{l}{\footnotesize{\textit{A scooter*}}} \\
        \includegraphics[width=0.123\linewidth]{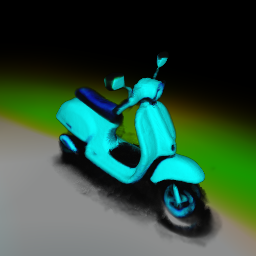}  
        &\includegraphics[width=0.123\linewidth]{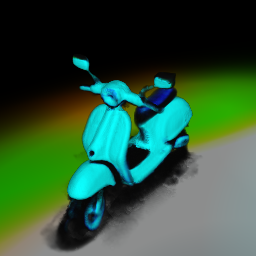} \hspace{1pt} & 
        &\includegraphics[width=0.123\linewidth]{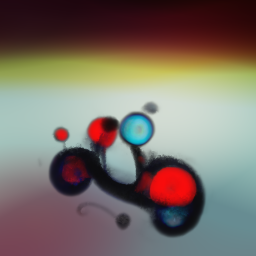}
        &\includegraphics[width=0.123\linewidth]{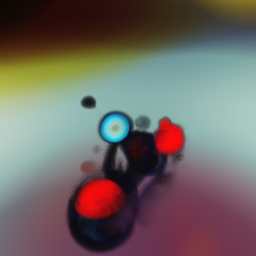} \hspace{1pt} &
        &\includegraphics[width=0.123\linewidth]{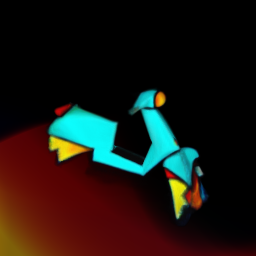}
        &\includegraphics[width=0.123\linewidth]{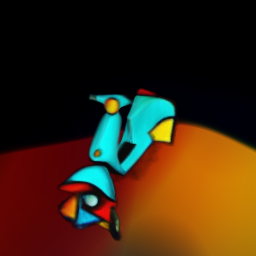}  \\        
        \end{tabular}

    \caption{\textbf{\name with image style transfer.}
    The style of the reference image is transferred to the 3D model by providing it to the diffusion model as conditional input. 
    We apply different styles during 3D synthesis with two different text prompts. 
    The first two columns show the 3D model optimized given the text without reference image. Afterwards, we show the reference and the corresponding 3D shapes.
    } 
    \label{fig:style_transfer}
\end{figure*} 
\begin{figure*}
    \centering \setlength{\tabcolsep}{0.5pt}
    \begin{tabular}{c cc cc cc cc} 
         ref img & 0,100 & 20,80 & 40,60 & 45,55 & 50,50 & 55,45 & 60,40 & 100,0 \\
         \includegraphics[width=0.10\linewidth, height=0.10\linewidth]{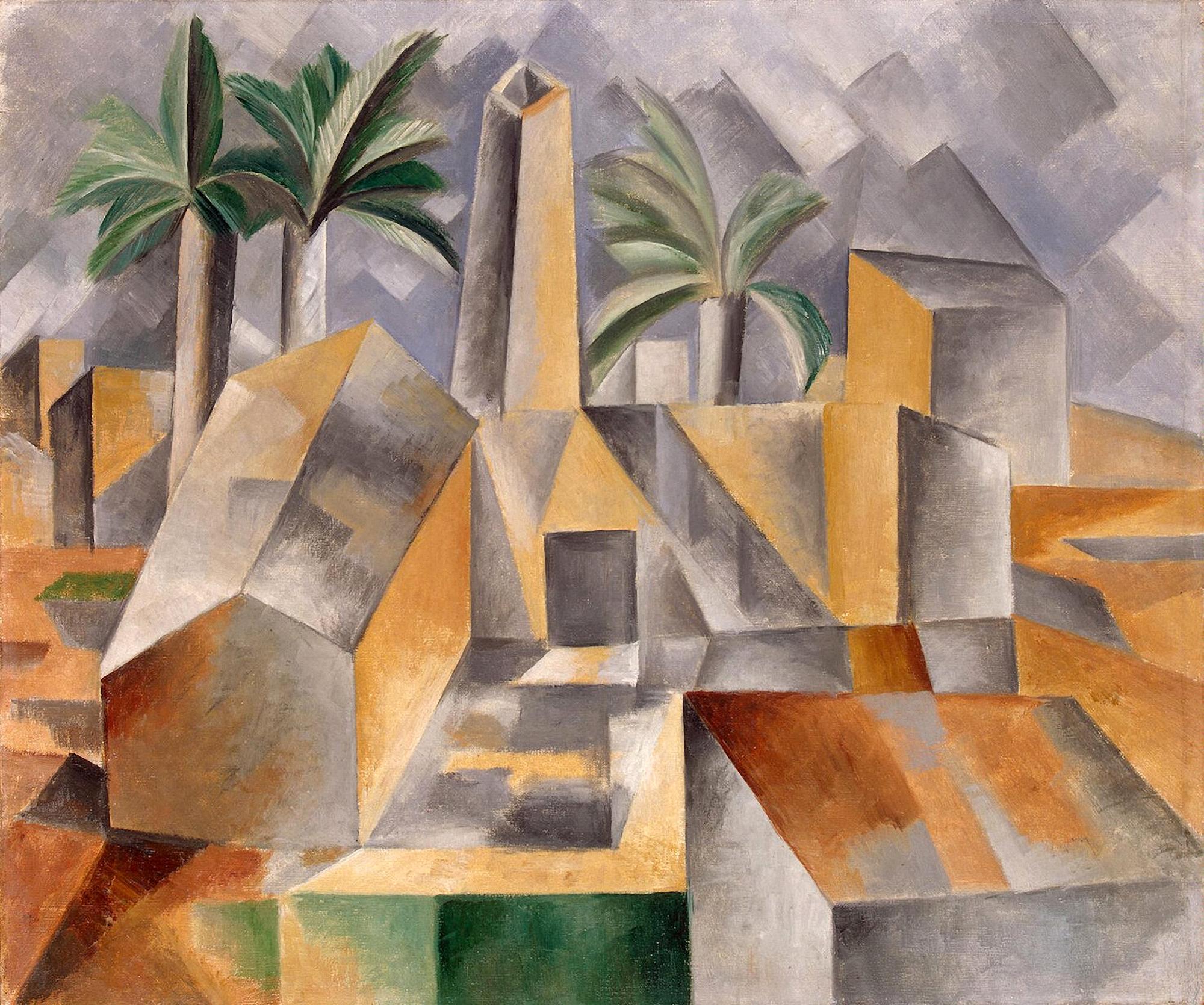} & 
         \includegraphics[width=0.10\linewidth]{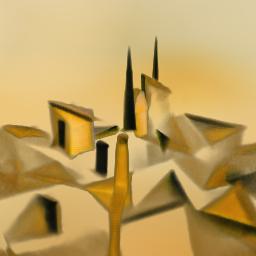} & 
         \includegraphics[width=0.10\linewidth]{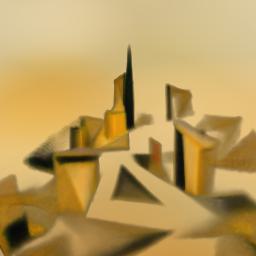} &
         \includegraphics[width=0.10\linewidth]{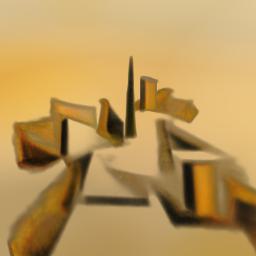} & 
         \includegraphics[width=0.10\linewidth]{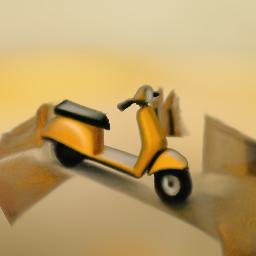} & 
         \includegraphics[width=0.10\linewidth]{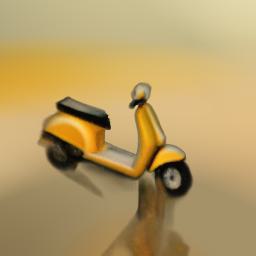} & 
         \includegraphics[width=0.10\linewidth]{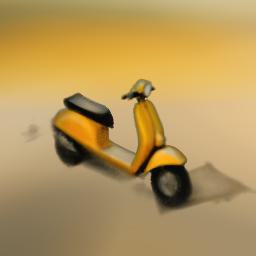} & 
         \includegraphics[width=0.10\linewidth]{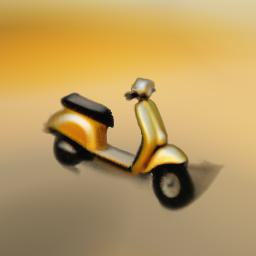} & 
         \includegraphics[width=0.10\linewidth]{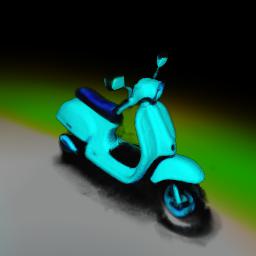} \\
         & \includegraphics[width=0.10\linewidth]{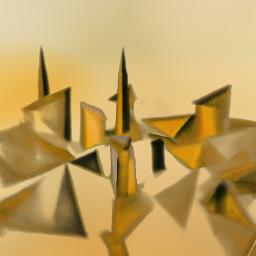} & 
         \includegraphics[width=0.10\linewidth]{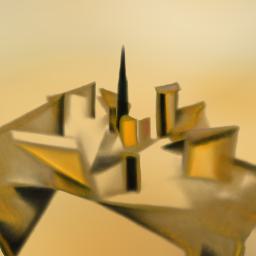} &
         \includegraphics[width=0.10\linewidth]{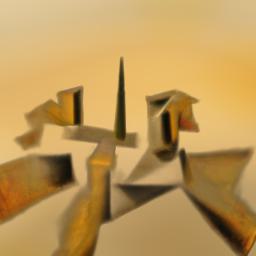} & 
         \includegraphics[width=0.10\linewidth]{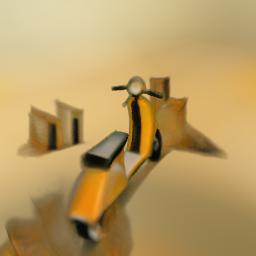} & 
         \includegraphics[width=0.10\linewidth]{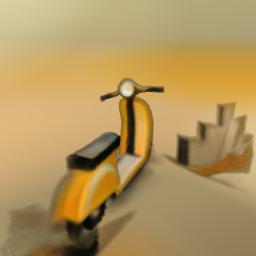} & 
         \includegraphics[width=0.10\linewidth]{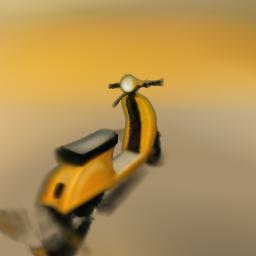} & 
         \includegraphics[width=0.10\linewidth]{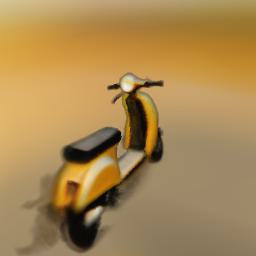} & 
         \includegraphics[width=0.10\linewidth]{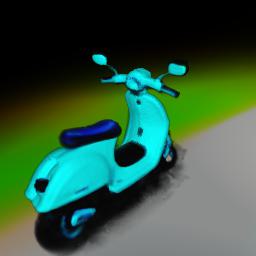} \\
         & \multicolumn{8}{c}{\textit{A DSLR photo of a scooter}}
    \end{tabular}
    \caption{\textbf{\name with image style transfer with different guidance weights.} Each column uses a different combination of guidance weights $\omega_{\textrm{text}}$, $\omega_{\textrm{joint}}$. All results are optimized with noise level threshold $t=1.0$. When $\omega_{\textrm{text}} = 0, \omega_{\textrm{joint}} = 100$, the setting is equivalent to using a single guidance weight. The style image dominates the resulting scene if the $\omega_{\textrm{joint}}$ is too large. We generally find that using a guidance weight combination around $50,50$ results in the best performance. }
    \label{fig:aba_style_transfer_guidance}
\end{figure*}
\begin{figure}
    \centering \setlength{\tabcolsep}{0.5pt}
    \begin{tabular}{c cc cc cc cc} 
         ref img & 0.3 & 0.5 & 1.0 \\
         \includegraphics[width=0.23\linewidth, height=0.23\linewidth]{figures/image_cond/style_transfer/brick.jpg} & 
         \includegraphics[width=0.23\linewidth]{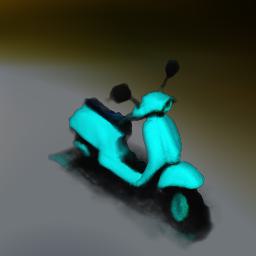} & 
         \includegraphics[width=0.23\linewidth]{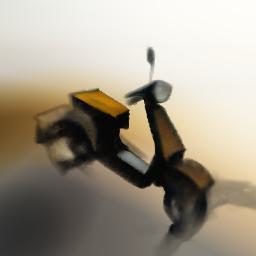} &
         \includegraphics[width=0.23\linewidth]{figures/image_cond/style_transfer_ablate_guid/g40_60_0.jpg}  \\
         & 
         \includegraphics[width=0.23\linewidth]{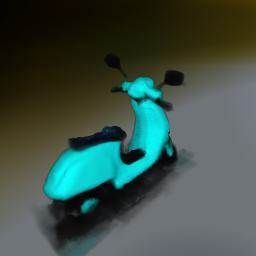} & 
         \includegraphics[width=0.23\linewidth]{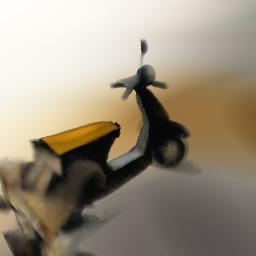} &
         \includegraphics[width=0.23\linewidth]{figures/image_cond/style_transfer_ablate_guid/g40_60_1.jpg}  \\
         & \multicolumn{3}{c}{\textit{A DSLR photo of a scooter}}
    \end{tabular}
    \caption{\textbf{\name with image style transfer with different noise level thresholds.} Each column uses a different noise level threshold $t$. All results are optimized with guidance weights $\omega_{\textrm{text}}$, $\omega_{\textrm{joint}}=40,60$. When the noise level threshold $t=0$, the setup is equivalent to using no style image guidance. We generally find that setting the threshold around $0.5$ provides the best performance. }
    \label{fig:aba_style_sigma}
\end{figure}
\begin{figure}
    \centering 
    \includegraphics[width=\linewidth, trim={0cm 0.0cm  5.0cm 1.5cm}, clip]{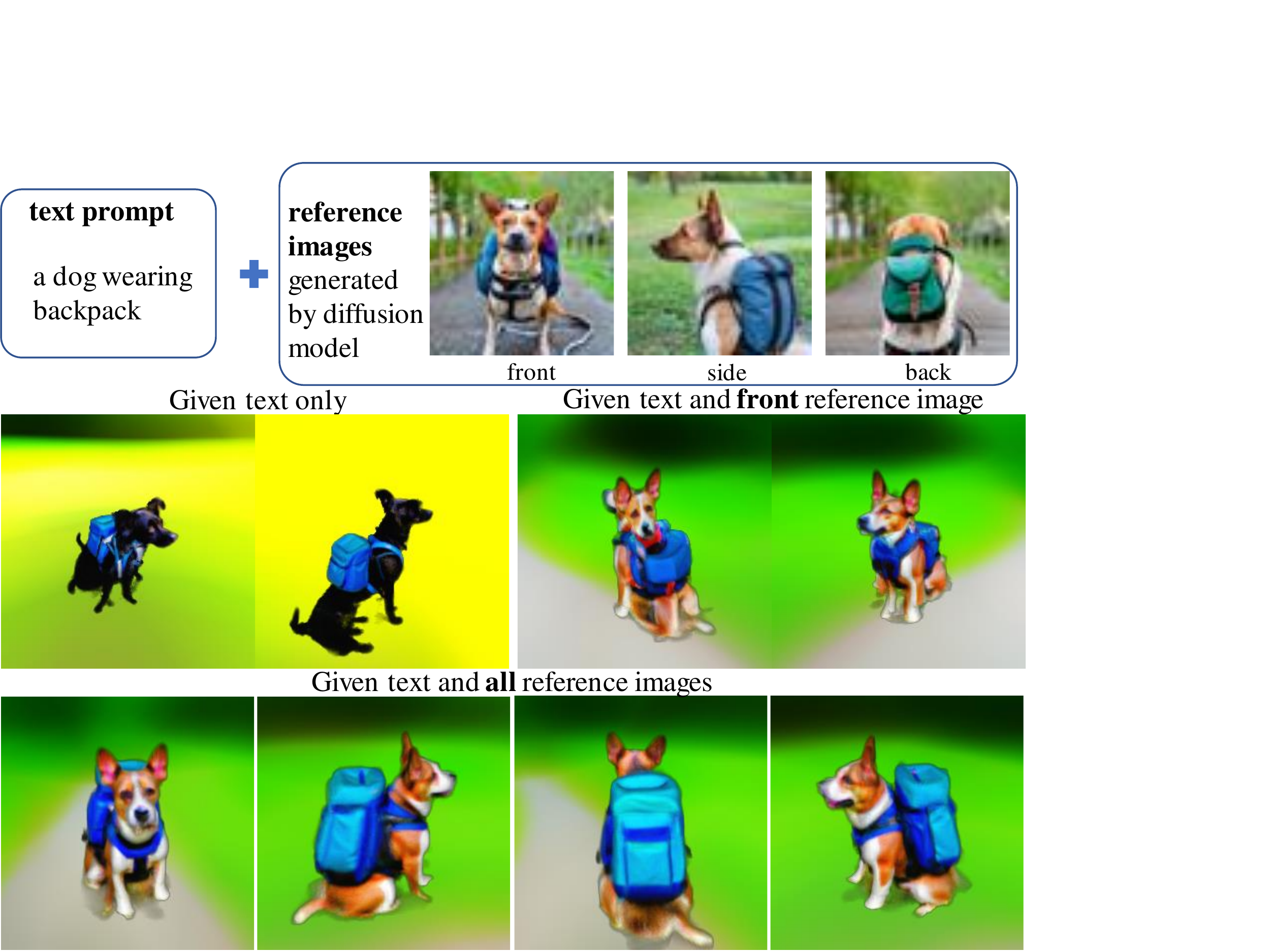} 
    \caption{\textbf{\name with image content transfer.} Given the reference images generated by the diffusion prior, we optimize 3D models that look similar to the object in the images. 
    We show the generated 3D models given \textit{(i)} text only, \textit{(ii)} text and front view's reference image only and \textit{(iii)} text and different view's reference images. \textit{(ii)} and \textit{(iii)} preserve the look of the dog in the image. With multiple reference images, \textit{(iii)} yields higher quality and more 3D-consistent outputs. 
  }
    \label{fig:sup_style_guide_content_transfer}
\end{figure}

\section{Implementation Details}

We follow the implementation details described by Poole~\etal~\cite{poole2022dreamfusion} as closely as possible.
We refer readers to the Dreamfusion paper~\cite{poole2022dreamfusion} for context and list the major differences below.

\vspace{4pt}
\noindent\textbf{Architectural details.}
As aforementioned in the main paper, we adopt a multi-resolution hash grid encoding architecture from Instant NGP~\cite{muller2022instant} instead of using a large global coordinate-based MLP architecture.
We use 16 levels of hash dictionaries of size $2^{19}$ and dimension $4$, spanning 3D gird resolutions from $2^4$ to $2^{12}$ with an exponential growth rate.
We use single-layer MLPs with $32$ hidden units to predict all of RGB color, volume density, and normal, where the inputs to the MLPs are the concatenated feature vectors from the multi-resolution hash encoding sampled with trilinear interpolation (we refer readers to the Instant NGP paper~\cite{muller2022instant} for more details in this representation).
We perform density-based pruning to sparsify the Instant NGP representation with an octree structure every 10 iterations.
This allows us to more efficiently render pixels using empty space skipping, even with 3D points as dense as $1024$ samples per ray.
We do not use the contracting reparametrization of unbounded scenes from Mip-NeRF 360~\cite{barron2022mip} as it is not supported by our sparse representation.

\vspace{4pt}
\noindent\textbf{Scene representation.}
For the coarse neural field representation, we use a bounding sphere of radius $2$ for our experiments.
We use the $\mathrm{softplus}$ activation for the density prediction and follow Poole~\etal~\cite{poole2022dreamfusion} to add an initial spatial density bias to encourage the optimization to focus on the object-centric neural field.
We empirically found that using a linear form of spatial density bias helps stabilize the optimization, more formally written as
\begin{align}
    \tau_\text{init}(\boldsymbol{\mu}) = \lambda_\tau \cdot \left( 1 - \frac{\|\boldsymbol{\mu}\|_2}{c} \right) \;,
\end{align}
where $\boldsymbol{\mu}$ is the 3D location, $\lambda_\tau = 10$ is the density bias scale, and $c=0.5$ is an offset scale.
Different from DreamFusion, however, we add this density bias to the \emph{pre}-activation; as a result, the post-activation of the density prediction will vary continuously from $\mathrm{softplus}(\lambda_\tau)$ to $0$ as a function of $\|\boldsymbol{\mu}\|_2$.

\vspace{4pt}
\noindent\textbf{Camera and light augmentations.}
We follow Poole~\etal~\cite{poole2022dreamfusion} to add random augmentations to the camera and light sampling for rendering the shaded images.
Differently,
(a) we sample the point light location such that the angular distance from the random camera center location (\wrt the origin) is sampled from $\psi_\text{cam} \sim \mathcal{U}(0, \pi/3)$ with a random point light distance $r_\text{cam} \sim \mathcal{U}(0.8, 1.5)$, and
(b) we use a ``soft'' version of the textureless and albedo-only augmentation such that various strengths of shading in the rendered images are seen during optimization.
(c) we sample the camera distance from $\mathcal{U}(1.5, 2)$, and the focal length  $\mathcal{U}(0.7, 1.35)$. When training with high resolution diffusion prior, we increase the focal length and sample from $\mathcal{U}(1.2, 1.8)$.

\vspace{4pt}
\noindent\textbf{Optimization.}
Unless otherwise specified, we optimize the coarse model with batch size $32$ using the Adam optimizer~\cite{kingma2014adam} with a learning rate of $1\times 10^{-2}$ without warmup and decay.
Note that the large global coordinate-based MLP architecture in DreamFusion~\cite{poole2022dreamfusion} limits its optimization to only an effective batch size of $8$.
For the coarse model, we add the opacity regularization as suggested by Poole~\etal~\cite{poole2022dreamfusion} to encourage sparsity in the volume density field, but we do not add the orientation regularization as we empirically found it to hurt optimization.

\vspace{4pt}
\noindent\textbf{Score Distillation Sampling.}
In the first stage, we sample the timestep $t\sim \mathcal{U}(0, 1)$ and set $w(t)=1$.
In the second stage, we find the range of timestep $t$ in SDS affects the quality. We sample $t \sim \mathcal{U}(0.02, 0.5)$ in our experiments. In general, setting $t_\text{max}$ in the range of $[0.5, 0.7]$ works well. We set $w(t)=\sigma_t^2$ in this stage.

\begin{figure}[t!]
\centering 
    \includegraphics[width=\linewidth]{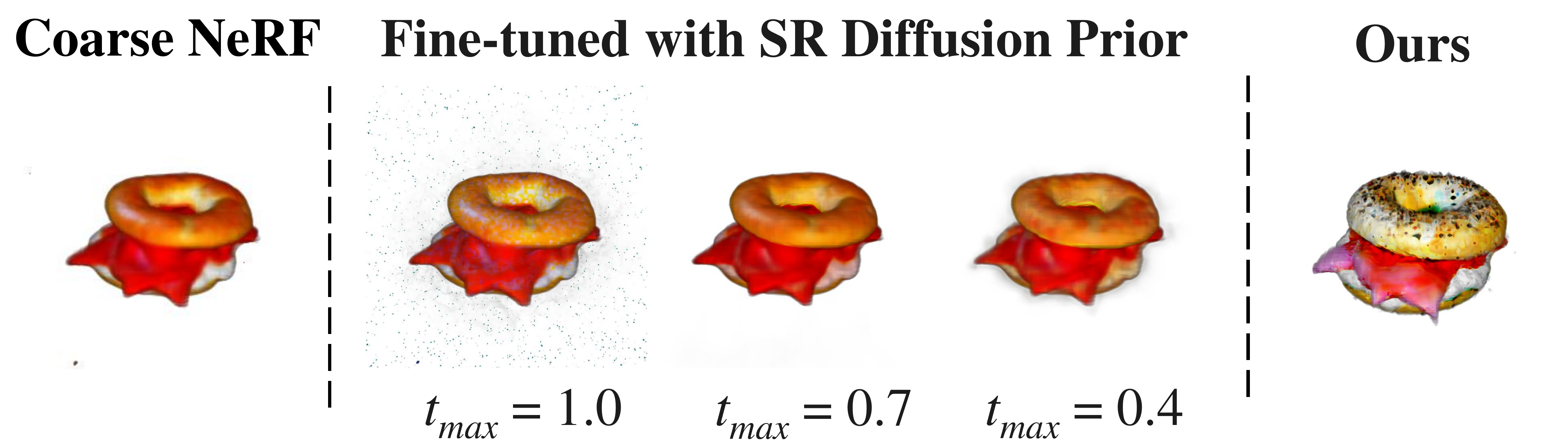}
    \vspace{-6mm}
    \caption{Fine-tuning NeRF with SR prior fails to add high-resolution details.
    $t_\text{max}$ is the maximum timestep sampled in SDS.
    }
    \label{fig:sr_nerf}
\end{figure}

\section{Alternative High-Resolution Prior}
In addition to LDM, we also consider using Super Resolution (SR) diffusion prior~\cite{balaji2022ediffi, saharia2022photorealistic} for increasing the resolution of a coarse model. This diffusion model is trained to %
generate a high-resolution image conditioning on a low-resolution input image. In SDS, this model predicts noises added in high resolution, i.e.,  $\epsilon_\phi(x_t;y,t,x_\text{low})$, where $x_\text{low}$ denotes a $64 \times 64$ low-resolution image. We render $x_\text{low}$ with a frozen coarse model to optimize the second-stage fine model. Fig.~\ref{fig:sr_nerf} shows this approach fails to add high-quality details to the input coarse model.

\section{Style-Guided Text-to-3D Synthesis}

We also explore controlling the 3D generation with multi-modal conditioning.
The \ediffi diffusion prior~\cite{balaji2022ediffi} is designed such that it can condition on a reference image when performing text-to-image generation.
Such an image conditioning design makes it easy to change the style of the generated output.
However, we find that na\"ively feeding the style image as input to the model when computing the SDS gradient can result in a poor 3D model that is essentially overfitting to the input image.
We hypothesize that the conditioning signal by the image is significantly stronger than the text prompt during optimization.
To better balance the guidance strength between image and text conditioning, we extend our model's classifier-free guidance scheme~\cite{ho2021classifierfree} and compute the final $\tilde{\epsilon}_\phi(x_t; y_{\textrm{text}}, y_{\textrm{image}}, t)$:
\begin{align} \label{eq:cfg_extended}
    \tilde{\epsilon}_\phi(x_t; &~ y_{\textrm{text}}, y_{\textrm{image}}, t) = \epsilon_\phi(x_t; t) \nonumber \\
    &+ \omega_{\textrm{text}} [\epsilon_\phi(x_t; y_{\textrm{text}}, t) - \epsilon_\phi(x_t; t)] \nonumber \\
    &+ \omega_{\textrm{joint}} [\epsilon_\phi(x_t; y_{\textrm{text}}, y_{\textrm{image}}, t) - \epsilon_\phi(x_t; t)] \;,
\end{align}
where $y_{\textrm{text}}$ and $y_{\textrm{image}}$ are text and image conditioning respectively,
and $\omega_{\textrm{text}}$ and $\omega_{\textrm{joint}}$ are the guidance weights for text and joint text-and-image conditioning respectively.
Note that for $\omega_{\textrm{joint}} = 0$, the scheme is equivalent to standard classifier-free guidance with respect to text conditioning only.

Fig.~\ref{fig:style_transfer} shows our style-guided text-to-3D generation results. When optimizing the 3D model, we feed the reference image to the \ediffi model. 
We set 
$\omega_{\textrm{text}},\omega_{\textrm{joint}}=50,50$ or $40,60$ and apply the image guidance when $t<0.5$ only.
We do not provide high-resolution results for this experiment because LDM does not support reference image conditioning.

\vspace{4pt}
\noindent\textbf{Guidance weight and noise level threshold.}
We ablate different combinations of guidance weights and noise level thresholds in Figs.~\ref{fig:aba_style_transfer_guidance} and~\ref{fig:aba_style_sigma}, respectively. The guidance weights $\omega_{\textrm{text}}$ and $\omega_{\textrm{joint}}$ balance the guidance strength during optimization (see Eq.~\ref{eq:cfg_extended}). A similar guidance formulation has also been used by Liu~\etal for compositional text-to-image generation~\cite{liu2022compositional}. 
We also find that applying the image conditioning only 
below a certain noise level threshold
can help control style transfer. The intuition is that image-based style guidance is most relevant for optimizing the generated 3D object's details, which are modeled at lower noise levels. Notice that we do not provide high-resolution results for this experiment because LDM does not support image conditioning inputs. 

\vspace{4pt}
\noindent\textbf{Content image as reference.}
We also explore using multiple images as inputs during 3D synthesis to transfer the content in the images to the 3D model, as shown in Fig.~\ref{fig:sup_style_guide_content_transfer}: Given a text prompt, we first ask the eDiff-I model to generate the front view, side view and back view images. When optimizing the 3D model for the same text prompt from different views, we then feed the corresponding generated view image as input to guide the 3D synthesis. This approach requires some degree of consistency with respect to subject identity across the different view images, which can be achieved by generating a set of different view images first and choosing accordingly. 
Overall, the experiment shows that we can apply the text-to-image diffusion model to generate images that can be used for guidance during 3D model optimization. As we see, this does not only provide enhanced control by preserving the identity of the subjects in the images, but also improves output quality and 3D consistency. Generally, depending on image type, image conditioning can be used either for object-centric content transfer to 3D (Fig.~\ref{fig:sup_style_guide_content_transfer}) or for abstract 3D stylization (Figs.~\ref{fig:style_transfer}, \ref{fig:aba_style_transfer_guidance}, and \ref{fig:aba_style_sigma}).

\section{Additional Results}

\begin{figure*}[t!] 
    \renewcommand{\arraystretch}{0.6}
    \centering 
    \setlength{\tabcolsep}{0.pt}
    \begin{tabular}{cccc ccc} 
    \includegraphics[align=c, width=0.1\linewidth, trim={0cm 0cm 0.3cm 0cm}, clip]{figures/prompt_edit/prompt_edit_base2.pdf} 
    & \includegraphics[align=c, width=0.15\linewidth, trim={0cm 35cm 0cm 0cm}, clip]{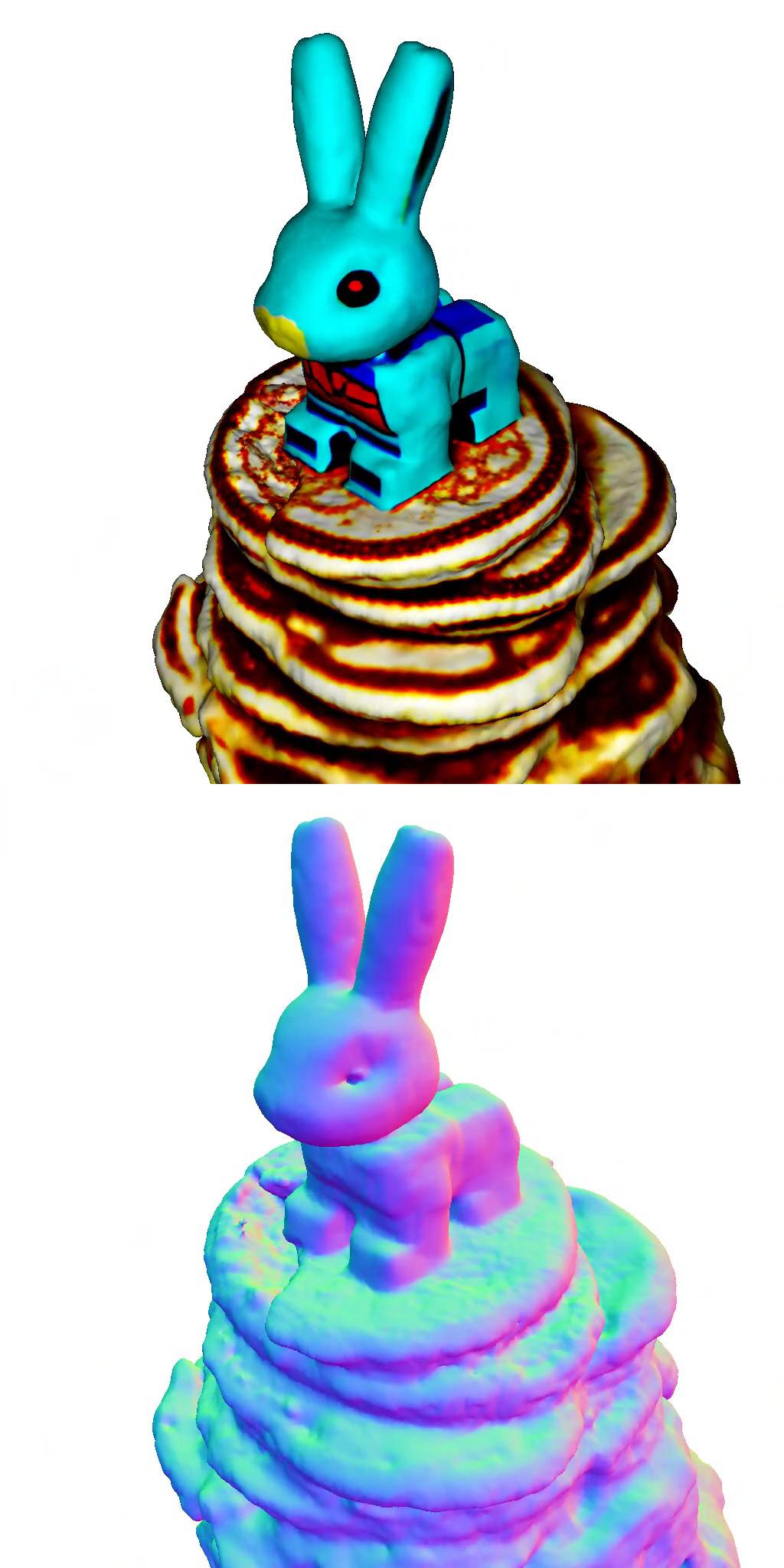}
    & \includegraphics[align=c, width=0.15\linewidth, trim={0cm 35cm 0cm 0cm}, clip]{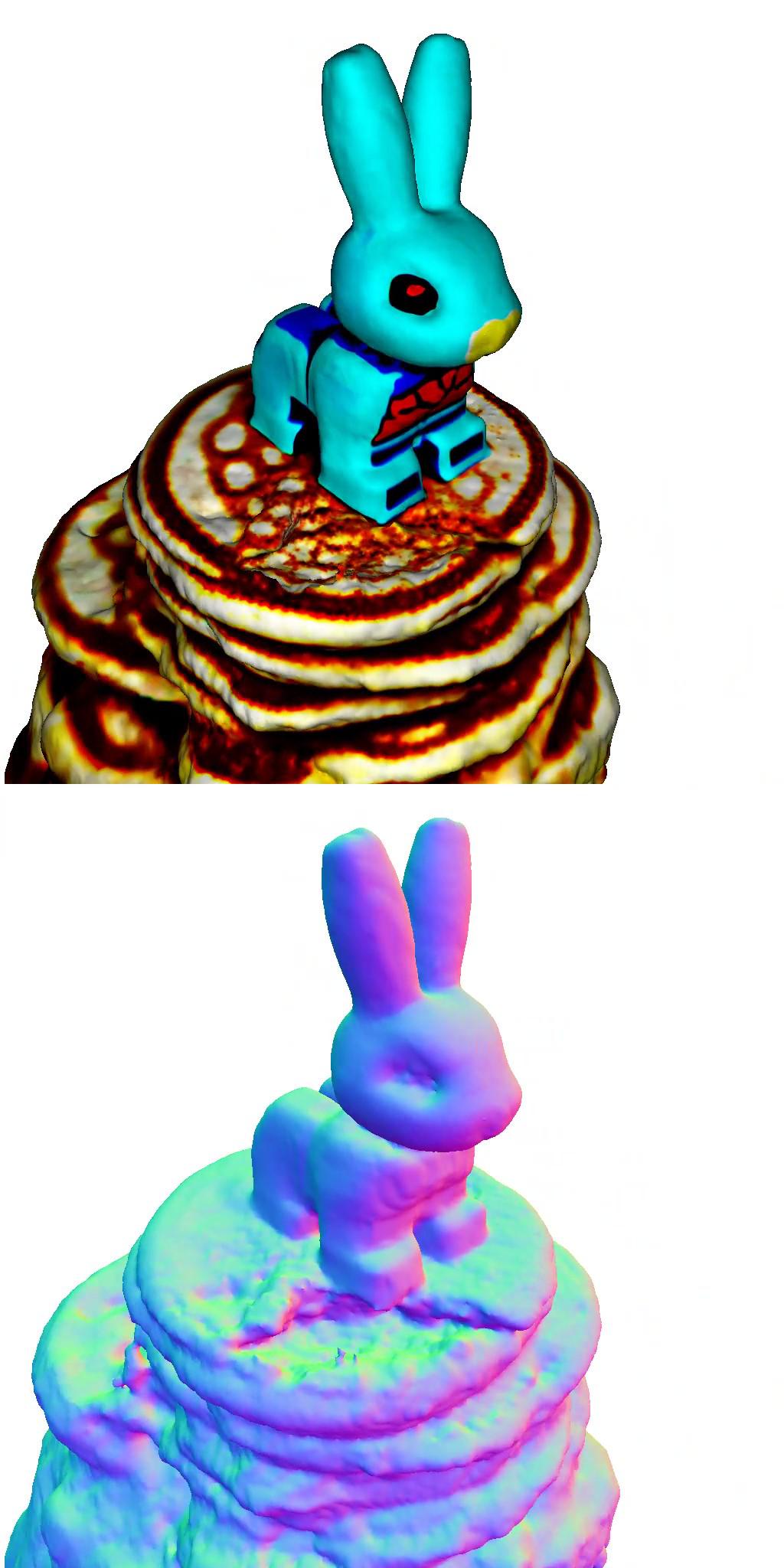}
    & \includegraphics[align=c, width=0.15\linewidth, trim={0cm 35cm 0cm 0cm}, clip]{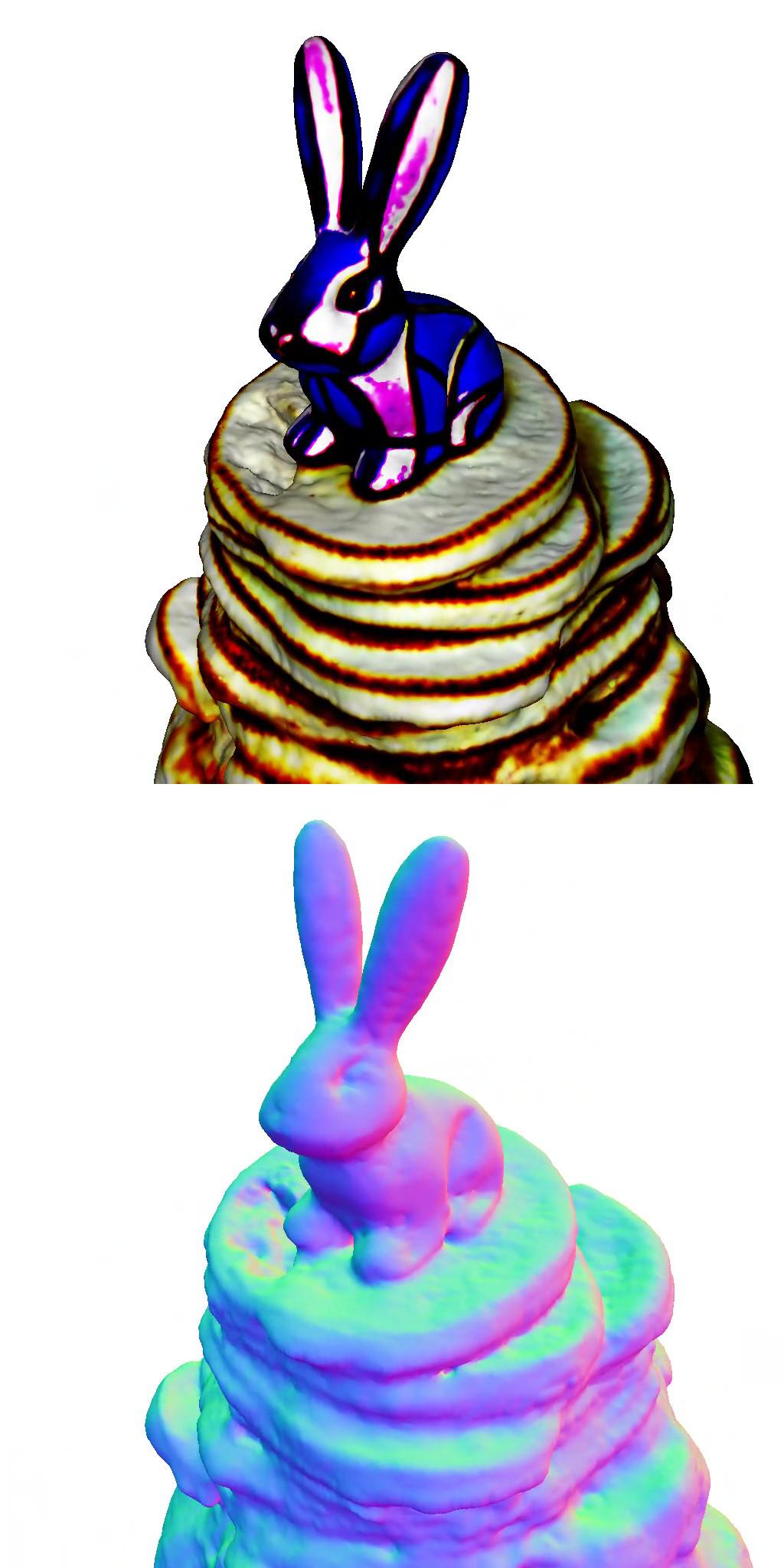}
    & \includegraphics[align=c, width=0.15\linewidth, trim={0cm 35cm 0cm 0cm}, clip]{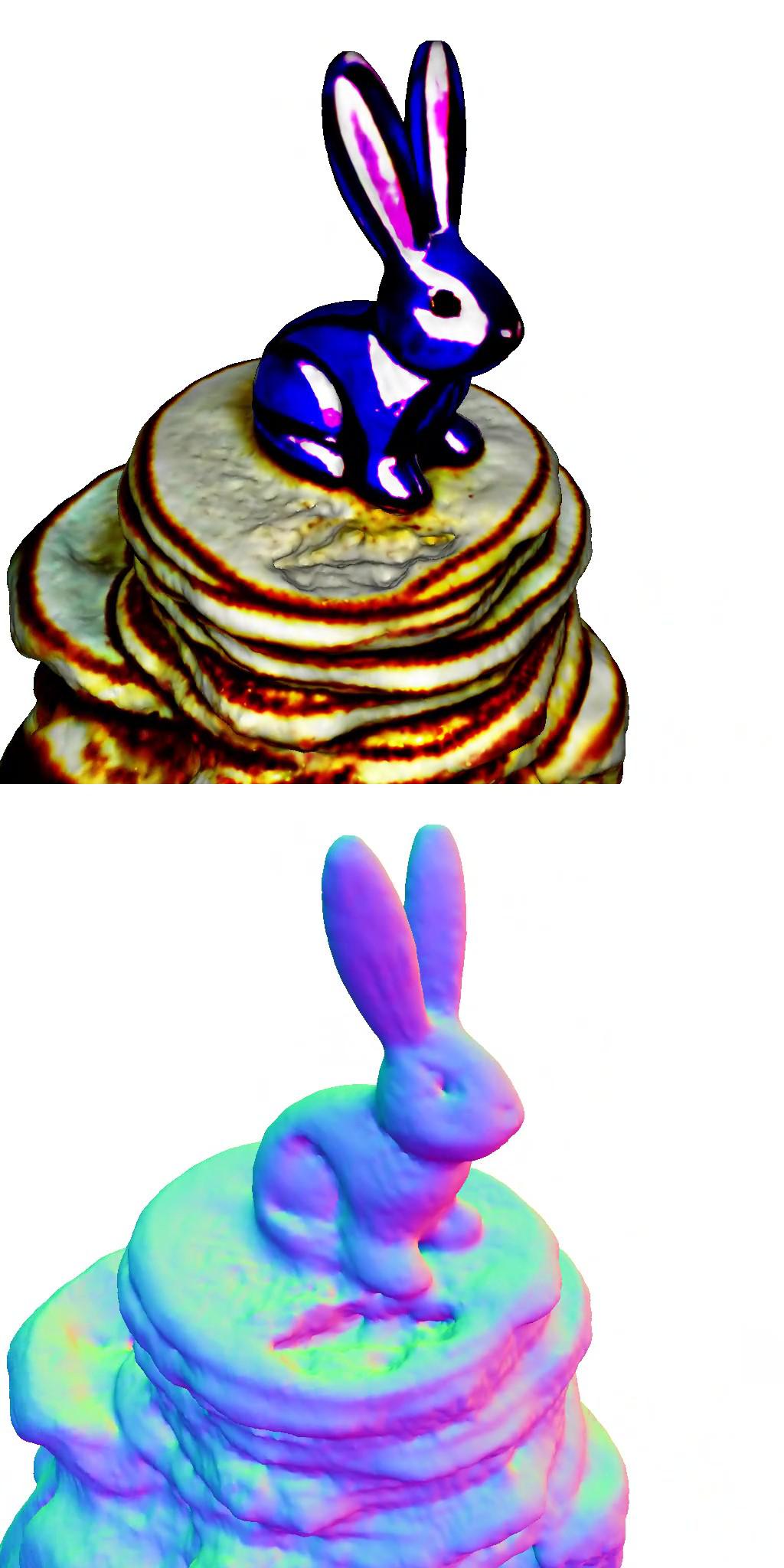}

    & \includegraphics[align=c, width=0.15\linewidth, trim={0cm 35cm 0cm 0cm}, clip]{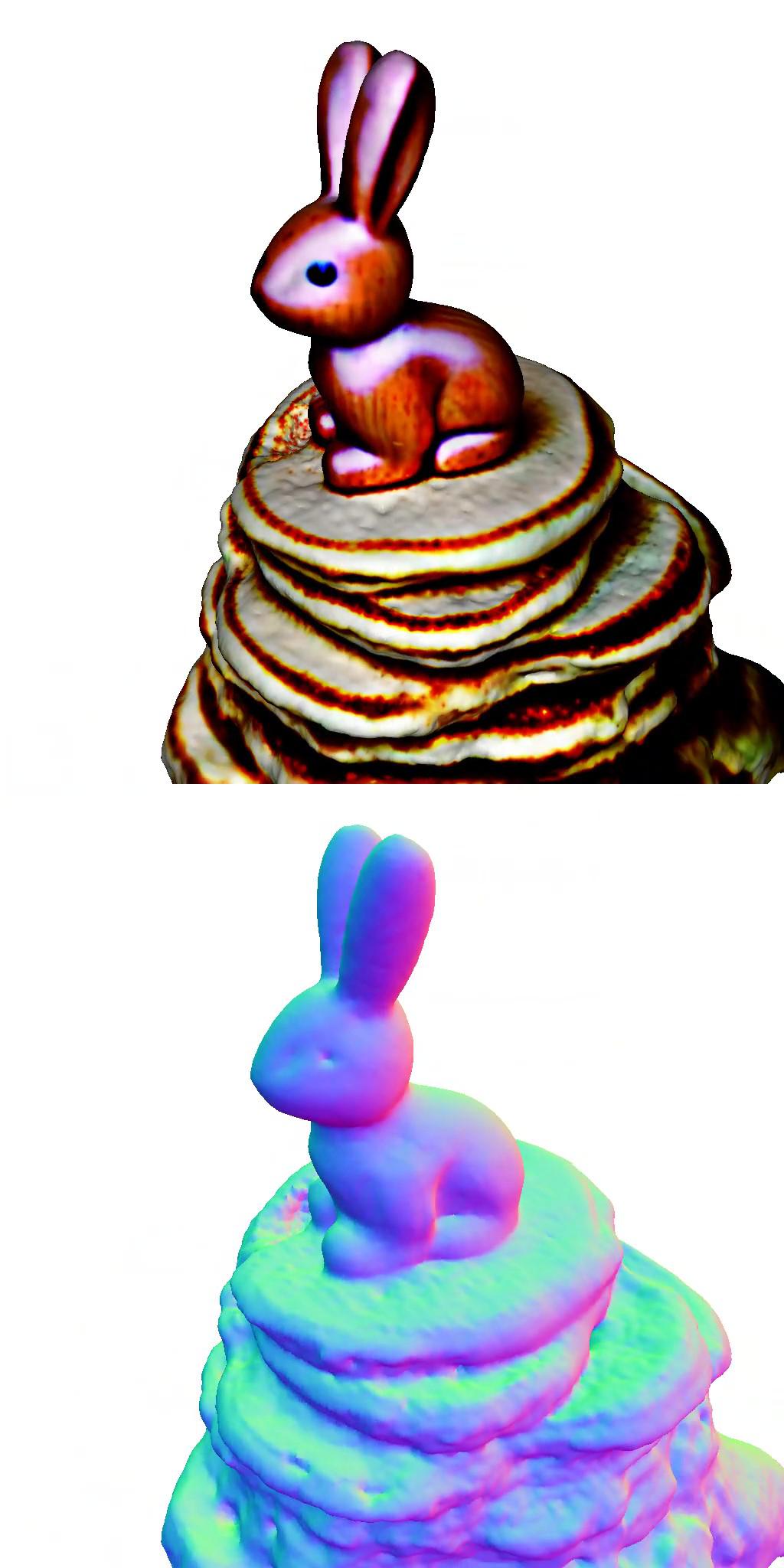}
    & \includegraphics[align=c, width=0.15\linewidth, trim={0cm 35cm 0cm 0cm}, clip]{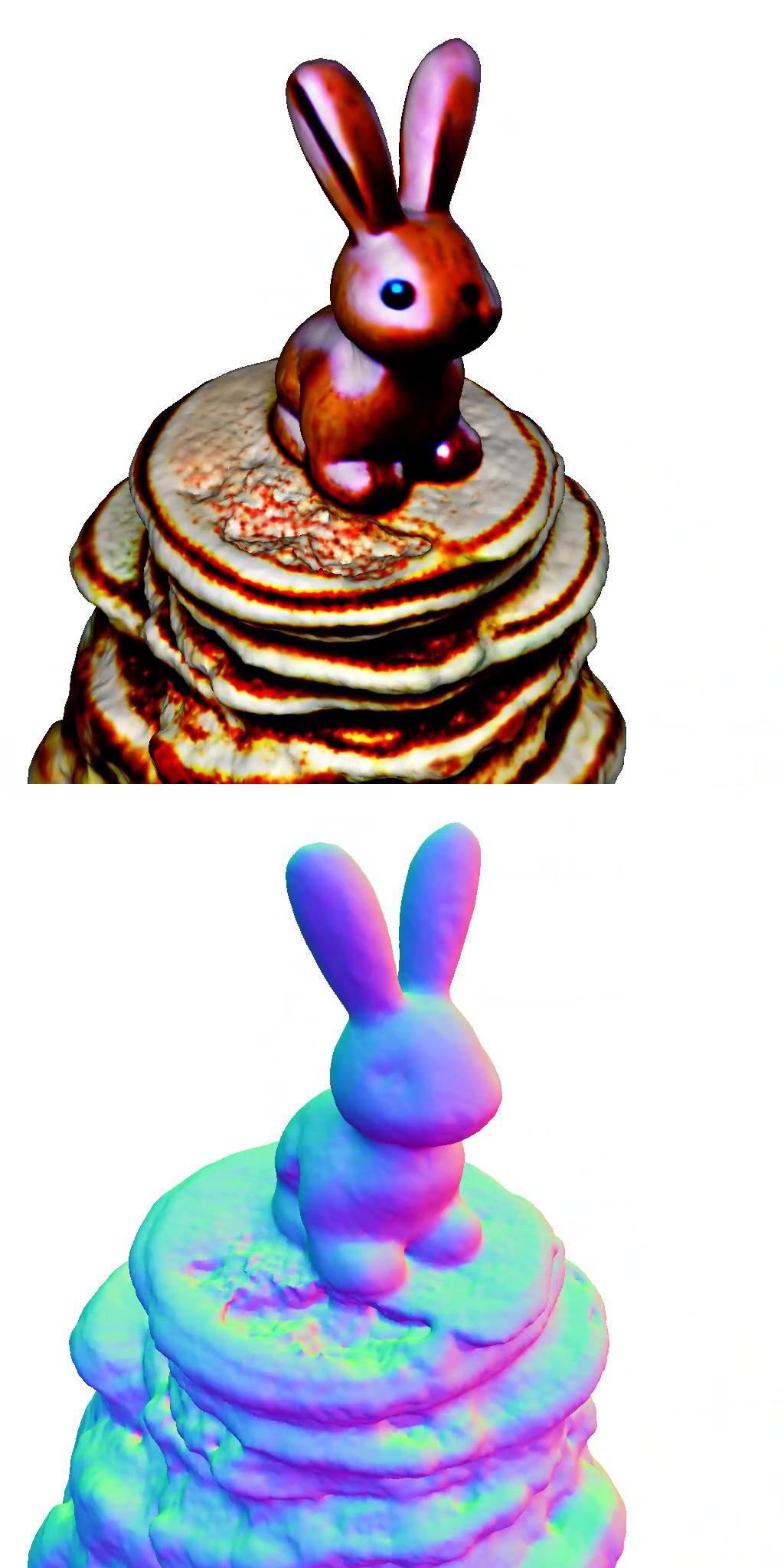}
    \\
     & \multicolumn{2}{c}{\footnotesize \textit{\textcolor{myorange}{lego bunny}, \textcolor{myblue}{a stack of pancakes}}} &
        \multicolumn{2}{c}{\footnotesize \textit{\textcolor{myorange}{stained glass bunny}, \textcolor{myblue}{a stack of pancakes}}} &
        \multicolumn{2}{c}{\footnotesize \textit{\textcolor{myorange}{wooden bunny}, \textcolor{myblue}{a stack of pancakes}}}  \\ 
    & \includegraphics[align=c, width=0.15\linewidth]{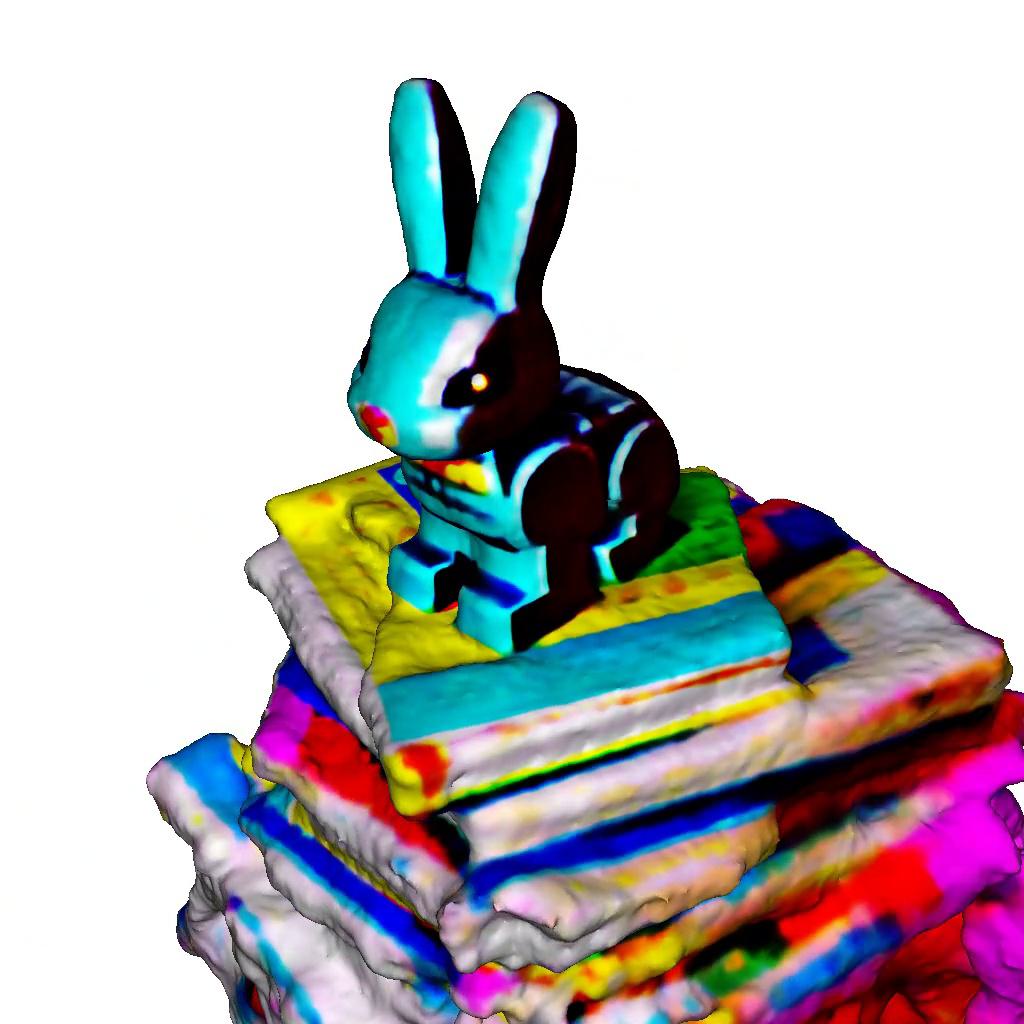} & 
        \includegraphics[align=c, width=0.15\linewidth]{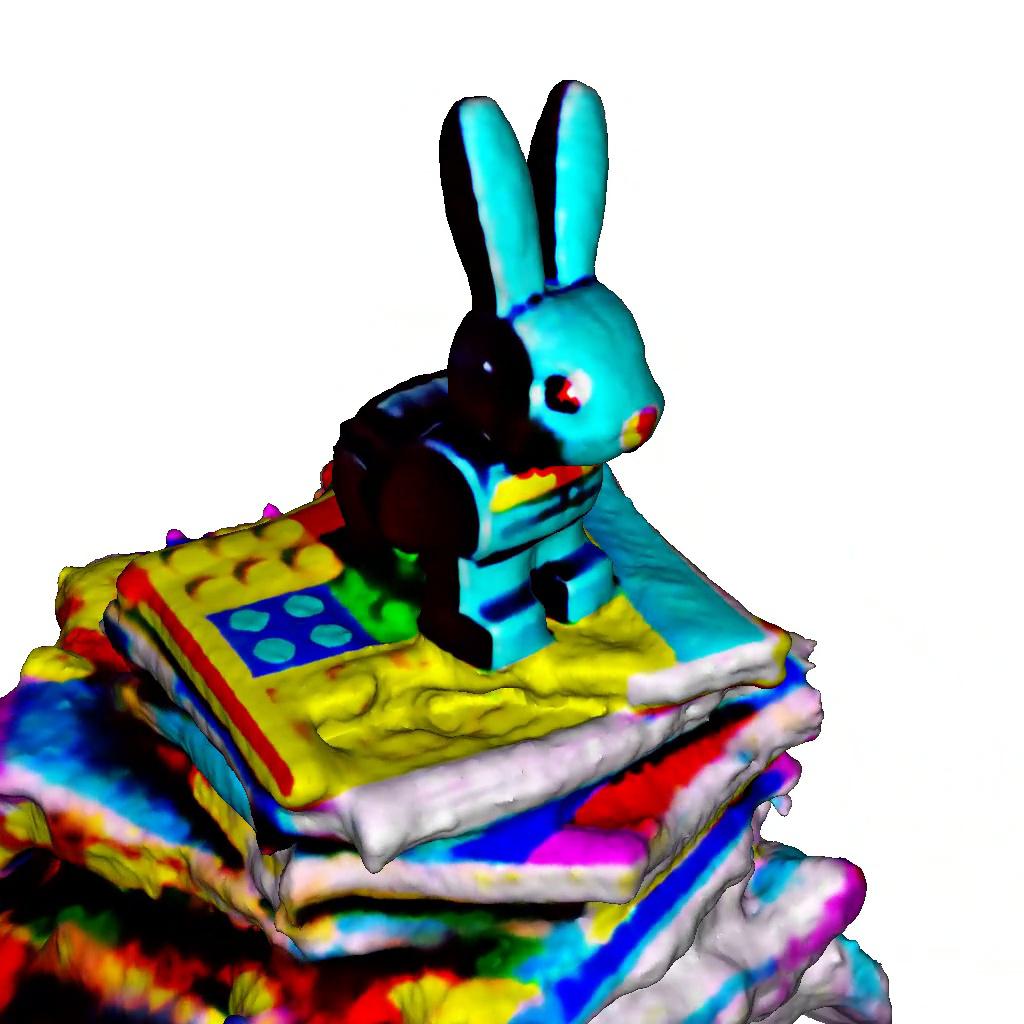} & 
        \includegraphics[align=c, width=0.15\linewidth]{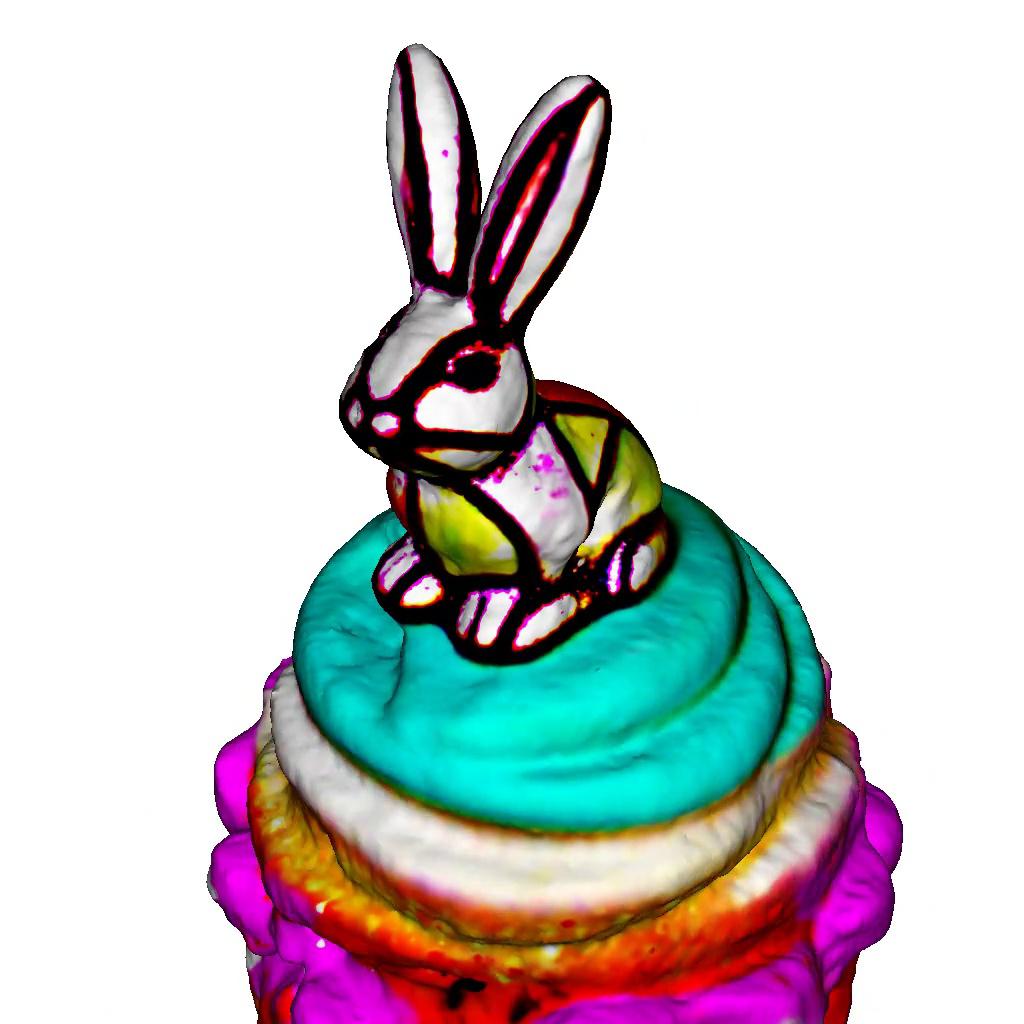} & 
        \includegraphics[align=c, width=0.15\linewidth]{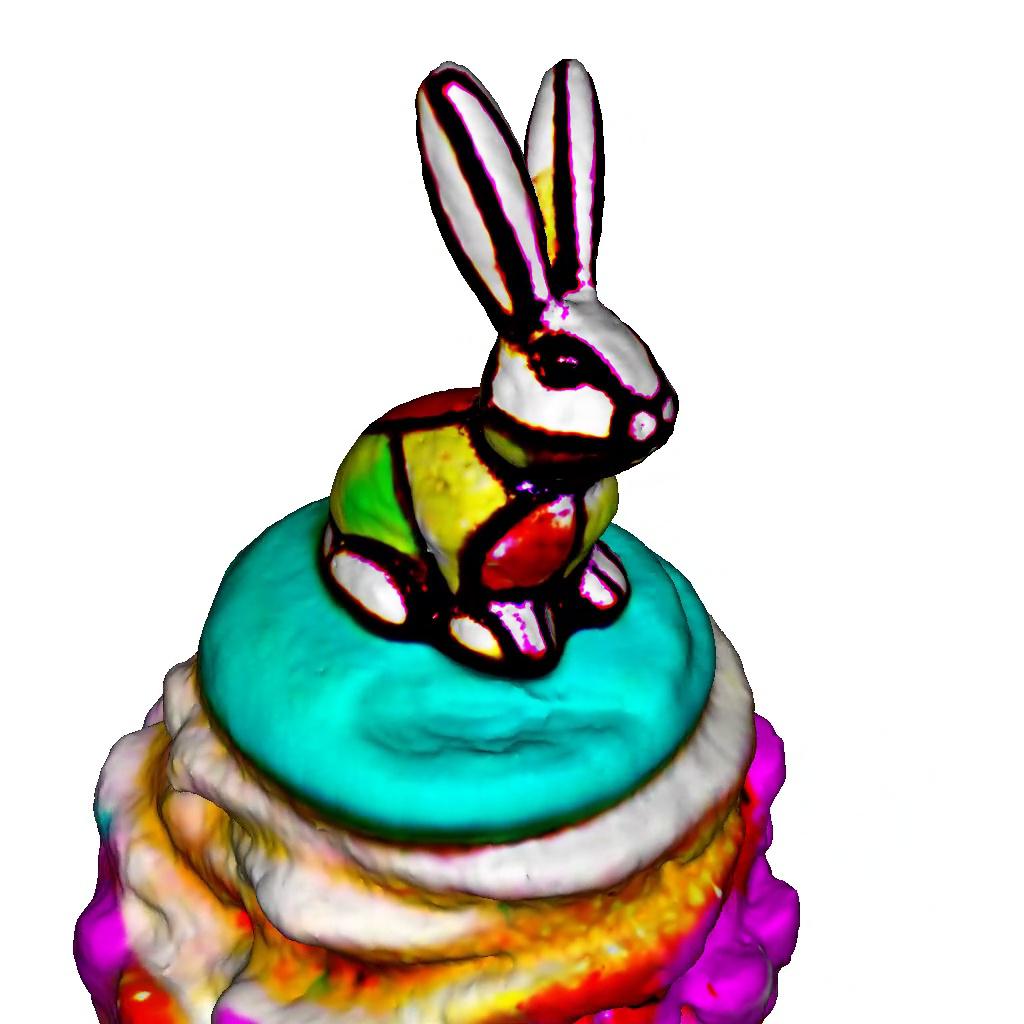} &
        \includegraphics[align=c, width=0.15\linewidth]{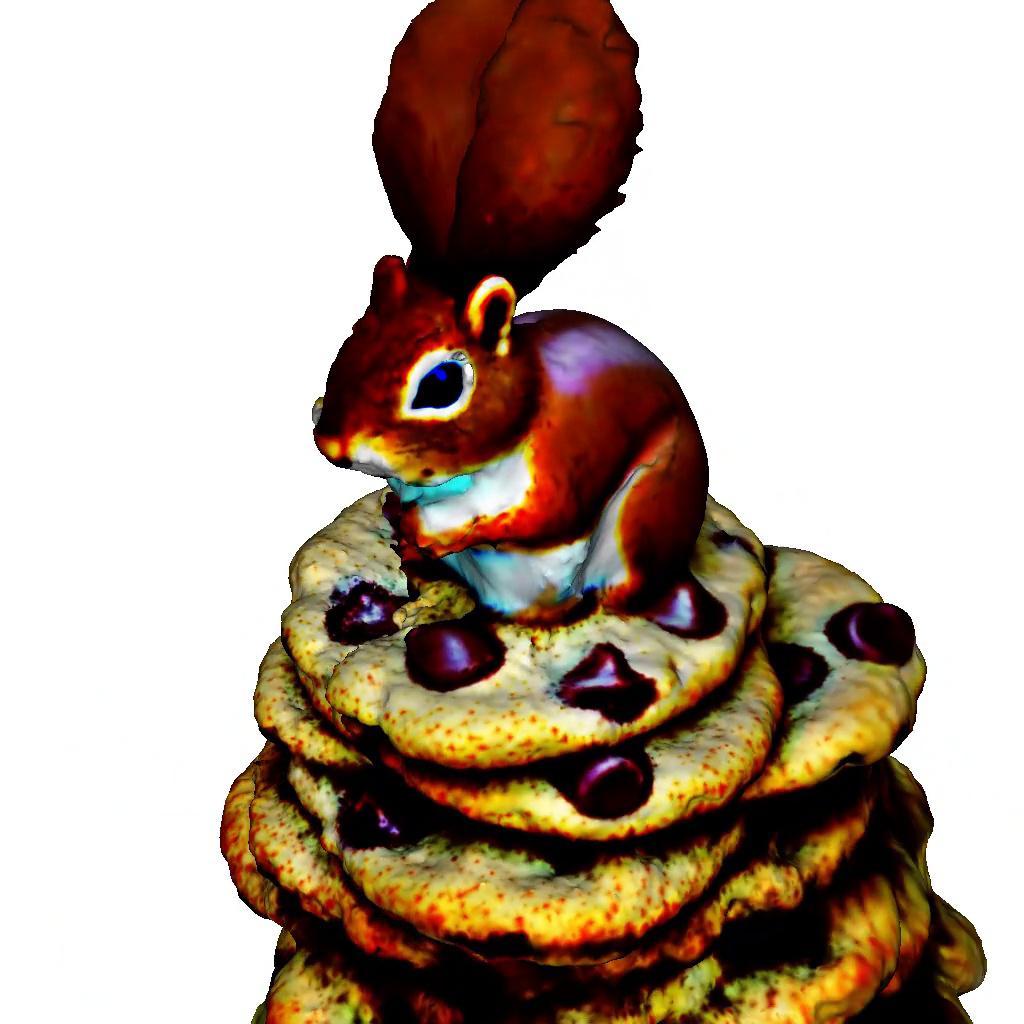} & 
        \includegraphics[align=c, width=0.15\linewidth]{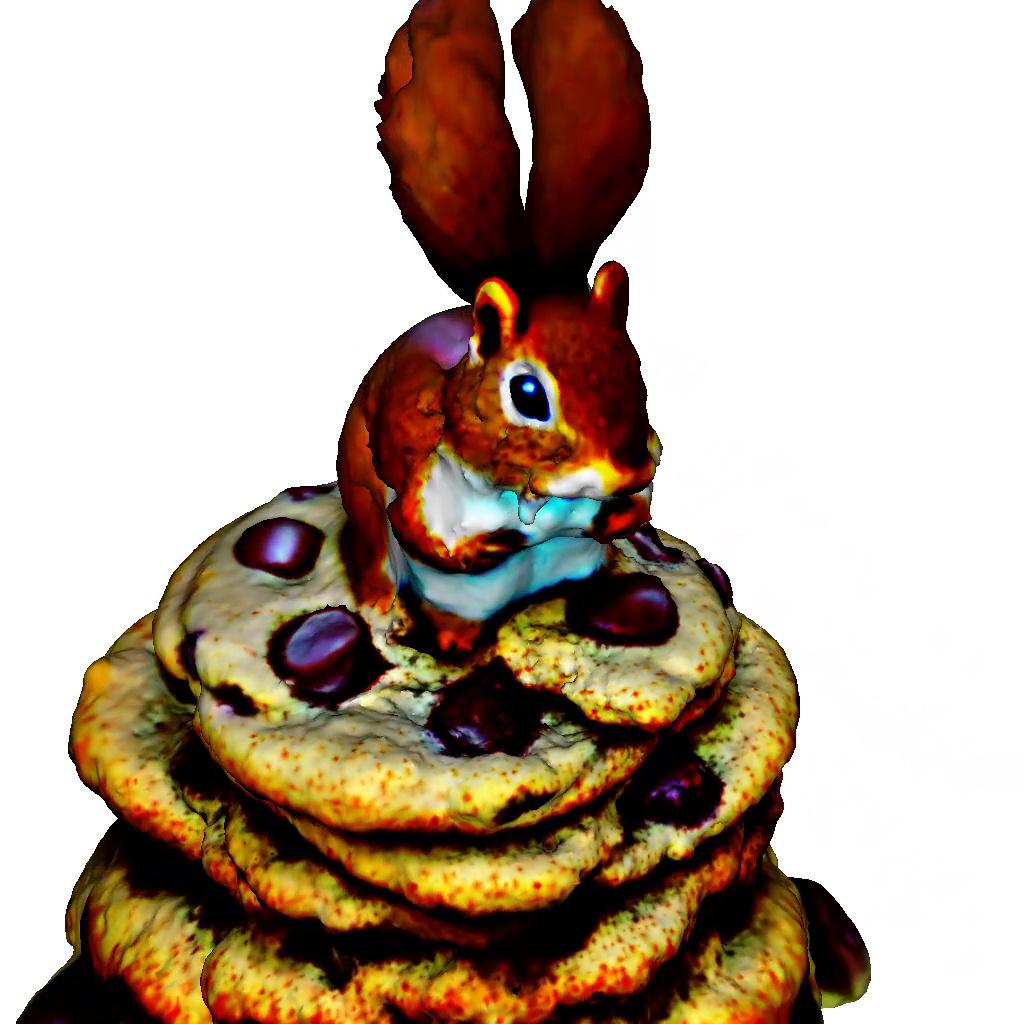} \\        
        & \multicolumn{2}{c}{\footnotesize \textit{\textcolor{myorange}{lego bunny}, \textcolor{myblue}{a stack of books}}} &
        \multicolumn{2}{c}{\footnotesize \textit{\textcolor{myorange}{stained glass bunny}, \textcolor{myblue}{a cupcake}}} &
        \multicolumn{2}{c}{\footnotesize \textit{\textcolor{myorange}{squirrel}, \textcolor{myblue}{a stack of chocolate cookies}}} 
        \\
    \includegraphics[align=c, width=0.1\linewidth, trim={0cm 0cm 0.3cm 0cm}, clip]{figures/prompt_edit/prompt_edit_base_sq.pdf}
       & \includegraphics[align=c, width=0.15\linewidth, trim={1.5cm 4cm 6cm 3.5cm}, clip]{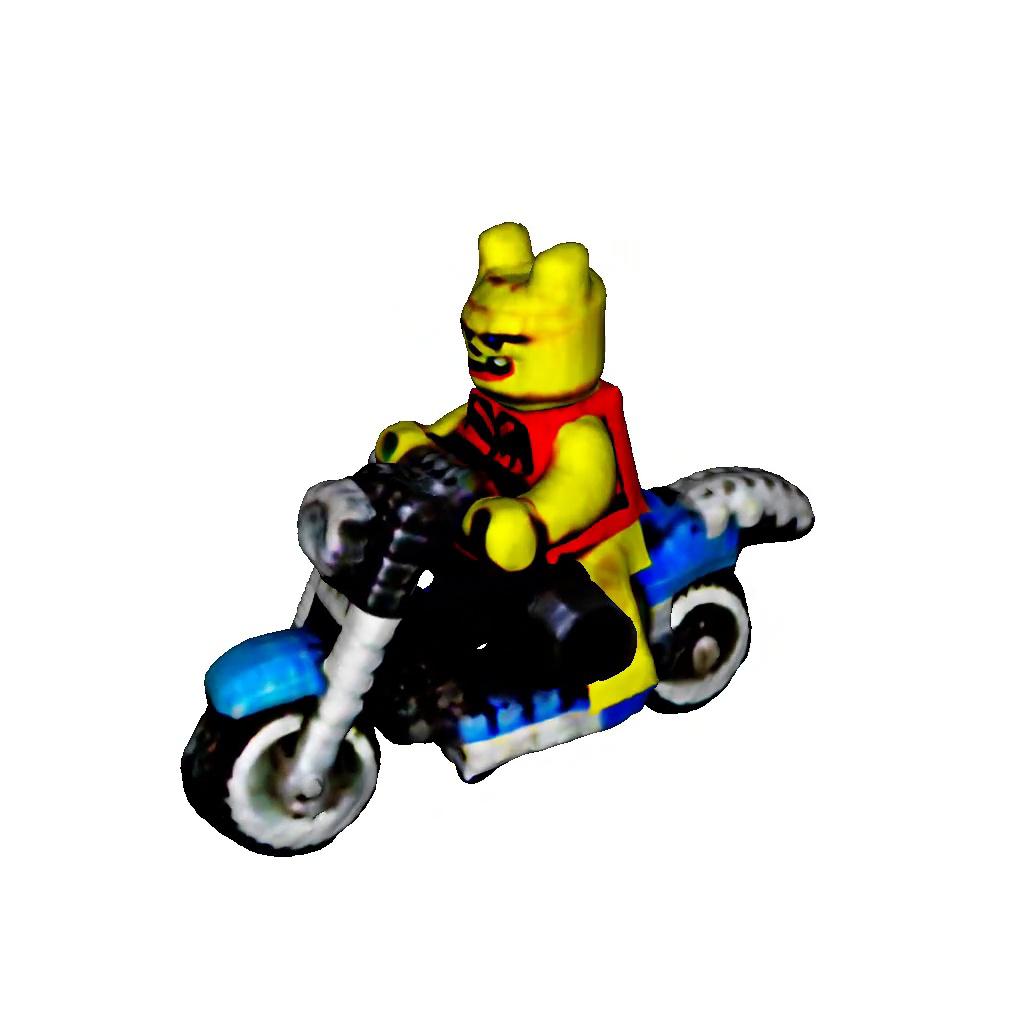} & 
        \includegraphics[align=c, width=0.15\linewidth, trim={3cm 1cm 2cm 4cm}, clip]{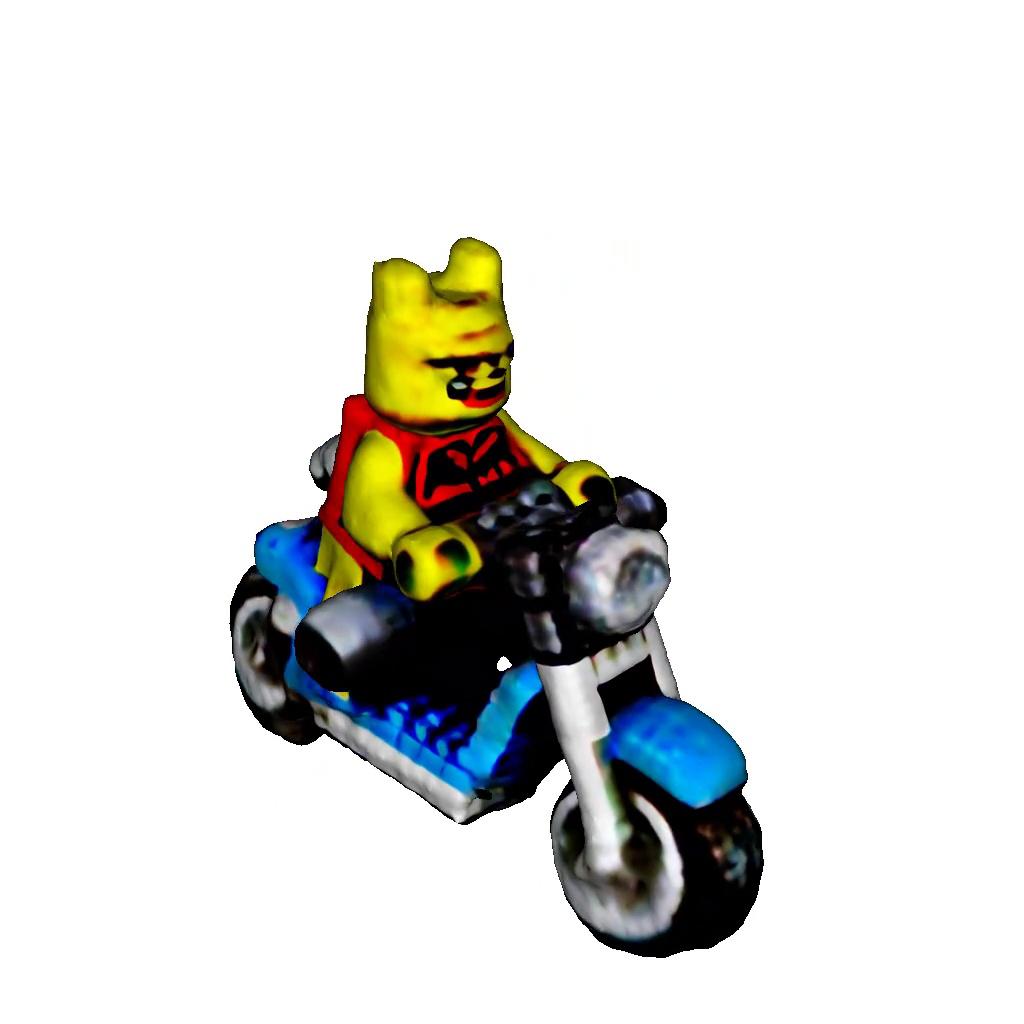} &
        \includegraphics[align=c, width=0.15\linewidth, trim={5cm 3cm 6cm 8cm}, clip]{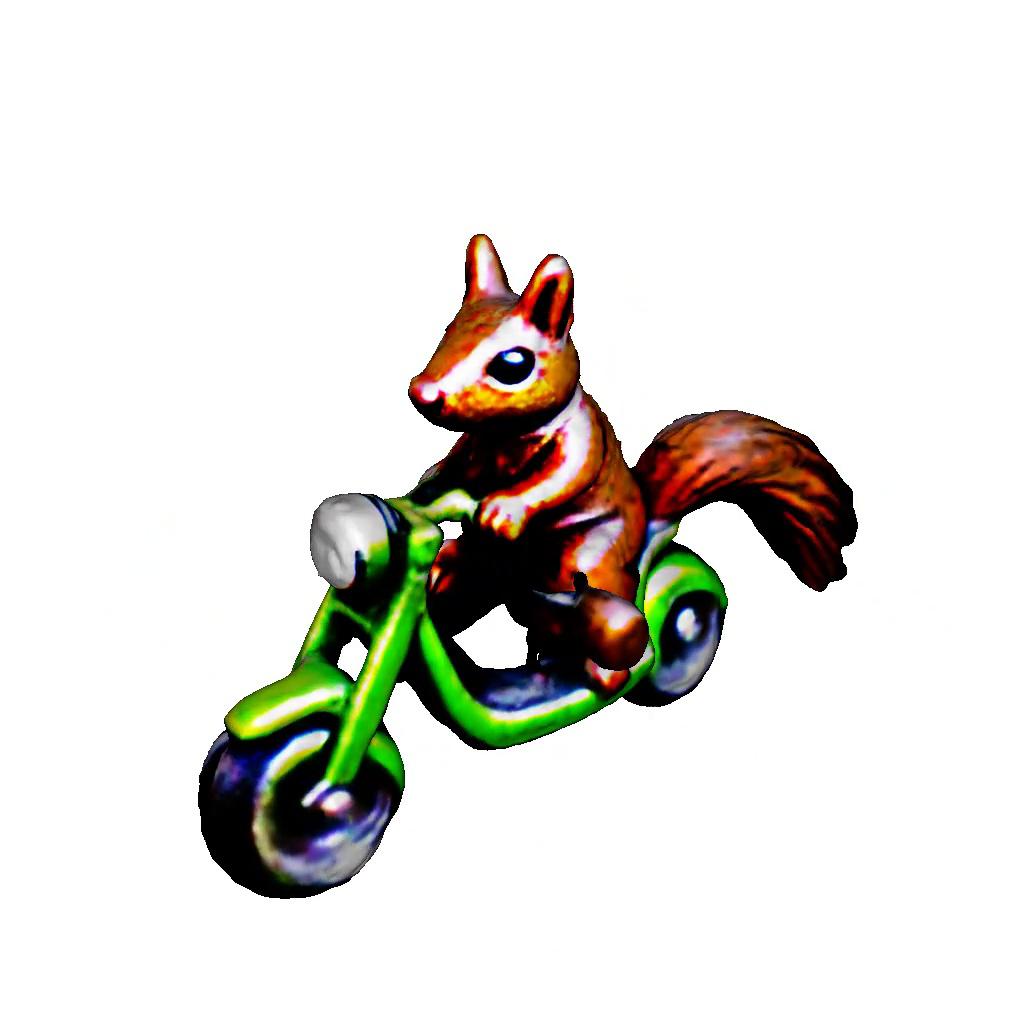} &  
        \includegraphics[align=c, width=0.15\linewidth, trim={5cm 3cm 6cm 8cm}, clip]{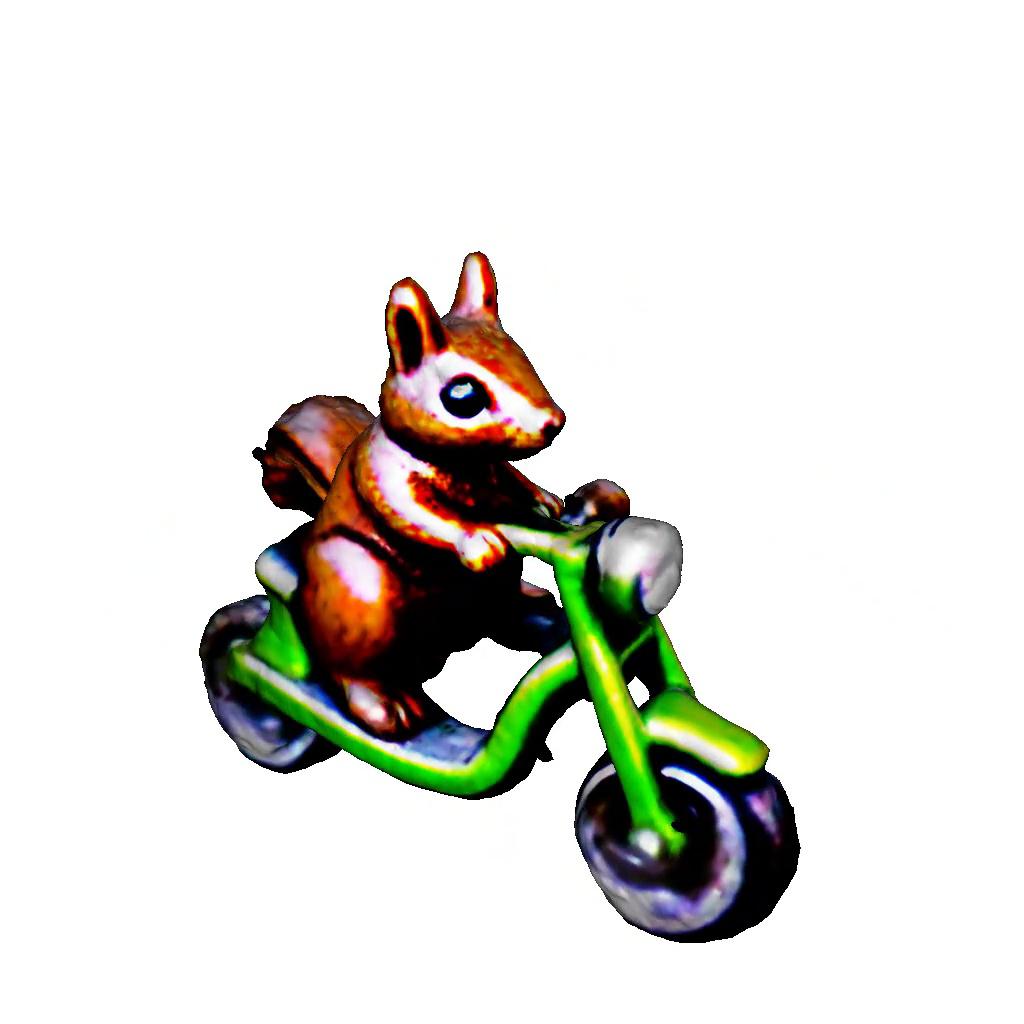} &
        \includegraphics[align=c, width=0.15\linewidth,trim={3cm 2cm 4cm 5cm}, clip]{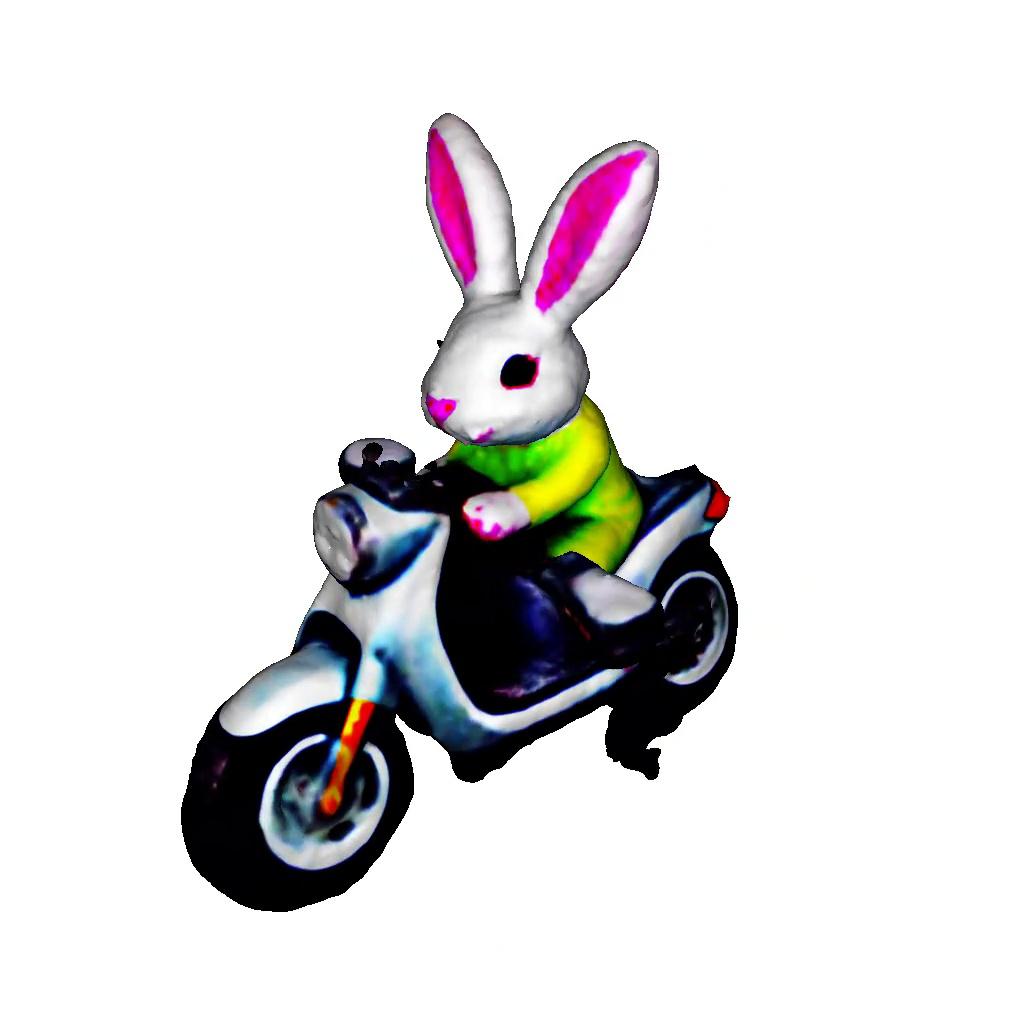} & 
        \includegraphics[align=c, width=0.15\linewidth,trim={3cm 2cm 4cm 5cm}, clip]{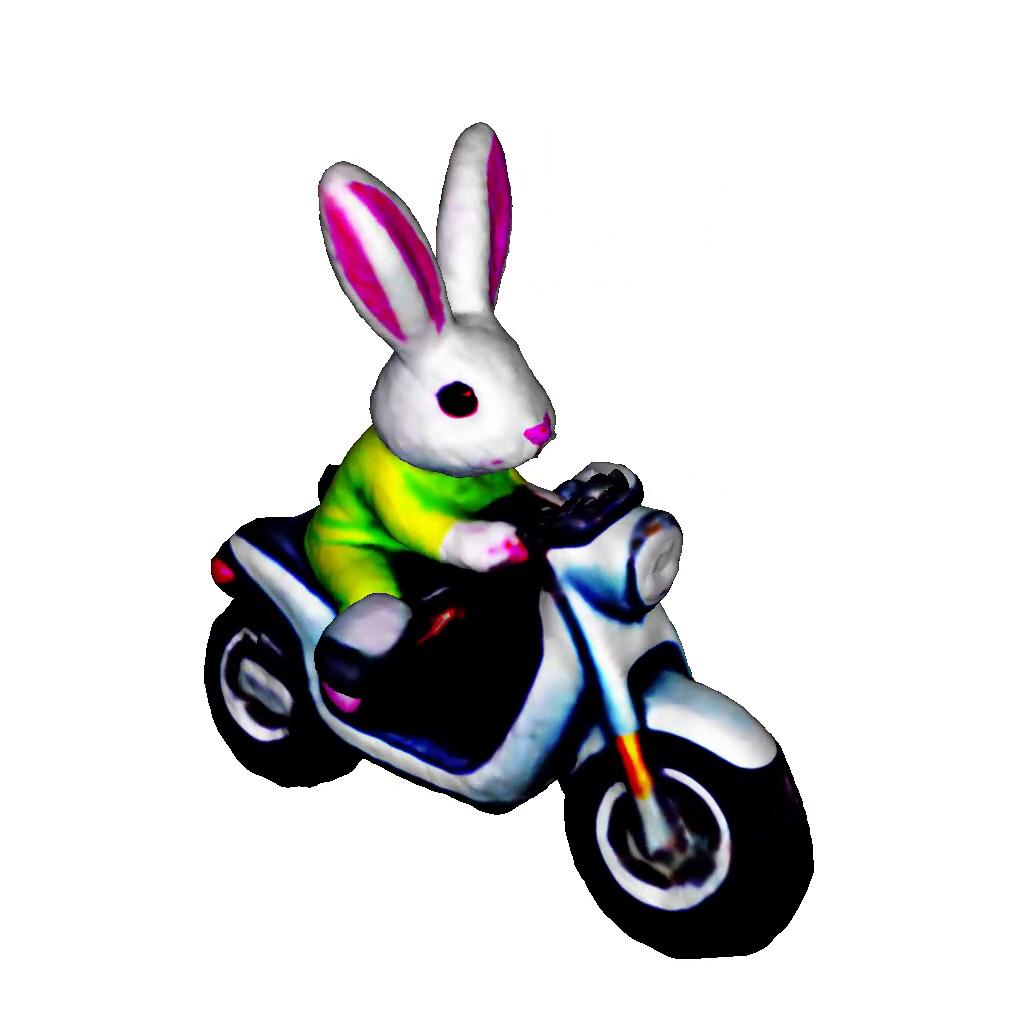} \\ 
        & \multicolumn{2}{c}{\footnotesize \textit{\textcolor{myorange}{lego squirrel}, \textcolor{myblue}{scooter}}} & 
        \multicolumn{2}{c}{\footnotesize \textit{\textcolor{myorange}{steampunk squirrel}, \textcolor{myblue}{scooter}}} & 
        \multicolumn{2}{c}{\footnotesize \textit{\textcolor{myorange}{bunny}, \textcolor{myblue}{scooter}}}  \\
        
        & 
        \includegraphics[align=c, width=0.15\linewidth, trim={1.5cm 4cm 6cm 3.5cm}, clip]{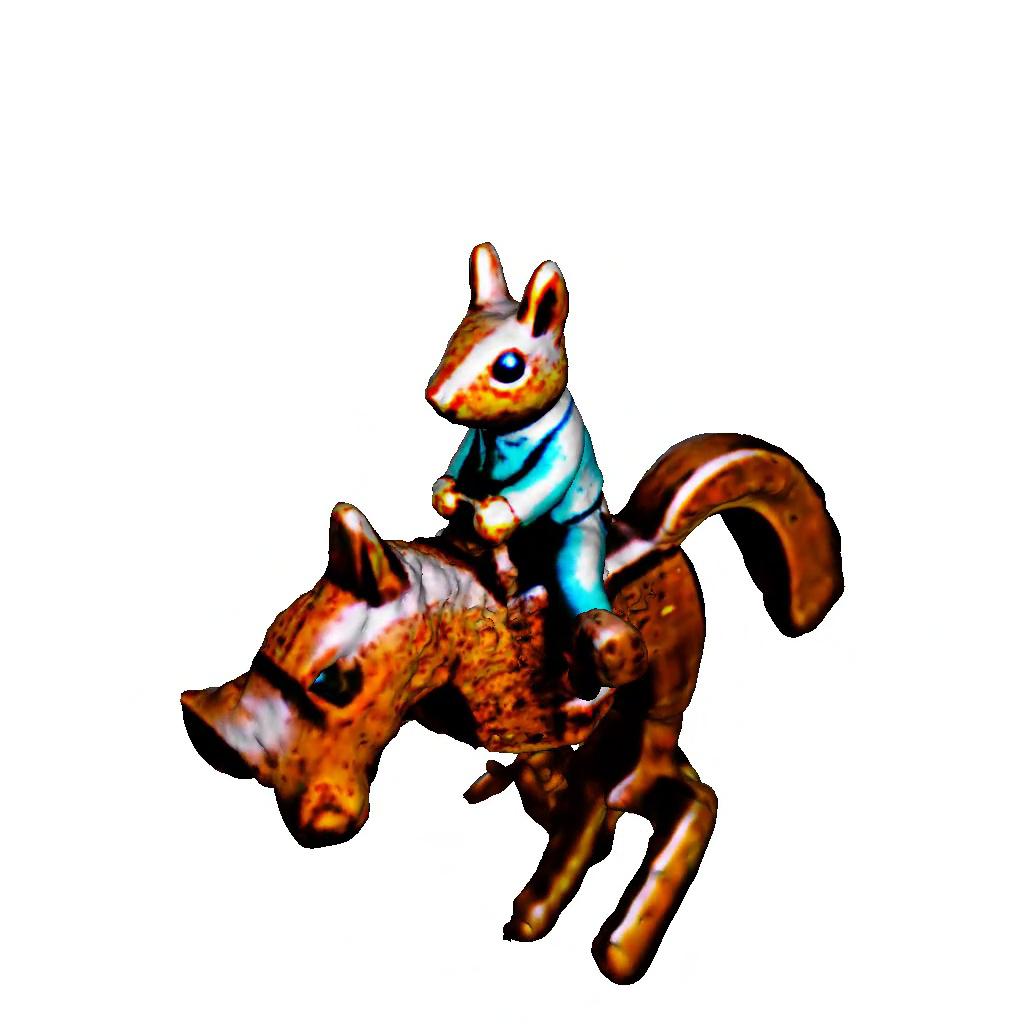} & 
        \includegraphics[align=c, width=0.15\linewidth, trim={3cm 1cm 2cm 4cm}, clip]{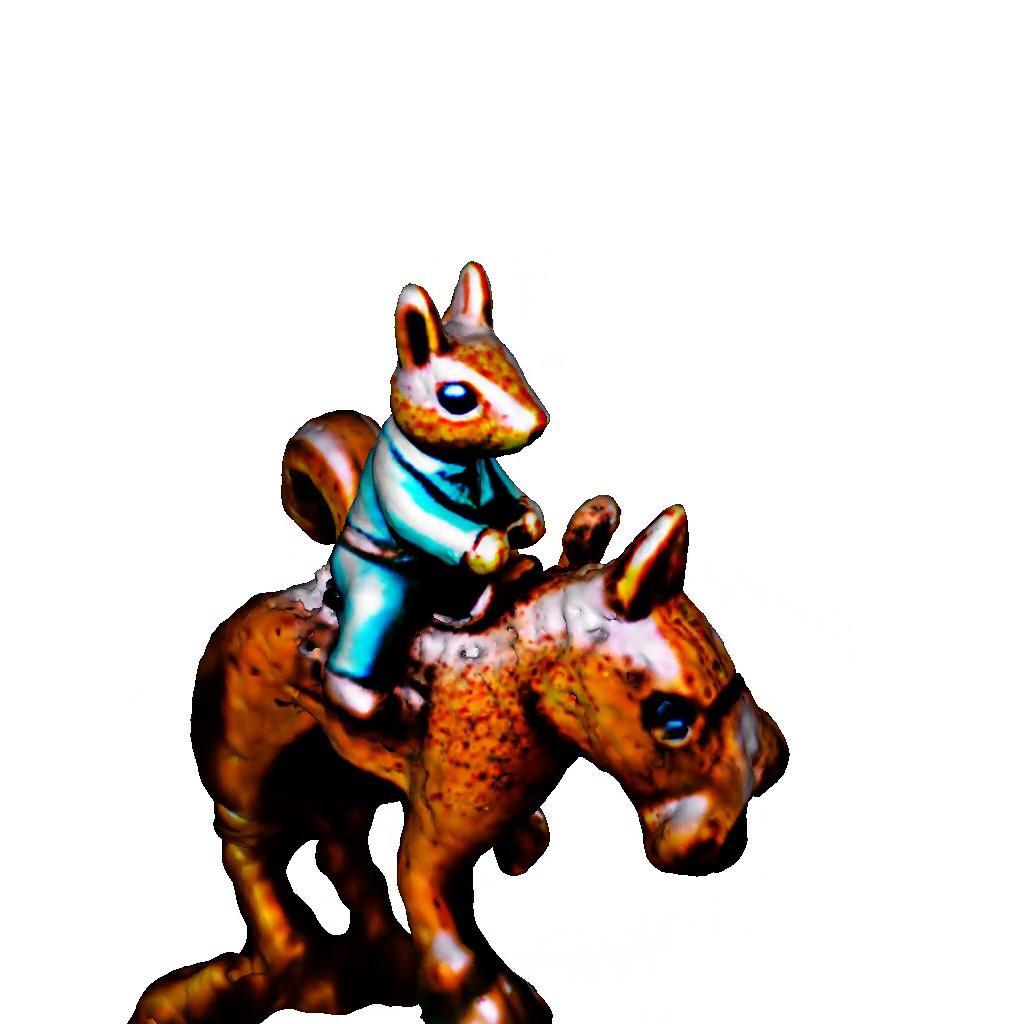} &

        \includegraphics[align=c, width=0.15\linewidth, trim={5cm 3cm 6cm 8cm}, clip]{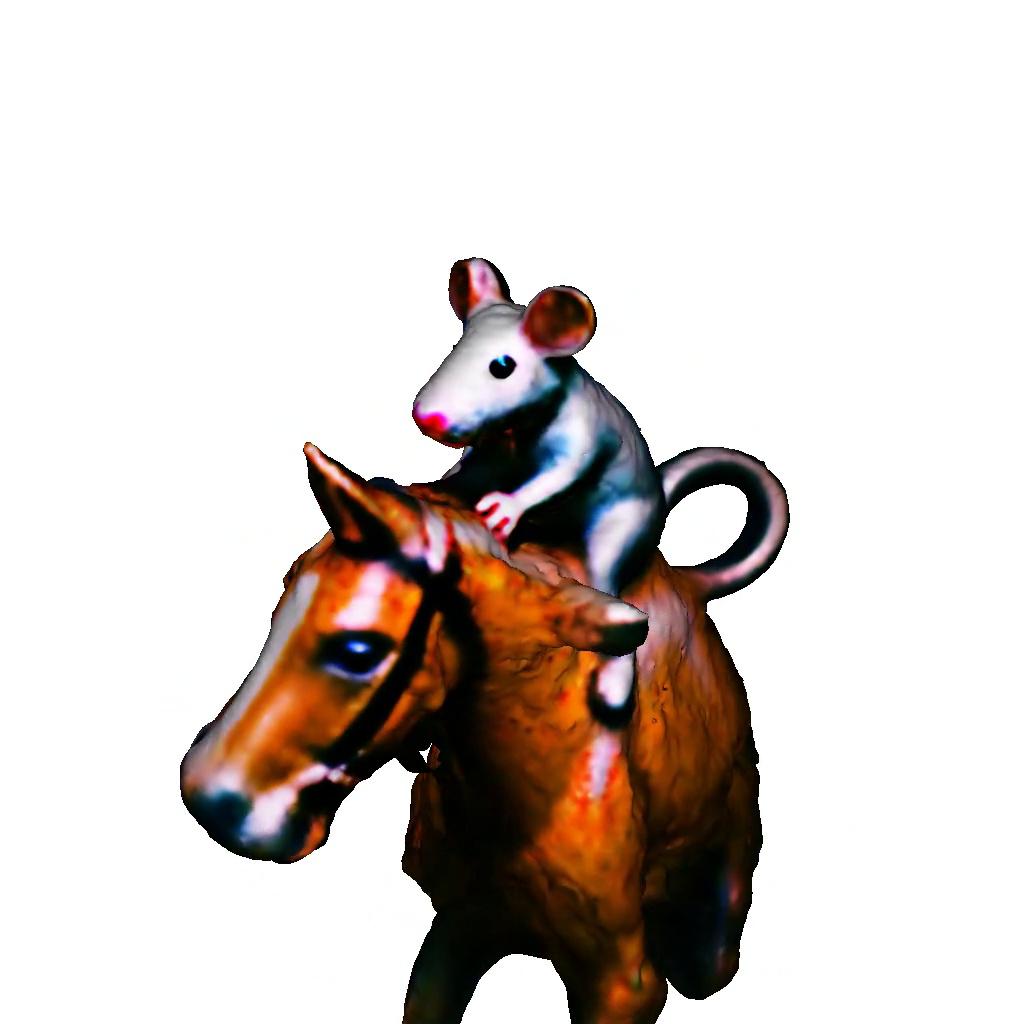} &  
        \includegraphics[align=c, width=0.15\linewidth, trim={5cm 3cm 6cm 8cm}, clip]{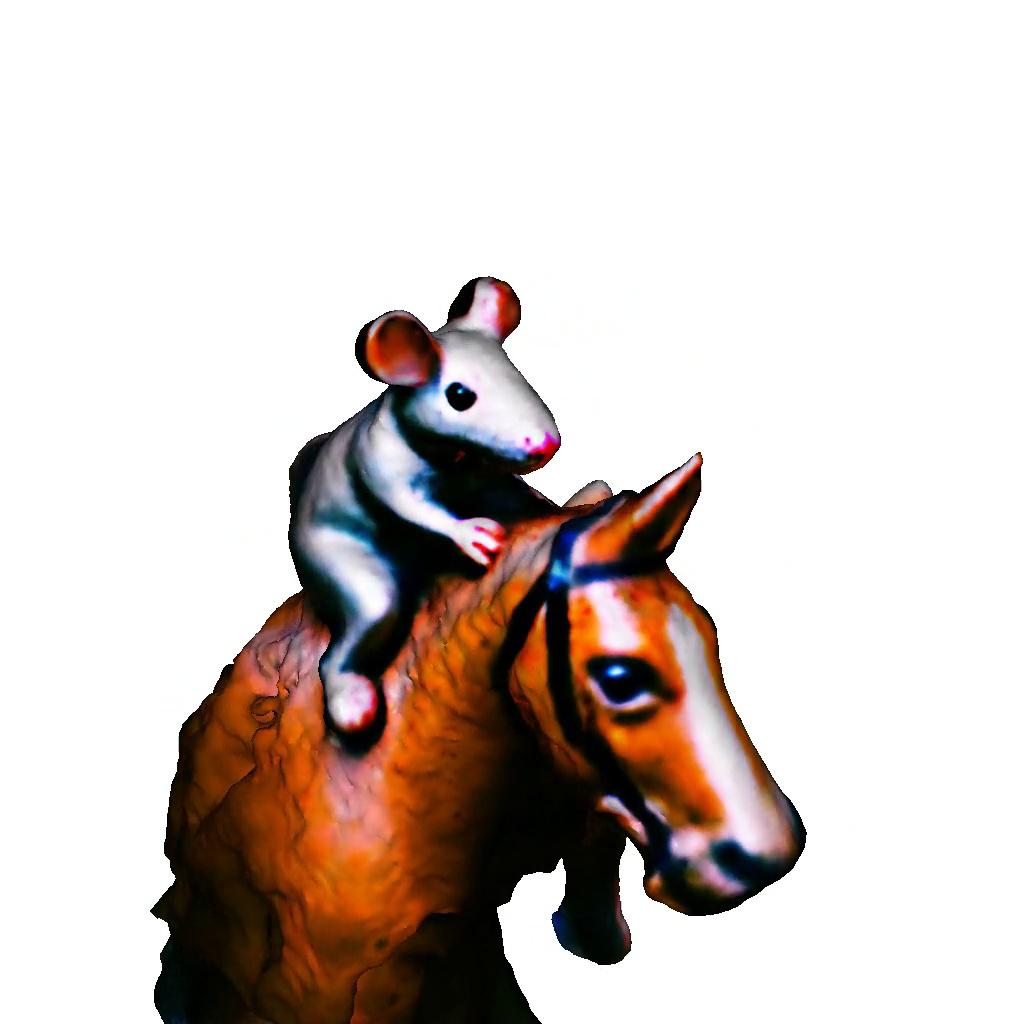} &
        \includegraphics[align=c, width=0.15\linewidth,trim={3cm 2cm 4cm 5cm}, clip]{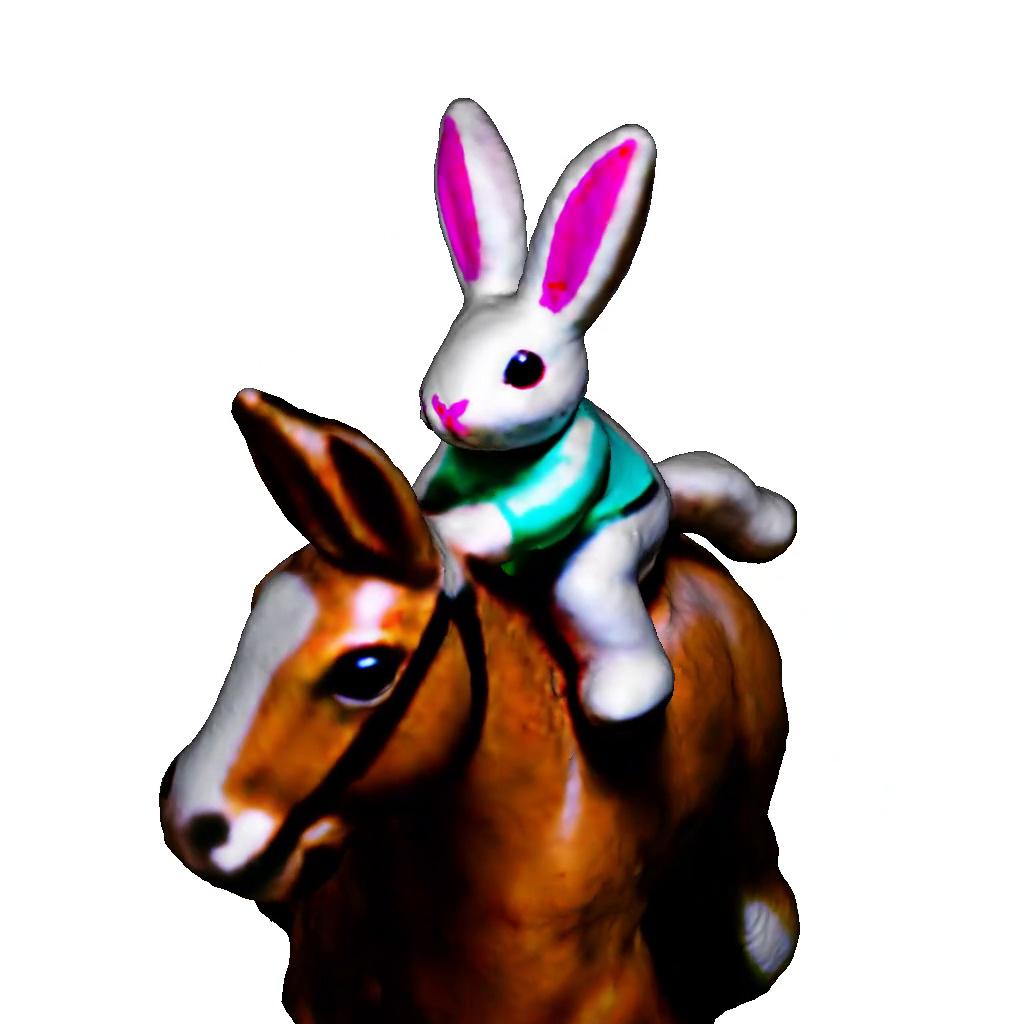} & 
        \includegraphics[align=c, width=0.15\linewidth,trim={3cm 2cm 4cm 5cm}, clip]{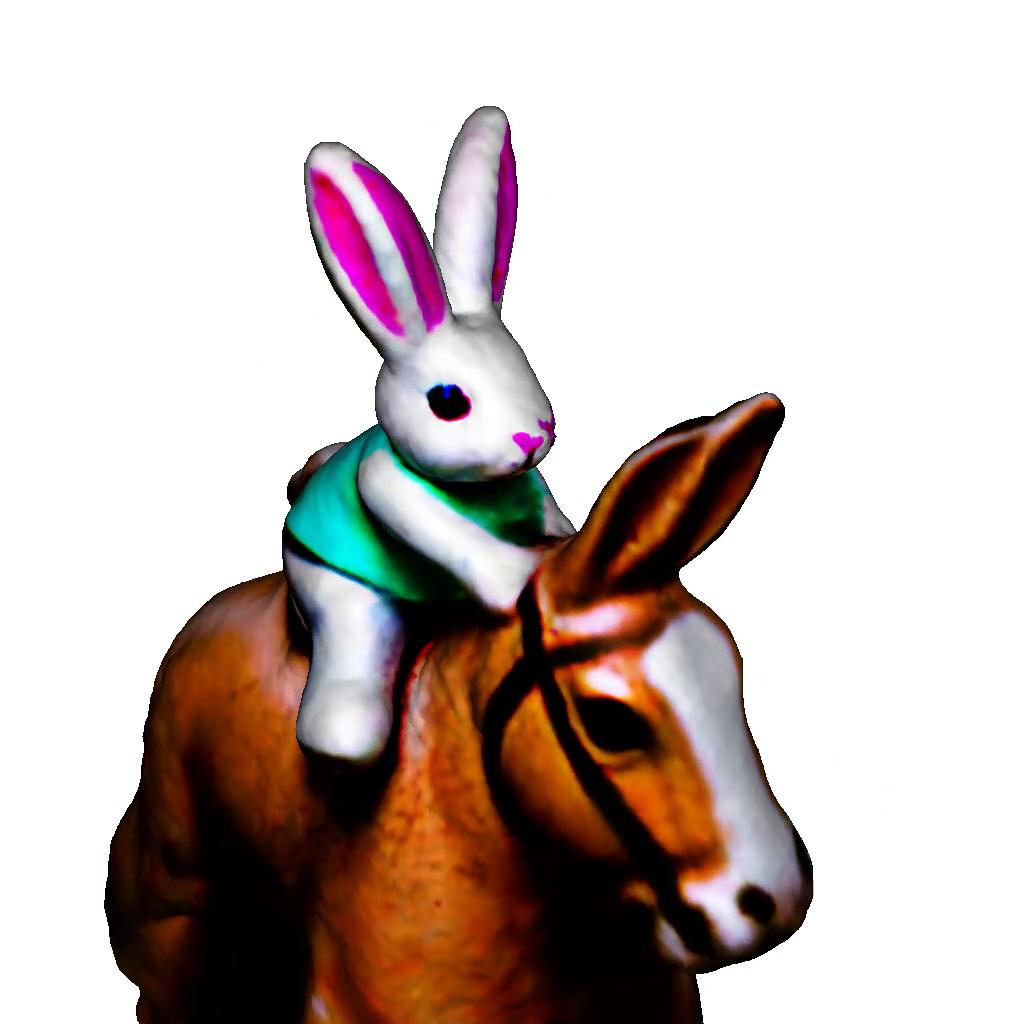} \\ 
        & 
        \multicolumn{2}{c}{\footnotesize \textit{\textcolor{myorange}{steampunk squirrel}, \textcolor{myblue}{horse}}} & 

        \multicolumn{2}{c}{\footnotesize \textit{\textcolor{myorange}{rat}, \textcolor{myblue}{horse}}} & 
        \multicolumn{2}{c}{\footnotesize \textit{\textcolor{myorange}{bunny}, \textcolor{myblue}{horse}}}  \\
        
        & \includegraphics[align=c, width=0.15\linewidth, trim={1.5cm 4cm 6cm 3.5cm}, clip]{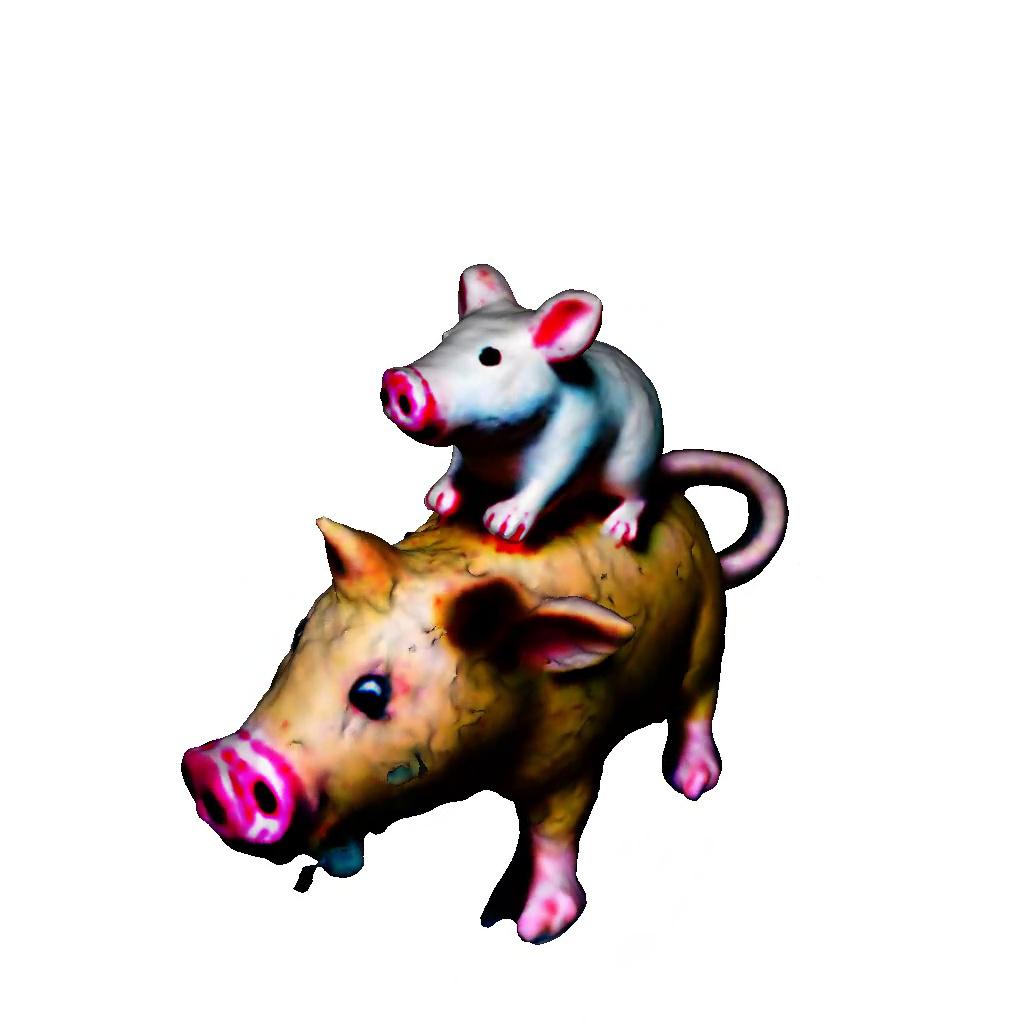} & 
        \includegraphics[align=c, width=0.15\linewidth, trim={3cm 1cm 2cm 4cm}, clip]{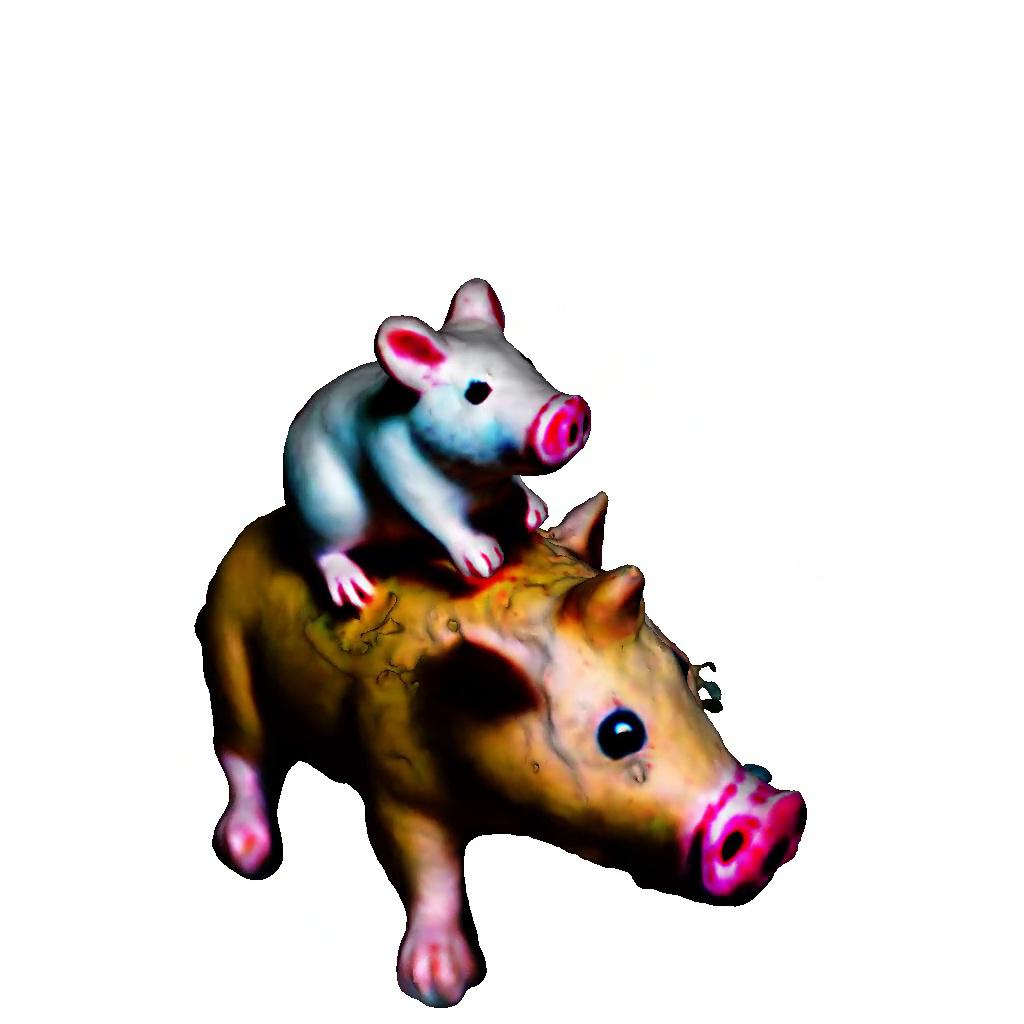} &
        \includegraphics[align=c, width=0.15\linewidth, trim={5cm 3cm 6cm 8cm}, clip]{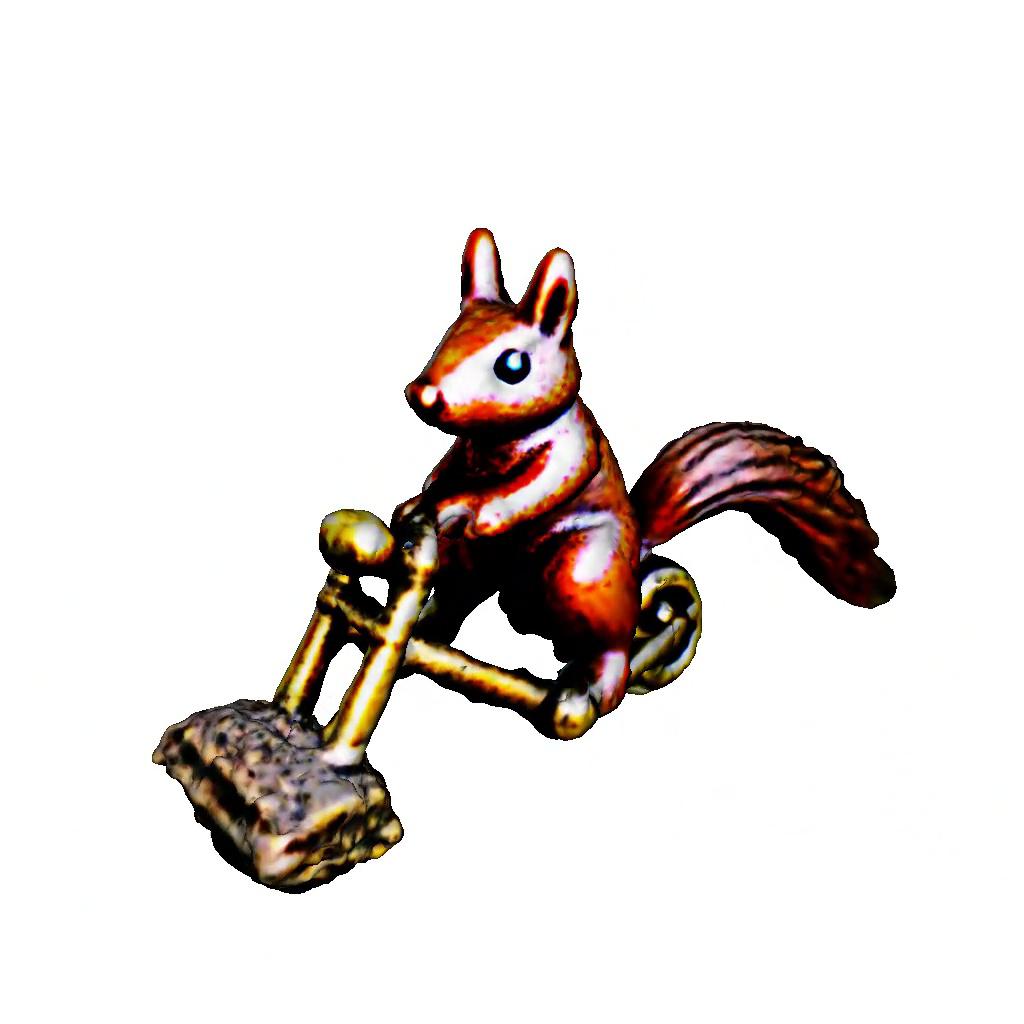} &  
        \includegraphics[align=c, width=0.15\linewidth, trim={5cm 3cm 6cm 8cm}, clip]{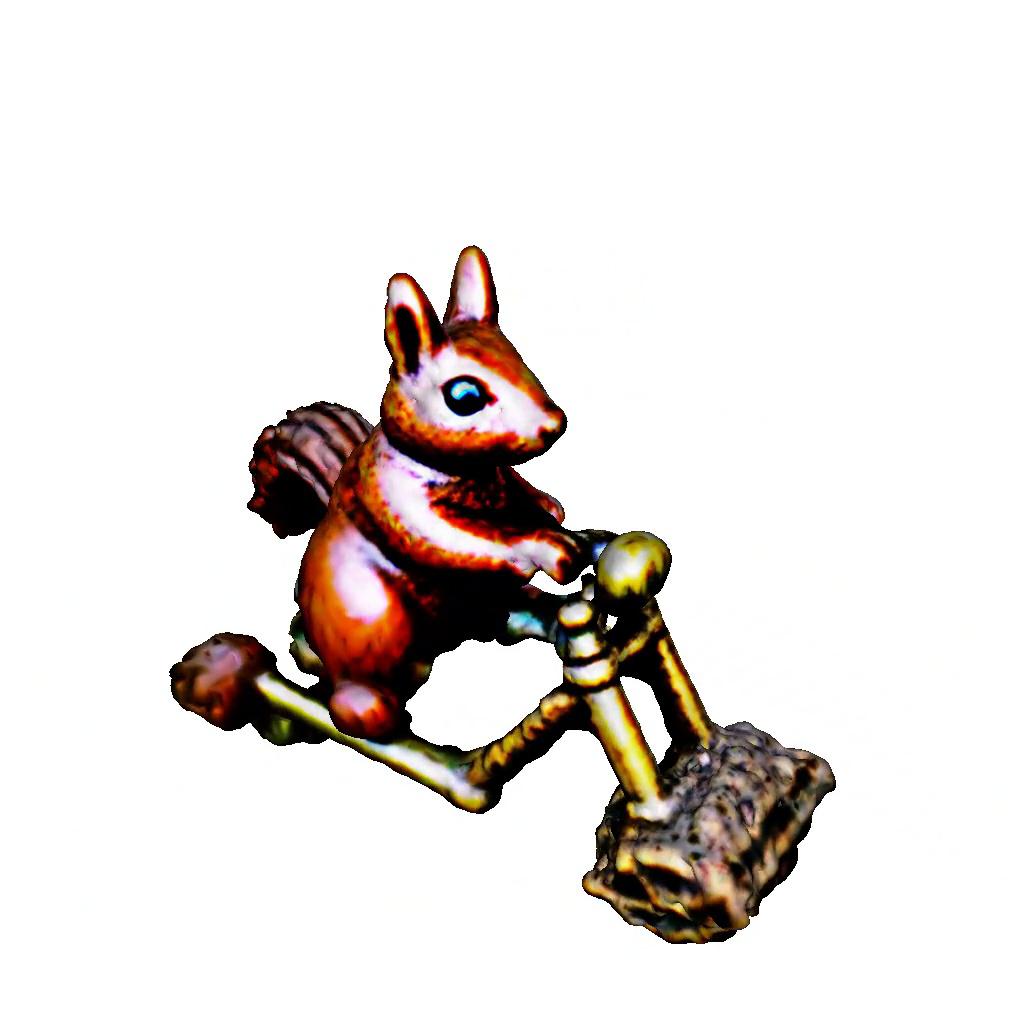} &
        \includegraphics[align=c, width=0.15\linewidth,trim={3cm 2cm 4cm 5cm}, clip]{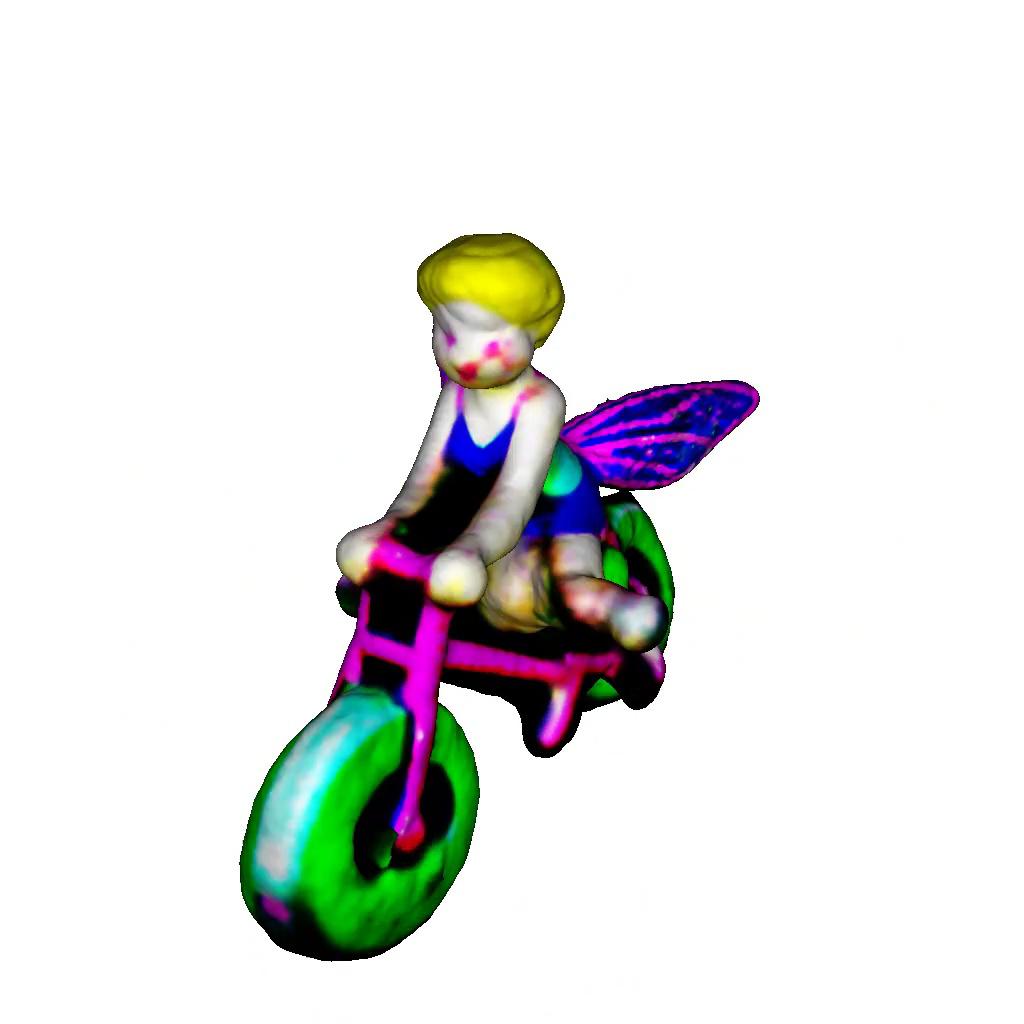} & 
        \includegraphics[align=c, width=0.15\linewidth,trim={3cm 2cm 4cm 5cm}, clip]{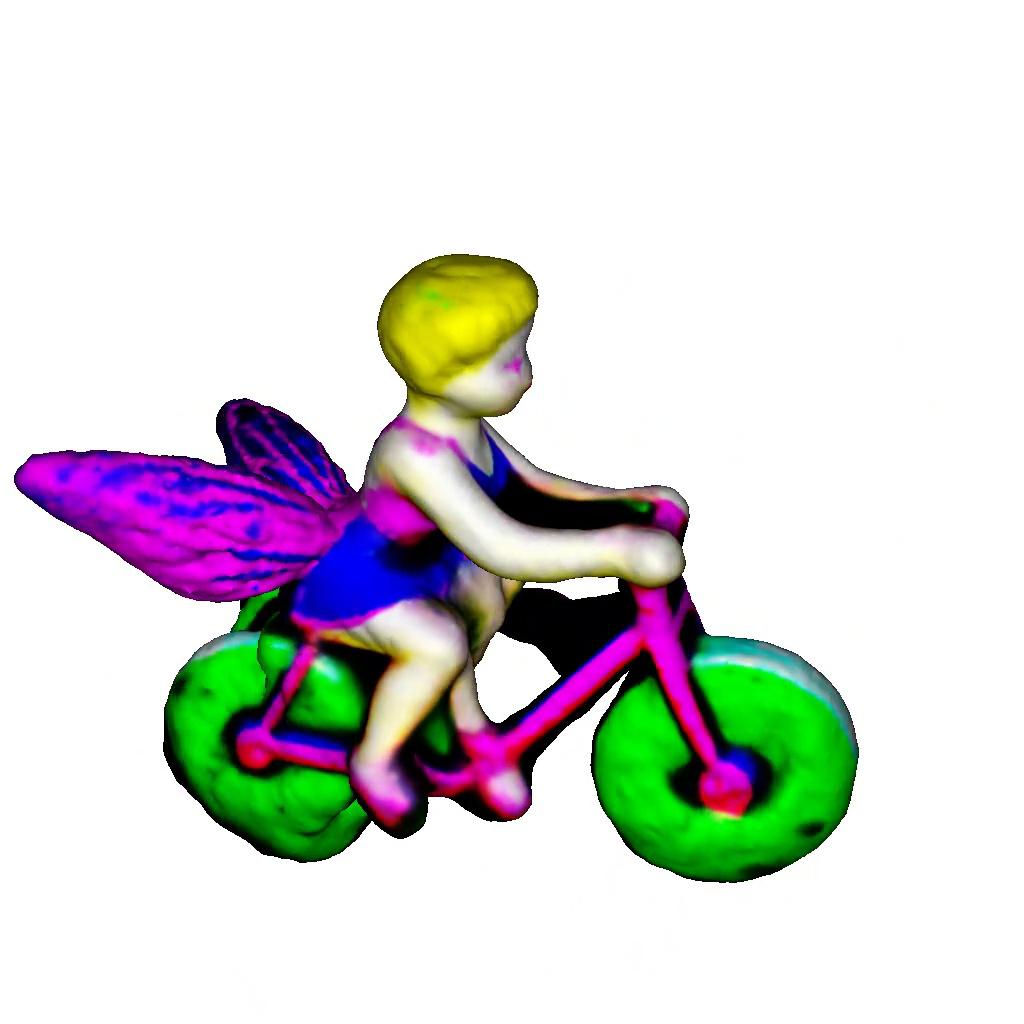} \\ 
        & 
        \multicolumn{2}{c}{\footnotesize \textit{\textcolor{myorange}{rat}, \textcolor{myblue}{pig}}} & 

        \multicolumn{2}{c}{\footnotesize \textit{\textcolor{myorange}{squirrel}, \textcolor{myblue}{broomstick}}} & 
        \multicolumn{2}{c}{\footnotesize \textit{\textcolor{myorange}{fairy}, \textcolor{myblue}{bike}}}  \\
        
    \end{tabular}
   \caption{\textbf{\name with prompt-based editing}. Given a coarse model (first column) generated with a base prompt, we replace the underscored text with new text and fine-tune the NeRF to get a high-resolution NeRF model with LDM. We further fine-tune the high-resolution mesh on top of the NeRF model. Such a prompt-based editing technique gives artists better control over the 3D generation output.}
   \label{fig:sup-fine-tuneSD-prompt}
\end{figure*}
\begin{figure*}
    \centering 
    \setlength{\tabcolsep}{0.2pt}
    \begin{tabular}{c} 
      \begin{tabularx}{\textwidth}{ >{\centering\arraybackslash}X >{\centering\arraybackslash}X}
        Ours & DreamFusion~\cite{poole2022dreamfusion}
    \end{tabularx}\\
        \includegraphics[align=c, width=\linewidth]{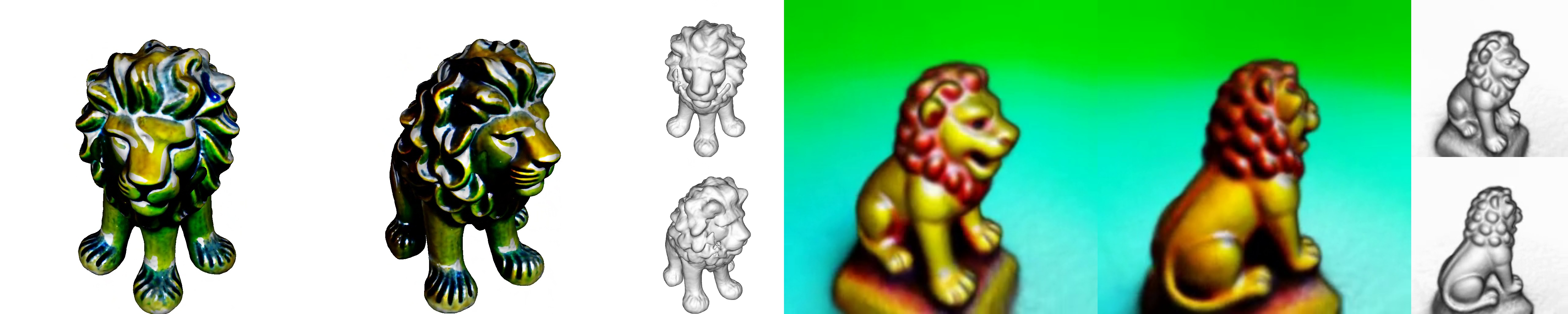} \\
         { \textit{a ceramic lion}}  \\
         \includegraphics[align=c, width=\linewidth]{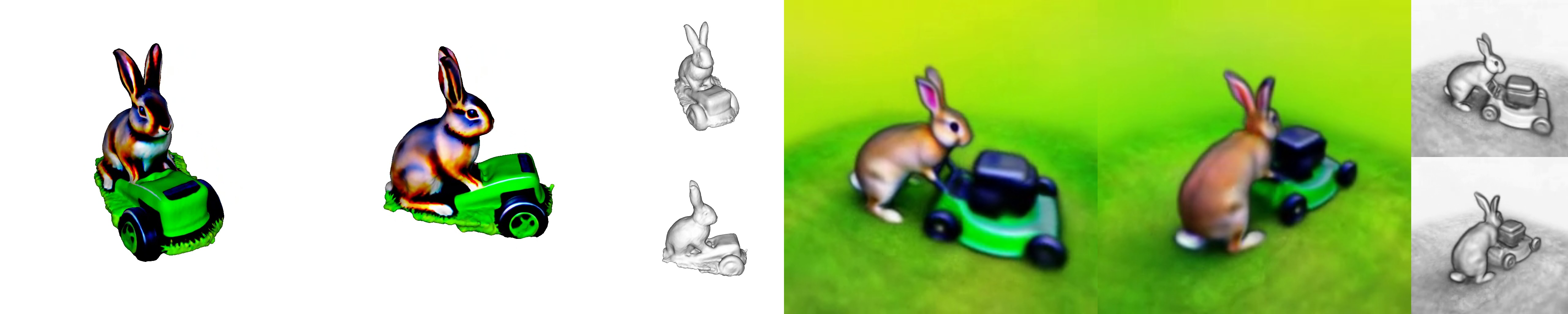}  \\
         { \textit{a rabbit cutting grass with a lawnmower$^{\dagger}$}}  \\
        \includegraphics[align=c, width=\linewidth]{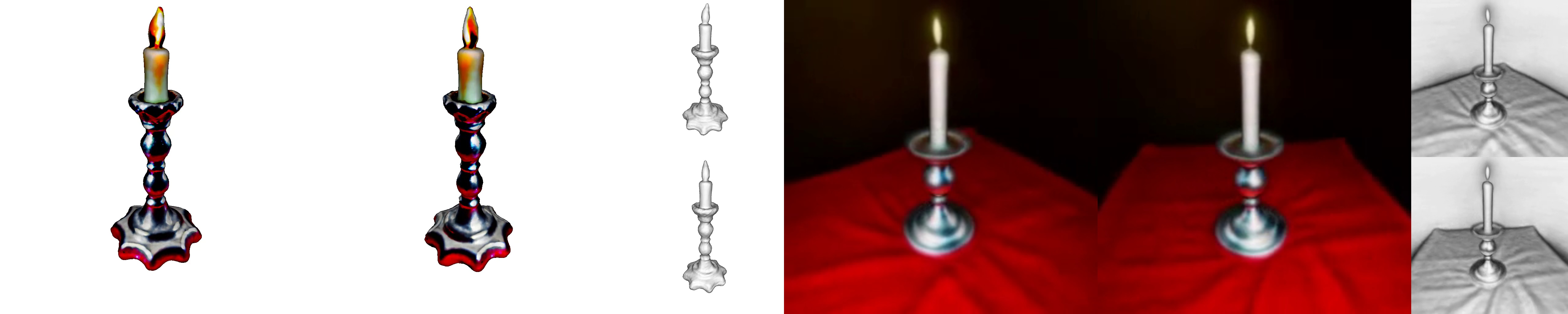} \\
         { \textit{ a silver candelabra sitting on a red velvet tablecloth, only one candle is lit$^{\dagger}$}}  \\
        \includegraphics[align=c, width=\linewidth]{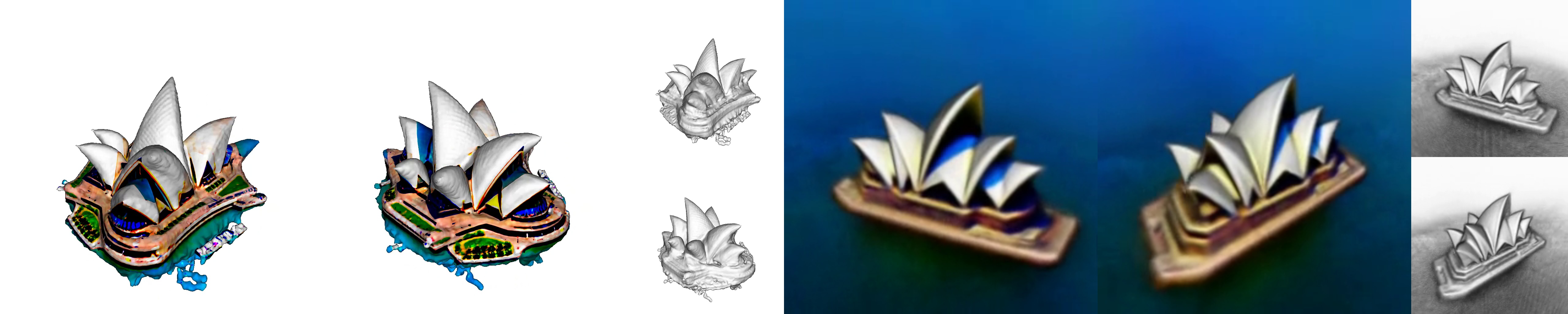} \\
         { \textit{Sydney opera house, aerial view$^{\dagger}$}} 
         \end{tabular}
        \vspace{-2mm}
    \caption{\textbf{Qualitative comparison with DreamFusion~\cite{poole2022dreamfusion}.} We use the same text prompt as in DreamFusion. For each 3D model, we render it from two views with a textureless rendering for each view and remove the background to focus on the 3D shape. For the DreamFusion results, we take frames from the videos published on the official webpage. \name generates much higher quality 3D shapes on both geometry and texture compared with DreamFusion.  
\textit{$\ast$ a DSLR photo of... $\dagger$ a zoomed out DSLR photo of...} }
    \label{fig:suppl_compare_dreamfusion_1}
    \vspace{-2mm}
\end{figure*}

\begin{figure*}
    \centering 
    \setlength{\tabcolsep}{0.2pt}
    \begin{tabular}{c} 
      \begin{tabularx}{\textwidth}{ >{\centering\arraybackslash}X >{\centering\arraybackslash}X}
        Ours & DreamFusion~\cite{poole2022dreamfusion}
    \end{tabularx}\\
         \includegraphics[align=c, width=\linewidth]{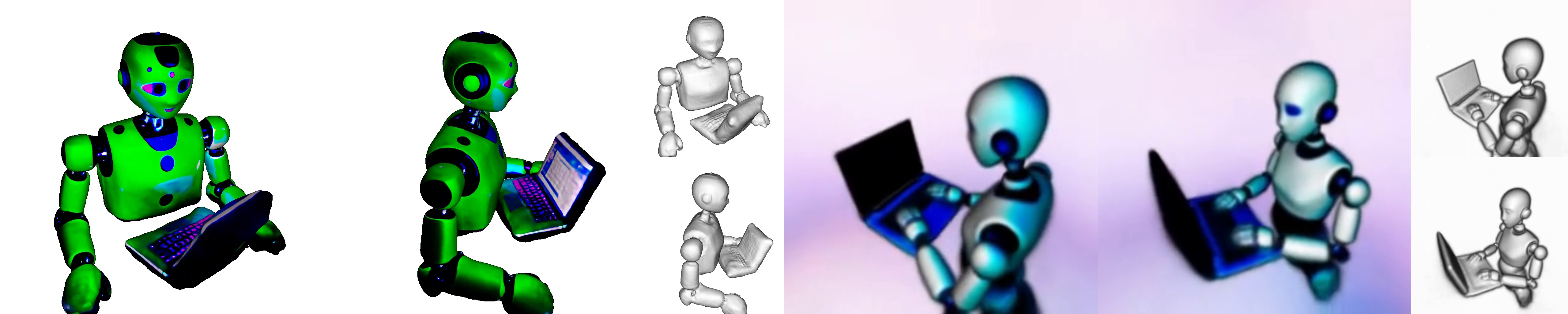} \\
        {\textit{a humanoid robot using a laptop$^{\ast}$}} \\
        \includegraphics[align=c, width=\linewidth]{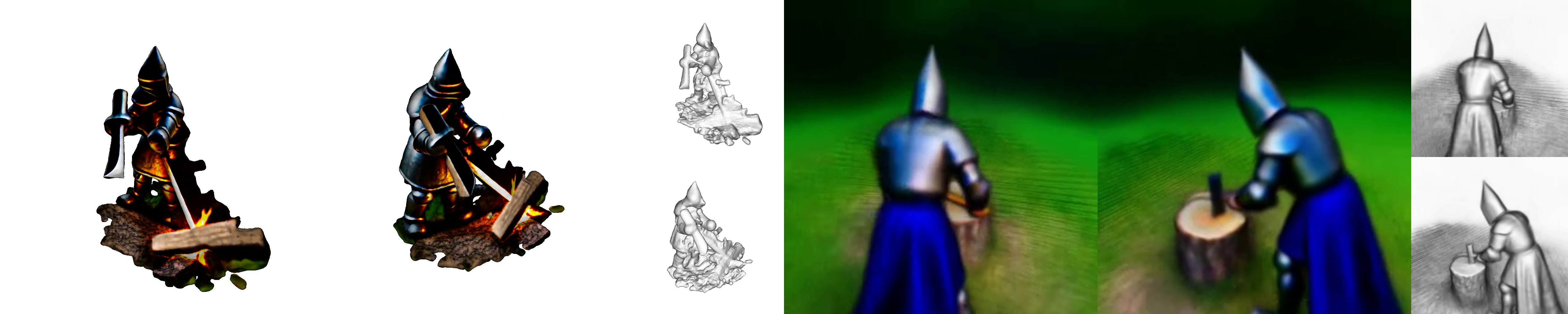} \\
        { \textit{a knight chopping wood$^{\ast}$}}  \\
        \includegraphics[align=c, width=\linewidth]{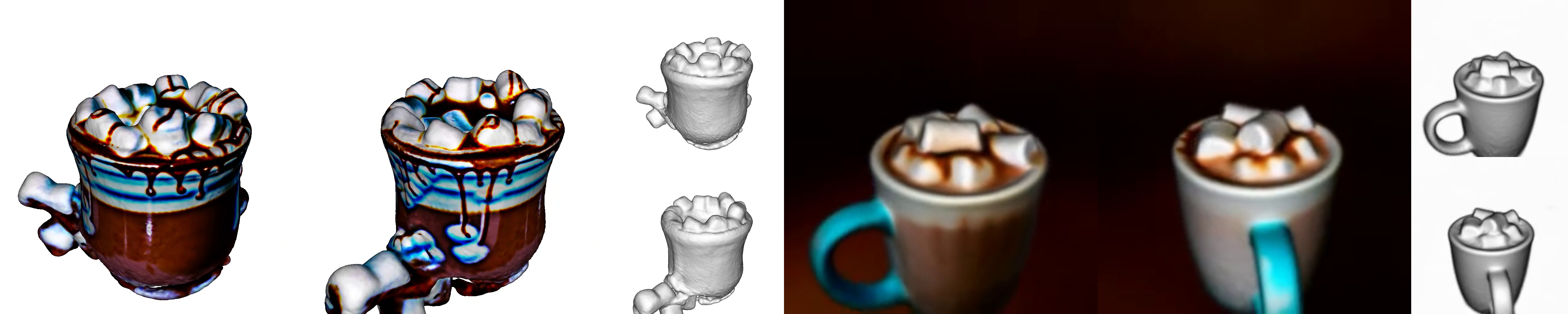}  \\
         { \textit{a mug of hot chocolate with whipped cream and marshmallows$^{\ast}$}}  \\
        \includegraphics[align=c, width=\linewidth]{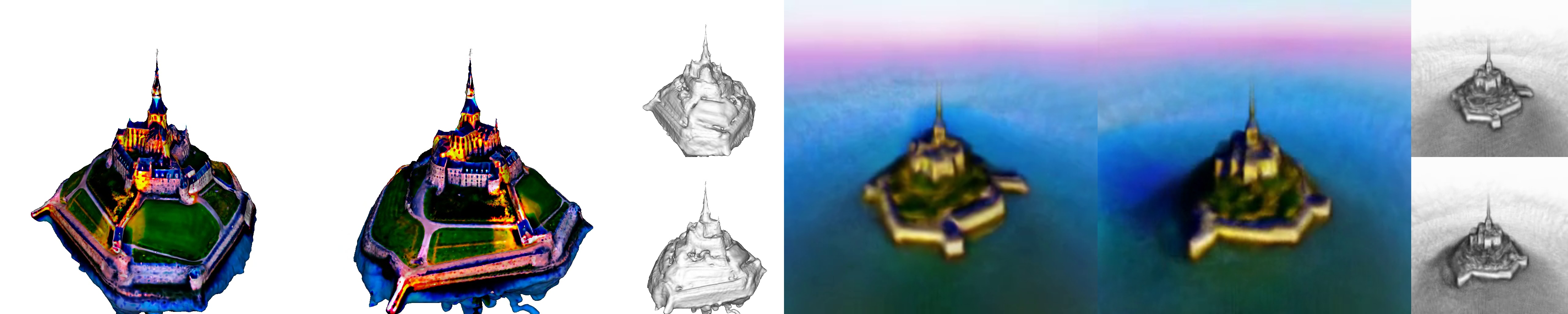} \\
         { \textit{an adorable piglet in a field$^{\ast}$}}  \\
        \includegraphics[align=c, width=\linewidth]{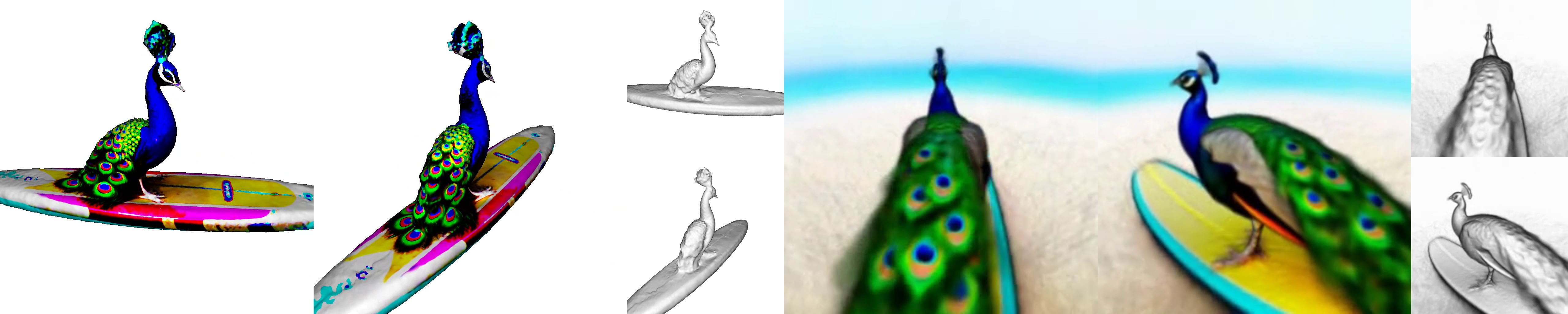} \\
         { \textit{a peacock on a surfboard$^{\ast}$}} 
         \end{tabular}
        \vspace{-2mm}
    \caption{\textbf{Qualitative comparison with DreamFusion~\cite{poole2022dreamfusion}.} We use the same text prompt as in DreamFusion. For each 3D model, we render it from two views with a textureless rendering for each view and remove the background to focus on the 3D shape. For the DreamFusion results, we take frames from the videos published on the official webpage. \name generates much higher quality 3D shapes on both geometry and texture compared with DreamFusion. 
\textit{$\ast$ a DSLR photo of... $\dagger$ a zoomed out DSLR photo of...} }
    \label{fig:suppl_compare_dreamfusion_2}
    \vspace{-2mm}
\end{figure*}

\begin{figure*}
    \centering 
    \setlength{\tabcolsep}{0.2pt}
    \begin{tabular}{c} 
      \begin{tabularx}{\textwidth}{ >{\centering\arraybackslash}X >{\centering\arraybackslash}X}
        Ours & DreamFusion~\cite{poole2022dreamfusion}
    \end{tabularx}\\
         \includegraphics[align=c, width=\linewidth]{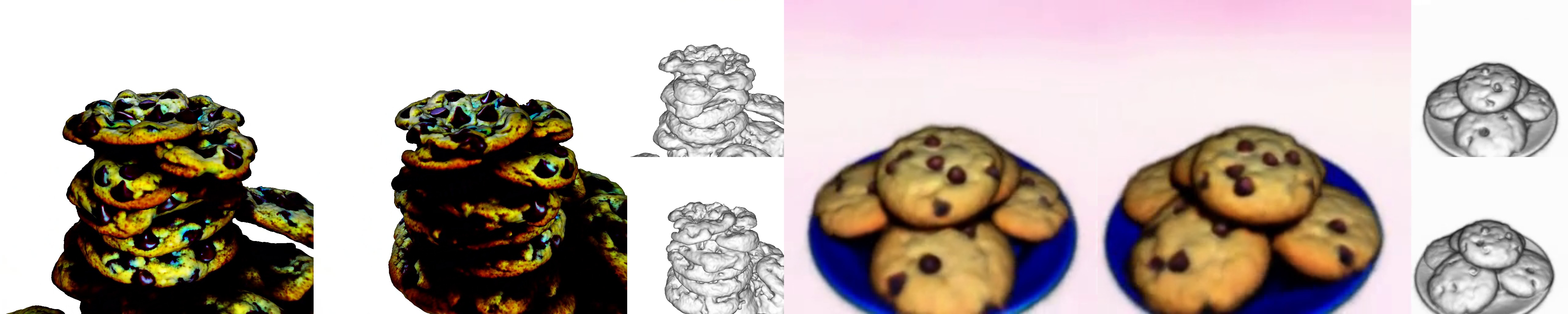} \\
        {\textit{a plate piled high with chocolate chip cookies$^{\dagger}$}} \\
        \includegraphics[align=c, width=\linewidth]{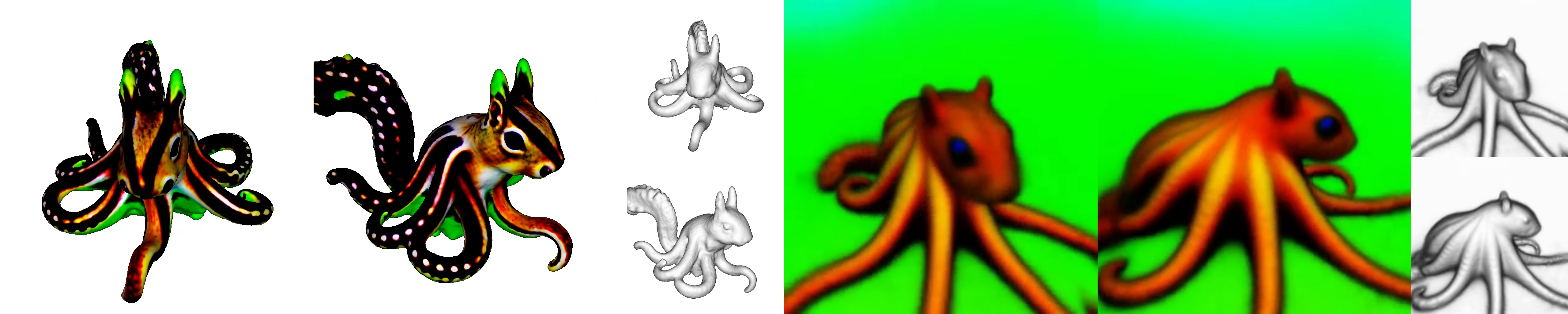} \\
        { \textit{a squirrel-octopus hybrid$^{\ast}$}}  \\
        \includegraphics[align=c, width=\linewidth]{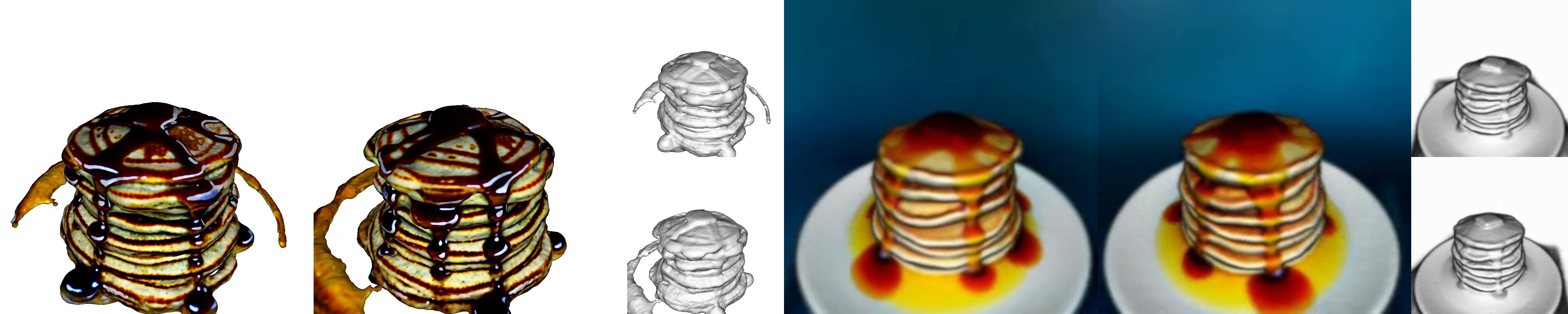}  \\
         { \textit{a stack of pancakes covered in maple syrup$^{\ast}$}}  \\
        \includegraphics[align=c, width=\linewidth]{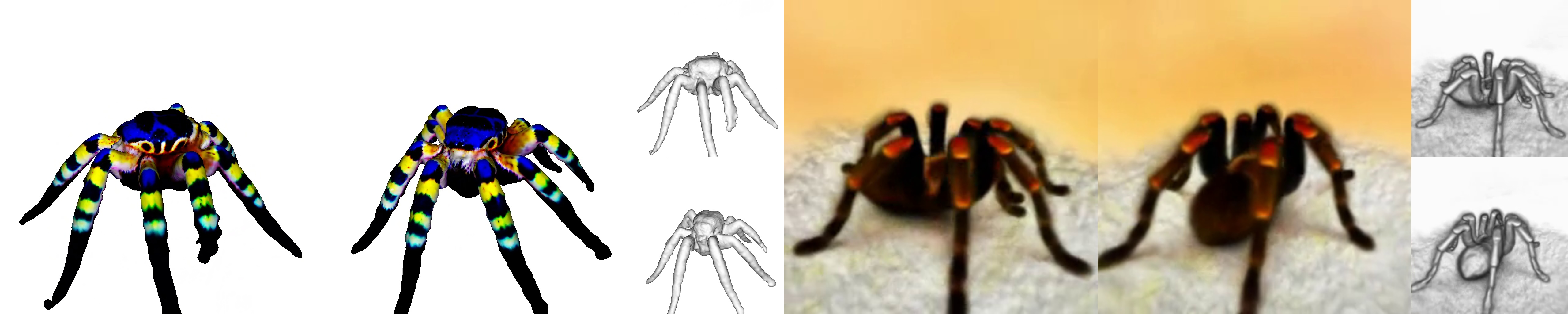} \\
         { \textit{a tarantula, highly detailed$^{\ast}$}}  \\
        \includegraphics[align=c, width=\linewidth]{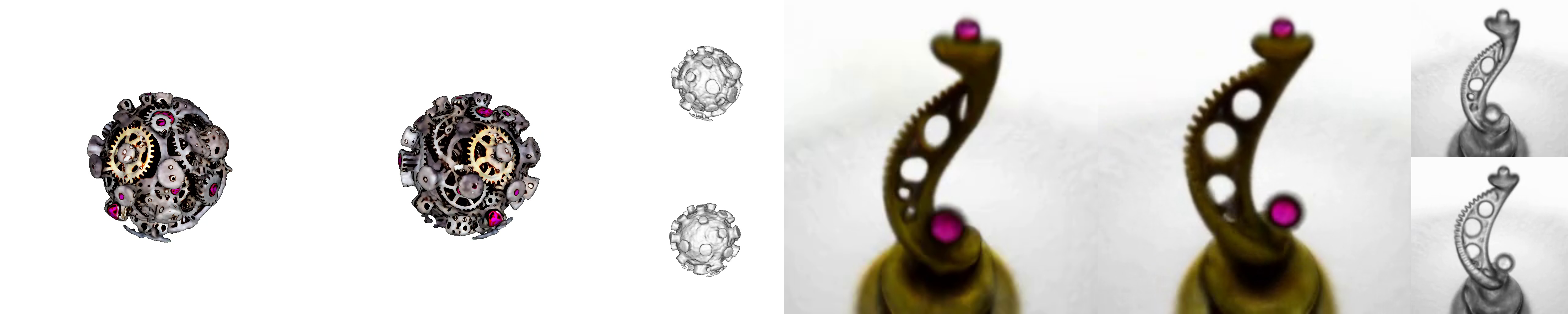} \\
        {\begin{tabular}[c]{@{}c@{}} \textit{ a very beautiful small organic sculpture made of fine clockwork and gears with tiny ruby bearings, very intricate, } \\ \textit{ caved, curved. Studio lighting, High resolution, white background$^{\ast}$}\end{tabular}}
         \end{tabular}
        \vspace{-2mm}
    \caption{\textbf{Qualitative comparison with DreamFusion~\cite{poole2022dreamfusion}.} We use the same text prompt as in DreamFusion. For each 3D model, we render it from two views with a textureless rendering for each view and remove the background to focus on the 3D shape. For the DreamFusion results, we take frames from the videos published on the official webpage. \name generates much higher quality 3D shapes on both geometry and texture compared with DreamFusion. 
\textit{$\ast$ a DSLR photo of... $\dagger$ a zoomed out DSLR photo of...} }
    \label{fig:suppl_compare_dreamfusion_3}
    \vspace{-2mm}
\end{figure*}

\begin{figure*}
    \centering 
    \setlength{\tabcolsep}{0.2pt}
    \begin{tabular}{c} 
      \begin{tabularx}{\textwidth}{ >{\centering\arraybackslash}X >{\centering\arraybackslash}X}
        Ours & DreamFusion~\cite{poole2022dreamfusion}
    \end{tabularx}\\
        \includegraphics[align=c, width=\linewidth]{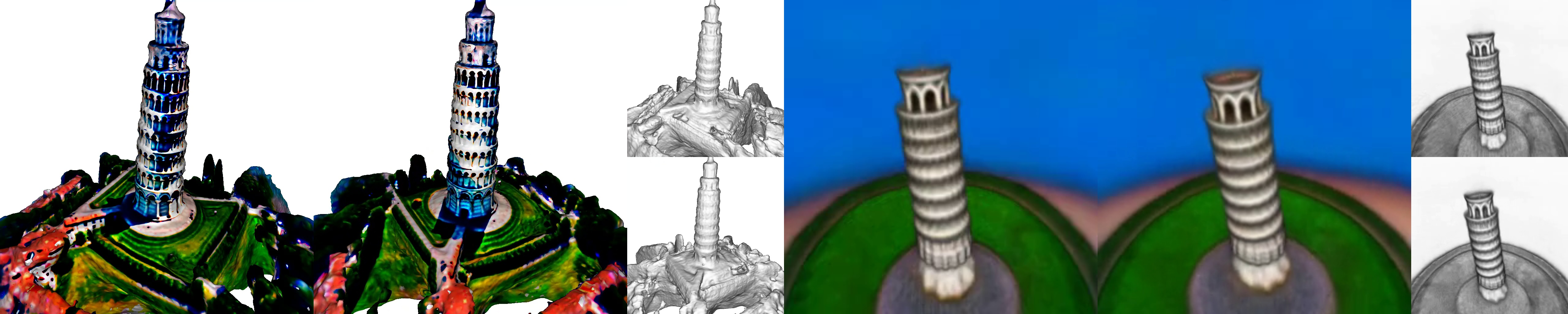}  \\
         { \textit{the leaning tower of Pisa, aerial view$^{\ast}$}}  \\
        \includegraphics[align=c, width=\linewidth]{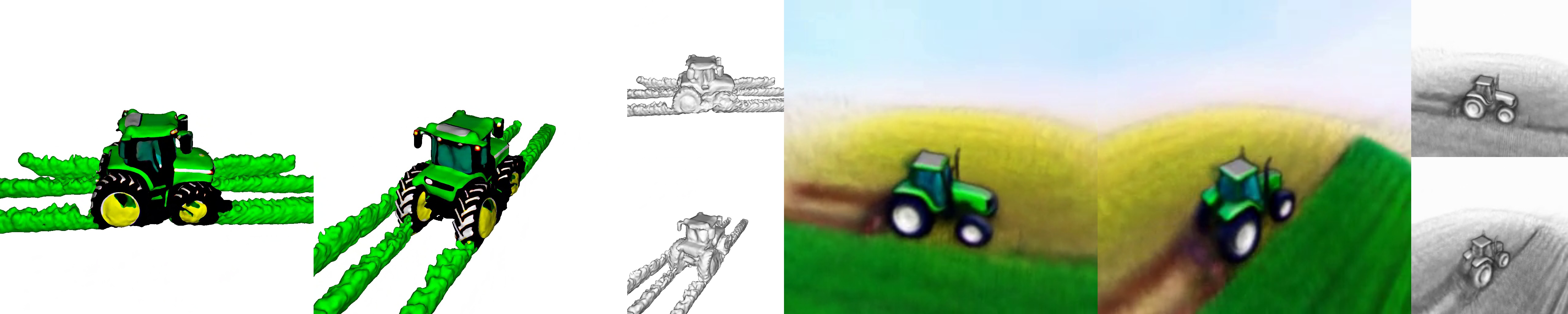} \\
         { \textit{a green tractor farming corn fields}} \\
         \includegraphics[align=c, width=\linewidth]{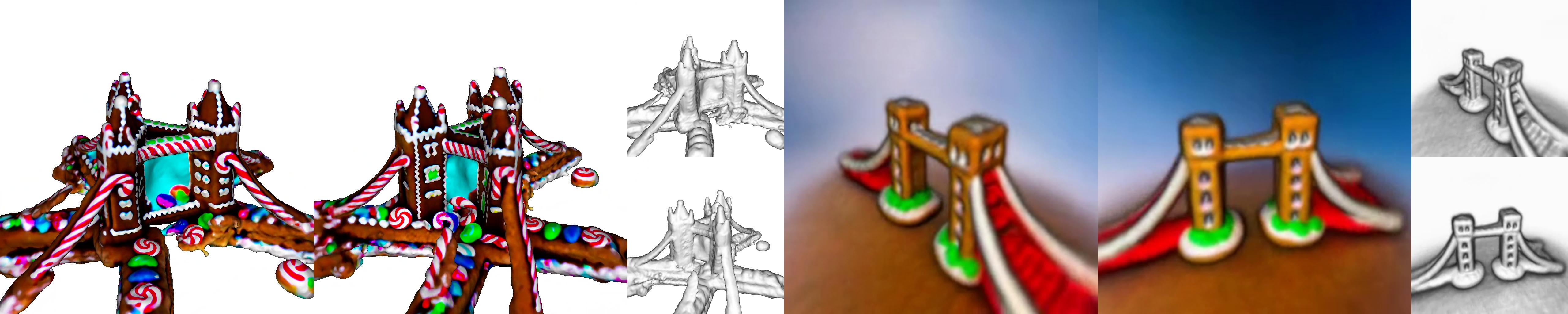} \\
        {\textit{a wide angle zoomed out DSLR photo of zoomed out view of Tower Bridge made out of gingerbread and candy$^{\dagger}$}} \\
        \includegraphics[align=c, width=\linewidth]{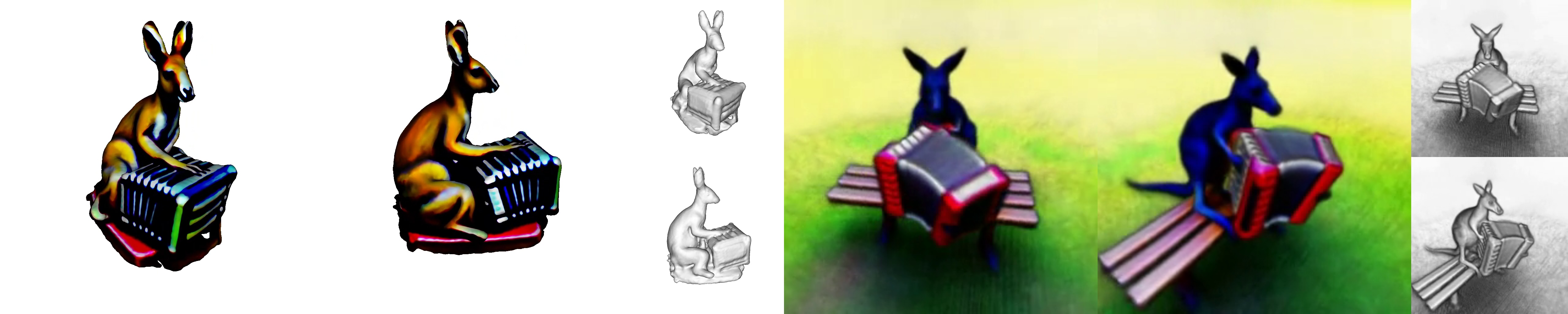} \\
        { \textit{a kangaroo sitting on a bench playing the accordion$^{\dagger}$}}  \\
         \end{tabular}
        \vspace{-2mm}
    \caption{\textbf{Qualitative comparison with DreamFusion~\cite{poole2022dreamfusion}.} We use the same text prompt as in DreamFusion. For each 3D model, we render it from two views with a textureless rendering for each view and remove the background to focus on the 3D shape. For the DreamFusion results, we take frames from the videos published on the official webpage. \name generates much higher quality 3D shapes on both geometry and texture compared with DreamFusion. 
\textit{$\ast$ a DSLR photo of... $\dagger$ a zoomed out DSLR photo of...} }
    \label{fig:suppl_compare_dreamfusion_4}
    \vspace{-2mm}
\end{figure*}

\begin{figure*}
    \centering 
    \setlength{\tabcolsep}{0.2pt}
    \begin{tabular}{c} 
      \begin{tabularx}{\textwidth}{ >{\centering\arraybackslash}X >{\centering\arraybackslash}X}
        Ours & DreamFusion~\cite{poole2022dreamfusion}
    \end{tabularx}\\
         \includegraphics[align=c, width=\linewidth]{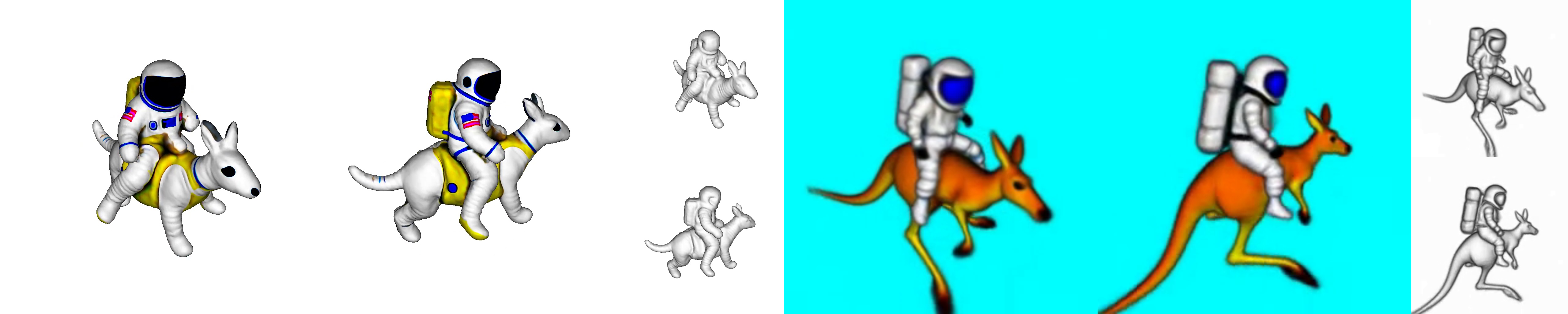} \\
        {\textit{an astronaut riding a kangaroo}} \\
        \includegraphics[align=c, width=\linewidth]{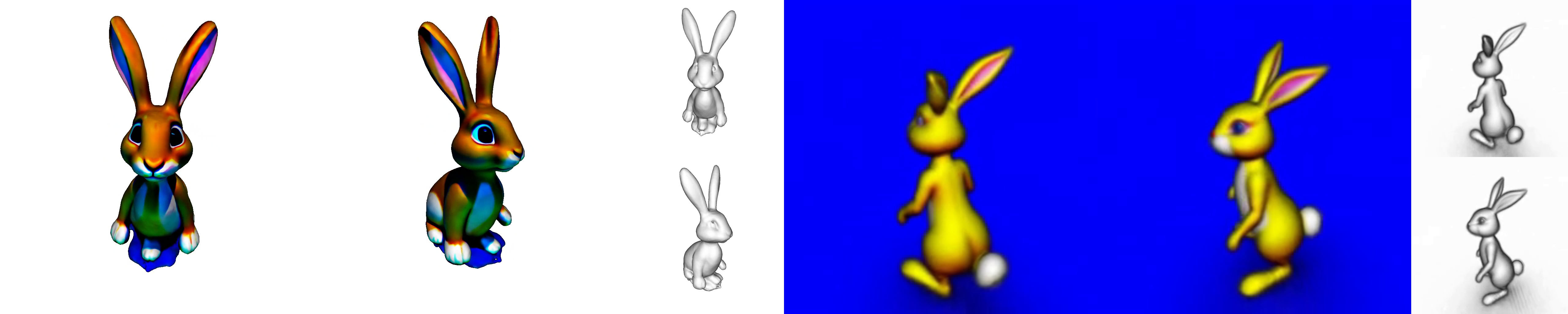} \\
        { \textit{ a rabbit, animated movie character, high detail 3d model}}  \\
        \includegraphics[align=c, width=\linewidth]{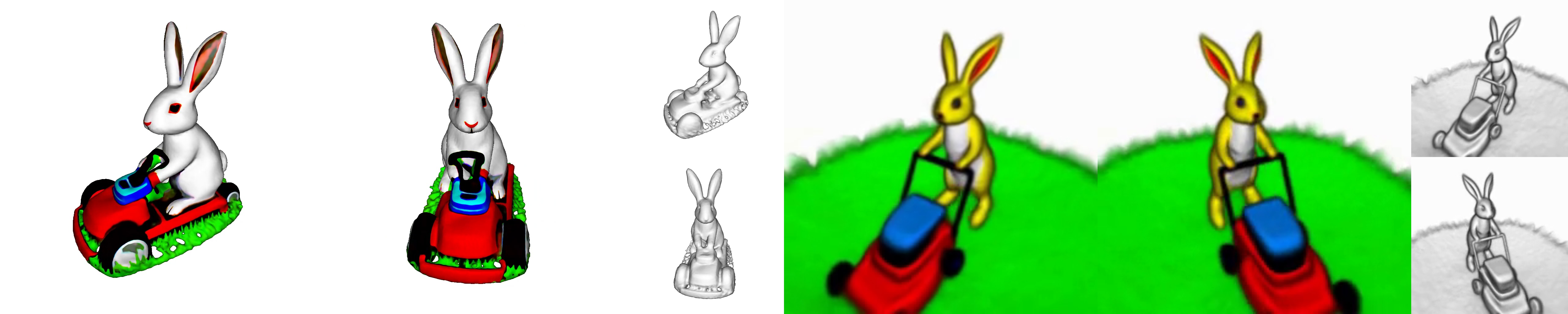}  \\
         { \textit{ a rabbit cutting grass with a lawnmower}}  \\
        \includegraphics[align=c, width=\linewidth]{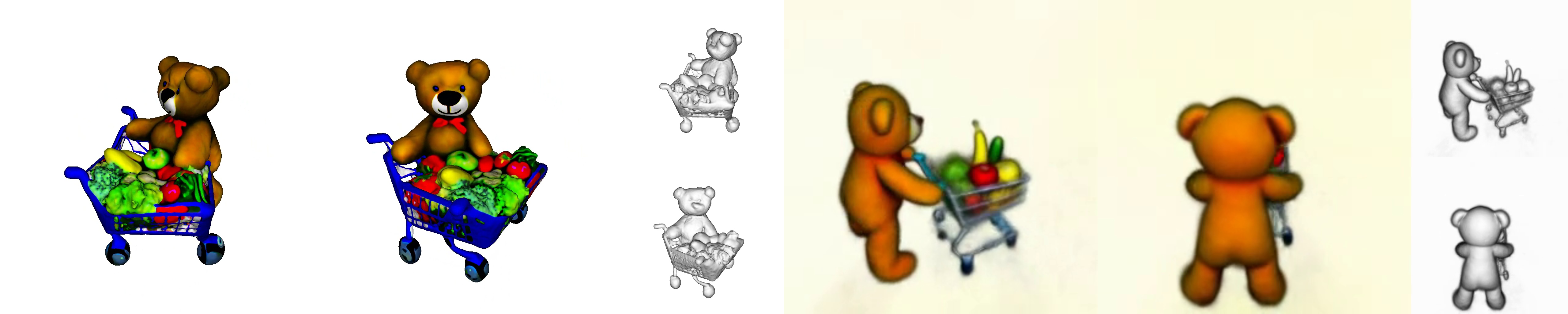} \\
         { \textit{a teddy bear pushing a shopping cart full of fruits and vegetables}} 
         \end{tabular}
        \vspace{-2mm}
    \caption{\textbf{Qualitative comparison with DreamFusion~\cite{poole2022dreamfusion}.} We use the same text prompt as in DreamFusion. For each 3D model, we render it from two views with a textureless rendering for each view and remove the background to focus on the 3D shape. For the DreamFusion results, we take frames from the videos published on the official webpage. \name generates much higher quality 3D shapes on both geometry and texture compared with DreamFusion. 
\textit{$\ast$ a DSLR photo of... $\dagger$ a zoomed out DSLR photo of...} }
    \label{fig:suppl_compare_dreamfusion_5}
    \vspace{-2mm}
\end{figure*}

We provide more qualitative comparisons with Dreamfusion~\cite{poole2022dreamfusion} in Figs.~\ref{fig:suppl_compare_dreamfusion_1},~\ref{fig:suppl_compare_dreamfusion_2},~\ref{fig:suppl_compare_dreamfusion_3},~\ref{fig:suppl_compare_dreamfusion_4},~\ref{fig:suppl_compare_dreamfusion_5}. Our \name achieved much higher quality in terms 3D geometry and texture. 

We also show more results on prompt-based editing in Fig.~\ref{fig:sup-fine-tuneSD-prompt}. Our \name enable high-quality editing of the 3D content through simple text prompt modification.

{\small
\bibliographystyle{ieee_fullname}
\bibliography{reference}

\begin{thebibliography}{10}\itemsep=-1pt

\bibitem{achlioptas2018learning}
Panos Achlioptas, Olga Diamanti, Ioannis Mitliagkas, and Leonidas Guibas.
\newblock Learning representations and generative models for 3d point clouds.
\newblock In {\em International conference on machine learning}, pages 40--49.
  PMLR, 2018.

\bibitem{balaji2022ediffi}
Yogesh Balaji, Seungjun Nah, Xun Huang, Arash Vahdat, Jiaming Song, Karsten
  Kreis, Miika Aittala, Timo Aila, Samuli Laine, Bryan Catanzaro, et~al.
\newblock ediff-i: Text-to-image diffusion models with an ensemble of expert
  denoisers.
\newblock {\em arXiv preprint arXiv:2211.01324}, 2022.

\bibitem{barron2022mip}
Jonathan~T Barron, Ben Mildenhall, Dor Verbin, Pratul~P Srinivasan, and Peter
  Hedman.
\newblock Mip-nerf 360: Unbounded anti-aliased neural radiance fields.
\newblock In {\em Proceedings of the IEEE/CVF Conference on Computer Vision and
  Pattern Recognition}, pages 5470--5479, 2022.

\bibitem{chan2022efficient}
Eric~R Chan, Connor~Z Lin, Matthew~A Chan, Koki Nagano, Boxiao Pan, Shalini
  De~Mello, Orazio Gallo, Leonidas~J Guibas, Jonathan Tremblay, Sameh Khamis,
  et~al.
\newblock Efficient geometry-aware 3d generative adversarial networks.
\newblock In {\em Proceedings of the IEEE/CVF Conference on Computer Vision and
  Pattern Recognition}, pages 16123--16133, 2022.

\bibitem{chan2021pi}
Eric~R Chan, Marco Monteiro, Petr Kellnhofer, Jiajun Wu, and Gordon Wetzstein.
\newblock pi-gan: Periodic implicit generative adversarial networks for
  3d-aware image synthesis.
\newblock In {\em Proceedings of the IEEE/CVF conference on computer vision and
  pattern recognition}, pages 5799--5809, 2021.

\bibitem{chen2019learning}
Zhiqin Chen and Hao Zhang.
\newblock Learning implicit fields for generative shape modeling.
\newblock In {\em Proceedings of the IEEE/CVF Conference on Computer Vision and
  Pattern Recognition}, pages 5939--5948, 2019.

\bibitem{gadelha20173d}
Matheus Gadelha, Subhransu Maji, and Rui Wang.
\newblock 3d shape induction from 2d views of multiple objects.
\newblock In {\em 2017 International Conference on 3D Vision (3DV)}, pages
  402--411. IEEE, 2017.

\bibitem{gao2020deftet}
Jun Gao, Wenzheng Chen, Tommy Xiang, Clement~Fuji Tsang, Alec Jacobson, Morgan
  McGuire, and Sanja Fidler.
\newblock Learning deformable tetrahedral meshes for 3d reconstruction.
\newblock In {\em Advances In Neural Information Processing Systems}, 2020.

\bibitem{gao2022get3d}
Jun Gao, Tianchang Shen, Zian Wang, Wenzheng Chen, Kangxue Yin, Daiqing Li, Or
  Litany, Zan Gojcic, and Sanja Fidler.
\newblock Get3d: A generative model of high quality 3d textured shapes learned
  from images.
\newblock {\em Advances in Neural Information Processing Systems}, 2022.

\bibitem{gu2021stylenerf}
Jiatao Gu, Lingjie Liu, Peng Wang, and Christian Theobalt.
\newblock Stylenerf: A style-based 3d aware generator for high-resolution image
  synthesis.
\newblock In {\em International Conference on Learning Representations}, 2022.

\bibitem{hao2021GANcraft}
Zekun Hao, Arun Mallya, Serge Belongie, and Ming-Yu Liu.
\newblock {GANcraft: Unsupervised 3D Neural Rendering of Minecraft Worlds}.
\newblock In {\em ICCV}, 2021.

\bibitem{henzler2019platonicgan}
Philipp Henzler, Niloy~J. Mitra, and Tobias Ritschel.
\newblock Escaping plato's cave: 3d shape from adversarial rendering.
\newblock In {\em The IEEE International Conference on Computer Vision (ICCV)},
  October 2019.

\bibitem{ho2020denoising}
Jonathan Ho, Ajay Jain, and Pieter Abbeel.
\newblock Denoising diffusion probabilistic models.
\newblock {\em Advances in Neural Information Processing Systems},
  33:6840--6851, 2020.

\bibitem{ho2021classifierfree}
Jonathan Ho and Tim Salimans.
\newblock Classifier-free diffusion guidance.
\newblock In {\em NeurIPS 2021 Workshop on Deep Generative Models and
  Downstream Applications}, 2021.

\bibitem{ibing2021octree}
Moritz Ibing, Gregor Kobsik, and Leif Kobbelt.
\newblock Octree transformer: Autoregressive 3d shape generation on
  hierarchically structured sequences.
\newblock {\em arXiv preprint arXiv:2111.12480}, 2021.

\bibitem{jain2021dreamfields}
Ajay Jain, Ben Mildenhall, Jonathan~T. Barron, Pieter Abbeel, and Ben Poole.
\newblock Zero-shot text-guided object generation with dream fields.
\newblock 2022.

\bibitem{khalid2022clip}
Nasir Khalid, Tianhao Xie, Eugene Belilovsky, and Tiberiu Popa.
\newblock Clip-mesh: Generating textured meshes from text using pretrained
  image-text models.
\newblock {\em ACM Transactions on Graphics (TOG), Proc. SIGGRAPH Asia}, 2022.

\bibitem{kingma2014adam}
Diederik Kingma and Jimmy Ba.
\newblock Adam: A method for stochastic optimization.
\newblock In {\em International Conference on Learning Representations}, 2015.

\bibitem{nvdiffrast}
Samuli Laine, Janne Hellsten, Tero Karras, Yeongho Seol, Jaakko Lehtinen, and
  Timo Aila.
\newblock Modular primitives for high-performance differentiable rendering.
\newblock {\em ACM Transactions on Graphics}, 39(6), 2020.

\bibitem{liu2020neural}
Lingjie Liu, Jiatao Gu, Kyaw Zaw~Lin, Tat-Seng Chua, and Christian Theobalt.
\newblock Neural sparse voxel fields.
\newblock {\em Advances in Neural Information Processing Systems},
  33:15651--15663, 2020.

\bibitem{liu2022compositional}
Nan Liu, Shuang Li, Yilun Du, Antonio Torralba, and Joshua~B Tenenbaum.
\newblock Compositional visual generation with composable diffusion models.
\newblock {\em arXiv preprint arXiv:2206.01714}, 2022.

\bibitem{lunz2020inverse}
Sebastian Lunz, Yingzhen Li, Andrew Fitzgibbon, and Nate Kushman.
\newblock Inverse graphics gan: Learning to generate 3d shapes from
  unstructured 2d data.
\newblock {\em arXiv preprint arXiv:2002.12674}, 2020.

\bibitem{luo2021diffusion}
Shitong Luo and Wei Hu.
\newblock Diffusion probabilistic models for 3d point cloud generation.
\newblock In {\em Proceedings of the IEEE/CVF Conference on Computer Vision and
  Pattern Recognition (CVPR)}, June 2021.

\bibitem{occnet}
Lars Mescheder, Michael Oechsle, Michael Niemeyer, Sebastian Nowozin, and
  Andreas Geiger.
\newblock Occupancy networks: Learning 3d reconstruction in function space.
\newblock In {\em Proceedings of the IEEE Conference on Computer Vision and
  Pattern Recognition}, pages 4460--4470, 2019.

\bibitem{mildenhall2020nerf}
Ben Mildenhall, Pratul~P. Srinivasan, Matthew Tancik, Jonathan~T. Barron, Ravi
  Ramamoorthi, and Ren Ng.
\newblock Nerf: Representing scenes as neural radiance fields for view
  synthesis.
\newblock In {\em ECCV}, 2020.

\bibitem{mo2019structurenet}
Kaichun Mo, Paul Guerrero, Li Yi, Hao Su, Peter Wonka, Niloy Mitra, and
  Leonidas Guibas.
\newblock Structurenet: Hierarchical graph networks for 3d shape generation.
\newblock {\em ACM Transactions on Graphics (TOG), Siggraph Asia 2019},
  38(6):Article 242, 2019.

\bibitem{muller2022instant}
Thomas M\"uller, Alex Evans, Christoph Schied, and Alexander Keller.
\newblock Instant neural graphics primitives with a multiresolution hash
  encoding.
\newblock {\em ACM Trans. Graph.}, 41(4):102:1--102:15, July 2022.

\bibitem{nvdiffrec}
Jacob Munkberg, Jon Hasselgren, Tianchang Shen, Jun Gao, Wenzheng Chen, Alex
  Evans, Thomas M{\"u}ller, and Sanja Fidler.
\newblock Extracting triangular 3d models, materials, and lighting from images.
\newblock In {\em CVPR}, pages 8280--8290, 2022.

\bibitem{nguyen2019hologan}
Thu Nguyen-Phuoc, Chuan Li, Lucas Theis, Christian Richardt, and Yong-Liang
  Yang.
\newblock Hologan: Unsupervised learning of 3d representations from natural
  images.
\newblock In {\em Proceedings of the IEEE/CVF International Conference on
  Computer Vision}, pages 7588--7597, 2019.

\bibitem{nichol2021glide}
Alex Nichol, Prafulla Dhariwal, Aditya Ramesh, Pranav Shyam, Pamela Mishkin,
  Bob McGrew, Ilya Sutskever, and Mark Chen.
\newblock Glide: Towards photorealistic image generation and editing with
  text-guided diffusion models.
\newblock {\em arXiv preprint arXiv:2112.10741}, 2021.

\bibitem{niemeyer2021giraffe}
Michael Niemeyer and Andreas Geiger.
\newblock Giraffe: Representing scenes as compositional generative neural
  feature fields.
\newblock In {\em Proceedings of the IEEE/CVF Conference on Computer Vision and
  Pattern Recognition}, pages 11453--11464, 2021.

\bibitem{orel2021styleSDF}
Roy Or-El, Xuan Luo, Mengyi Shan, Eli Shechtman, Jeong~Joon Park, and Ira
  Kemelmacher-Shlizerman.
\newblock Stylesdf: High-resolution 3d-consistent image and geometry
  generation.
\newblock In {\em Proceedings of the IEEE/CVF Conference on Computer Vision and
  Pattern Recognition}, pages 13503--13513, 2022.

\bibitem{poole2022dreamfusion}
Ben Poole, Ajay Jain, Jonathan~T Barron, and Ben Mildenhall.
\newblock Dreamfusion: Text-to-3d using 2d diffusion.
\newblock {\em arXiv preprint arXiv:2209.14988}, 2022.

\bibitem{radford2021learning}
Alec Radford, Jong~Wook Kim, Chris Hallacy, Aditya Ramesh, Gabriel Goh,
  Sandhini Agarwal, Girish Sastry, Amanda Askell, Pamela Mishkin, Jack Clark,
  et~al.
\newblock Learning transferable visual models from natural language
  supervision.
\newblock In {\em International Conference on Machine Learning}, pages
  8748--8763. PMLR, 2021.

\bibitem{ramesh2022hierarchical}
Aditya Ramesh, Prafulla Dhariwal, Alex Nichol, Casey Chu, and Mark Chen.
\newblock Hierarchical text-conditional image generation with clip latents.
\newblock {\em arXiv preprint arXiv:2204.06125}, 2022.

\bibitem{Rombach_2022_CVPR}
Robin Rombach, Andreas Blattmann, Dominik Lorenz, Patrick Esser, and Bj\"orn
  Ommer.
\newblock High-resolution image synthesis with latent diffusion models.
\newblock In {\em Proceedings of the IEEE/CVF Conference on Computer Vision and
  Pattern Recognition (CVPR)}, pages 10684--10695, June 2022.

\bibitem{ruiz2022dreambooth}
Nataniel Ruiz, Yuanzhen Li, Varun Jampani, Yael Pritch, Michael Rubinstein, and
  Kfir Aberman.
\newblock Dreambooth: Fine tuning text-to-image diffusion models for
  subject-driven generation.
\newblock {\em arXiv preprint arXiv:2208.12242}, 2022.

\bibitem{saharia2022photorealistic}
Chitwan Saharia, William Chan, Saurabh Saxena, Lala Li, Jay Whang, Emily
  Denton, Seyed Kamyar~Seyed Ghasemipour, Burcu~Karagol Ayan, S~Sara Mahdavi,
  Rapha~Gontijo Lopes, et~al.
\newblock Photorealistic text-to-image diffusion models with deep language
  understanding.
\newblock {\em arXiv preprint arXiv:2205.11487}, 2022.

\bibitem{sanghi2022clip}
Aditya Sanghi, Hang Chu, Joseph~G Lambourne, Ye Wang, Chin-Yi Cheng, Marco
  Fumero, and Kamal~Rahimi Malekshan.
\newblock Clip-forge: Towards zero-shot text-to-shape generation.
\newblock In {\em Proceedings of the IEEE/CVF Conference on Computer Vision and
  Pattern Recognition}, pages 18603--18613, 2022.

\bibitem{schwarz2022voxgraf}
Katja Schwarz, Axel Sauer, Michael Niemeyer, Yiyi Liao, and Andreas Geiger.
\newblock Voxgraf: Fast 3d-aware image synthesis with sparse voxel grids.
\newblock In {\em Advances in Neural Information Processing Systems (NeurIPS)},
  2022.

\bibitem{dmtet}
Tianchang Shen, Jun Gao, Kangxue Yin, Ming-Yu Liu, and Sanja Fidler.
\newblock Deep marching tetrahedra: a hybrid representation for high-resolution
  3d shape synthesis.
\newblock In {\em Advances in Neural Information Processing Systems (NeurIPS)},
  2021.

\bibitem{smith2017improved}
Edward~J Smith and David Meger.
\newblock Improved adversarial systems for 3d object generation and
  reconstruction.
\newblock In {\em Conference on Robot Learning}, pages 87--96. PMLR, 2017.

\bibitem{sohl2015deep}
Jascha Sohl-Dickstein, Eric Weiss, Niru Maheswaranathan, and Surya Ganguli.
\newblock Deep unsupervised learning using nonequilibrium thermodynamics.
\newblock In {\em International Conference on Machine Learning}, pages
  2256--2265. PMLR, 2015.

\bibitem{song2019generative}
Yang Song and Stefano Ermon.
\newblock Generative modeling by estimating gradients of the data distribution.
\newblock {\em Advances in Neural Information Processing Systems}, 32, 2019.

\bibitem{takikawa2021neural}
Towaki Takikawa, Joey Litalien, Kangxue Yin, Karsten Kreis, Charles Loop, Derek
  Nowrouzezahrai, Alec Jacobson, Morgan McGuire, and Sanja Fidler.
\newblock Neural geometric level of detail: Real-time rendering with implicit
  3d shapes.
\newblock In {\em Proceedings of the IEEE/CVF Conference on Computer Vision and
  Pattern Recognition}, pages 11358--11367, 2021.

\bibitem{KaolinWispLibrary}
Towaki Takikawa, Or Perel, Clement~Fuji Tsang, Charles Loop, Joey Litalien,
  Jonathan Tremblay, Sanja Fidler, and Maria Shugrina.
\newblock Kaolin wisp: A pytorch library and engine for neural fields research.
\newblock \url{https://github.com/NVIDIAGameWorks/kaolin-wisp}, 2022.

\bibitem{wu2016learning}
Jiajun Wu, Chengkai Zhang, Tianfan Xue, Bill Freeman, and Josh Tenenbaum.
\newblock Learning a probabilistic latent space of object shapes via 3d
  generative-adversarial modeling.
\newblock {\em Advances in neural information processing systems}, 29, 2016.

\bibitem{pointflow}
Guandao Yang, Xun Huang, Zekun Hao, Ming-Yu Liu, Serge Belongie, and Bharath
  Hariharan.
\newblock Pointflow: 3d point cloud generation with continuous normalizing
  flows.
\newblock In {\em Proceedings of the IEEE/CVF International Conference on
  Computer Vision}, pages 4541--4550, 2019.

\bibitem{zeng2022lion}
Xiaohui Zeng, Arash Vahdat, Francis Williams, Zan Gojcic, Or Litany, Sanja
  Fidler, and Karsten Kreis.
\newblock Lion: Latent point diffusion models for 3d shape generation.
\newblock In {\em Advances in Neural Information Processing Systems (NeurIPS)},
  2022.

\bibitem{zhang2020image}
Yuxuan Zhang, Wenzheng Chen, Huan Ling, Jun Gao, Yinan Zhang, Antonio Torralba,
  and Sanja Fidler.
\newblock Image gans meet differentiable rendering for inverse graphics and
  interpretable 3d neural rendering.
\newblock In {\em International Conference on Learning Representations}, 2021.

\bibitem{zhou2021pvd}
Linqi Zhou, Yilun Du, and Jiajun Wu.
\newblock 3d shape generation and completion through point-voxel diffusion.
\newblock In {\em Proceedings of the IEEE/CVF International Conference on
  Computer Vision}, pages 5826--5835, 2021.

\end{thebibliography}
}

\end{document}